\documentclass{article} 
\usepackage[final]{colm2026_conference}

\usepackage{microtype}
\usepackage{hyperref}
\usepackage{url}

\usepackage{booktabs}
\usepackage{graphicx}
\usepackage{amsmath}
\usepackage{amssymb}
\usepackage{xcolor}
\usepackage{multirow}
\usepackage{placeins}
\usepackage{lineno}
\usepackage{subcaption}
\usepackage{float}

\definecolor{darkblue}{rgb}{0, 0, 0.5}
\definecolor{langconsistent}{HTML}{FFC20A}
\definecolor{langagnostic}{HTML}{D55E00}
\definecolor{langspecific}{HTML}{57C4AB}
\hypersetup{colorlinks=true, citecolor=darkblue, linkcolor=darkblue, urlcolor=darkblue}

\title{LangFIR: Discovering Sparse Language-Specific Features from Monolingual Data for Language Steering}

\author{%
\begin{minipage}{\linewidth}
\normalfont\normalsize\raggedright
\textbf{Sing Hieng Wong}\textsuperscript{1}, \quad
\textbf{Hassan Sajjad}\textsuperscript{2}, \quad
\textbf{A.B. Siddique}\textsuperscript{1} \\[0.35em]
\textsuperscript{1}Department of Computer Science, University of Kentucky, USA \\
\textsuperscript{2}Department of Computer Science, Dalhousie University, Canada \\
Correspondence: \texttt{singhieng.wong@uky.edu}
\end{minipage}
}

\begin{document}

\ifcolmsubmission
\linenumbers
\fi

\maketitle

\begin{abstract}
Large language models~(LLMs) show strong multilingual capabilities, yet reliably controlling the language of their outputs remains difficult. 
Representation-level steering addresses this by adding language-specific vectors to model activations at inference time, but identifying language-specific directions in the residual stream often relies on multilingual or parallel data that can be expensive to obtain.
Sparse autoencoders~(SAEs) decompose residual activations into interpretable, sparse feature directions and offer a natural basis for this search, yet existing SAE-based approaches face the same data constraint.
We introduce \textbf{LangFIR} (\textbf{Lang}uage \textbf{F}eature \textbf{I}dentification via \textbf{R}andom-token Filtering), a method that discovers language-specific SAE features using only a small amount of monolingual data and random-token sequences.
Many SAE features consistently activated by target-language inputs do not encode language identity. Random-token sequences surface these language-agnostic features, allowing LangFIR to filter them out and isolate a sparse set of language-specific features.
We show that these features are extremely sparse, highly selective for their target language, and causally important: directional ablation increases cross-entropy loss only for the corresponding language.
Using these features to construct steering vectors for the multilingual generation control task, LangFIR achieves the best average accuracy $\times$ BLEU among steering methods across three models (Gemma~3~1B, Gemma~3~4B, and Llama~3.1~8B), three datasets, and twelve target languages, outperforming the strongest monolingual baseline by up to $4.7\times$ and surpassing methods that use parallel data. 
Our results suggest that language identity in multilingual LLMs is localized in a sparse set of feature directions discoverable with monolingual data. Code is available at \url{https://github.com/JaMussCraft/LangFIR}.

\end{abstract}

\section{Introduction}

Large language models (LLMs) exhibit impressive capabilities across text generation, instruction following, and reasoning~\citep{grattafiori2024llama, Kamath2025Gemma3T}. 
Considerable effort has also been devoted to improving their multilingual abilities, motivated by the growing need to deploy them across diverse linguistic settings.
While multilingual LLMs can generate text in many languages, reliably controlling which language they output (a task known as multilingual generation control) remains a challenge~\citep{marchisio2024understanding, gurgurov2026clas}.
Representation-level steering provides a training-free alternative that enables direct intervention on model activations at inference time, while also offering insights into how models encode language information.

Recent representation-level approaches to multilingual control locate language-specific structure in model activations in different ways.
Neuron-based methods~\citep{tang2024language, kojima2024multilingual} detect language-discriminative neurons, but require logging activations over millions of tokens across multiple languages.
Sparse autoencoders~(SAEs) decompose activations into sparse, interpretable feature directions, providing a natural basis for identifying language-specific directions.
Yet SAE-based approaches~\citep{deng-etal-2025-unveiling, chou2025causal} similarly rely on multilingual data to identify language-specific features.
More recently, \citet{zhong2025language} propose a data-efficient method, but its stronger variant still depends on parallel sentence pairs.
A key limitation of most existing approaches is their reliance on multilingual data, often parallel corpora, which is restrictive in domains where such resources are unavailable or costly to curate.
Identifying language-specific features from small amounts of monolingual data would therefore substantially broaden the applicability of inference-time language control.


\begin{figure*}[t]
  \centering
  \includegraphics[width=\textwidth]{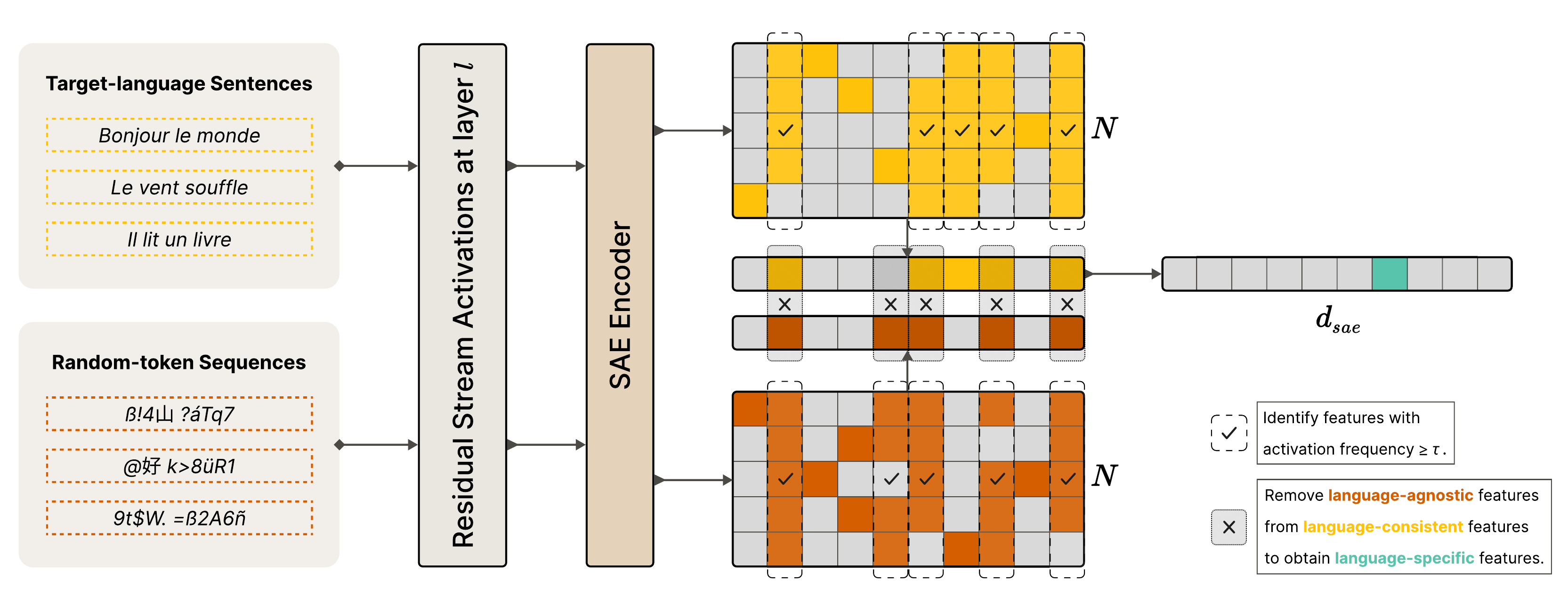}
  \caption{Overview of LangFIR. (1) For each target-language sentence, a random-token sequence is generated by sampling from the tokenizer's vocabulary. (2) Residual-stream activations are extracted at a layer $l$ and encoded through SAE. (3) Sample-wise filtering is performed with activation frequency threshold $\tau$ to identify features that activate frequently for target-language sentences (\textcolor{langconsistent}{language-consistent}) and for random-token sequences (\textcolor{langagnostic}{language-agnostic}). (4) Removing language-agnostic features from language-consistent features yields a sparse set of \textcolor{langspecific}{language-specific features}.}
  \label{fig:langfir_method}
\end{figure*}

We observe that when encoding target-language text through an SAE, many highly activated features are not language-specific: they activate across all languages, responding to non-linguistic tokens such as punctuation patterns and whitespace.
We hypothesize that random-token sequences, which lack coherent linguistic content, provide a simple proxy for surfacing such language-agnostic features. Removing them from the features activated consistently by target-language text should leave features that genuinely encode language identity.

Building on this intuition, we introduce \textbf{LangFIR} (\textbf{Lang}uage \textbf{F}eature \textbf{I}dentification via \textbf{R}andom-token Filtering). Our method collects a set of target-language sentences and generates a set of random-token sequences, encoding their residual-stream activations using an SAE. 
It then identifies features that activate frequently for the target language while filtering out those that also activate frequently for random tokens. 
This yields a sparse set of language-specific features.

We show these features activate selectively for their target language, are causally important, as ablating them substantially increases cross-entropy loss only for the target language, and concentrate their effects most strongly in later model layers. 
We then use language-specific features to construct steering vectors for controlled multilingual generation.
Across three models, three datasets, and twelve target languages, steering vectors constructed from these features achieve the best average accuracy $\times$ BLEU. They outperform the strongest monolingual baseline by up to $4.7\times$ and the strongest overall baseline by up to 2.7 points.

This work makes the following contributions:

\begin{itemize}
\item We propose LangFIR, a monolingual-data-only method that identifies sparse language-specific SAE features via random-token filtering.
\item We show these features are language-selective and causally important under directional ablation, with effects concentrated in later model layers.
\item We demonstrate that steering vectors built from these features outperform strong baselines across models and datasets on the multilingual generation control task.
\end{itemize}

\section{Related Work}

\subsection{Multilingual Representations in LLMs}

Prior work has revealed a layer-wise progression in how LLMs process multilingual input. \citet{wendler-etal-2024-llamas} find that in English-centric models such as Llama~2, early layers encode general lexical and semantic features, intermediate layers map representations into an English-aligned concept space, and later layers shift toward target-language token spaces. \citet{dumas-etal-2025-separating} provide further evidence for language-agnostic concept representations in intermediate layers, showing that language and concept information can be separated through activation patching. 
Similarly, \citet{gurgurov2026clas} demonstrate that language-specific structure emerges predominantly in later layers and that steering directions cluster by language family.
In a contrastive approach related to LangFIR, \citet{alkhamissi2025llm} compare sentences against strings of non-words to localize residual-stream units selective for natural language in general, rather than for a specific target language.

\subsection{Language Steering}

A broad line of work explores controlling an LLM's behavior by manipulating its internal representations. \citet{turner2023steering} and~\citet{rimsky2024steering} demonstrate that simple vector-based interventions can steer models toward desired behaviors such as improved truthfulness. Language steering applies this paradigm to multilingual generation, using representation interventions to control which language a model outputs. 
\citet{gurgurov2026clas} propose CLaS-Bench, a standardized benchmark for evaluating multilingual language steering, finding that the simple DiffMean method on residual activations~\citep{marks2023geometry} consistently outperforms more complex approaches.

\textbf{Neuron-based methods.} \citet{kojima2024multilingual} train binary classifiers on neuron activations to identify language-discriminative neurons and propose replacement-based manipulation. \citet{tang2024language} introduce Language Activation Probability Entropy (LAPE), which identifies language-specific neurons through entropy-based filtering and steers generation by selectively activating and deactivating them. 
However, neuron-based methods can be unreliable due to superposition~\citep{elhage2022toy}, where multiple concepts are encoded in a single neuron, and they typically require large-scale multilingual corpora.

\textbf{Dimension-based methods.} \citet{zhong2025language} identify sparse language-specific dimensions in the residual stream and steer multilingual generation by overriding activations in these dimensions. 
Their approach requires a small amount of monolingual or parallel data. 
While data-efficient, the method can be sensitive to hyperparameter choices and model architecture, and its stronger variant relies on parallel data.

\textbf{SAE-based methods.} SAEs have been used for uncovering interpretable, language-specific features. \citet{deng-etal-2025-unveiling} introduce a monolinguality metric for SAE features and show that ablating language-specific features selectively impairs performance in one language while leaving others unaffected. They also demonstrate that these features can enhance steering vectors. \citet{chou2025causal} show that modulating a small number of SAE features in mid-to-late layers can steer generation toward target languages with high success rates. However, these methods typically require multilingual parallel corpora for feature identification. 
In contrast, our method removes this requirement by identifying language-specific SAE features using only monolingual data.

\section{Preliminaries}
\label{sec:prelim}

\textbf{Sparse Autoencoders.}
Given a residual-stream activation $\mathbf{x} \in \mathbb{R}^{d_{\text{res}}}$ at layer $l$, an SAE computes a sparse latent activation $\mathbf{f}(\mathbf{x}) \in \mathbb{R}^{d_{\text{sae}}}$ and reconstructs the input as $\hat{\mathbf{x}}$:
\begin{equation}
\mathbf{f}(\mathbf{x}) := \text{ReLU}(W_\text{enc}\,\mathbf{x} + \mathbf{b}_\text{enc}), \qquad
\hat{\mathbf{x}}(\mathbf{f}) := W_{\mathrm{dec}}\,\mathbf{f} + \mathbf{b}_\text{dec}. \label{eq:sae}
\end{equation}
To encourage sparsity, \citet{bricken2023monosemanticity} and~\citet{huben2023sparse} incorporate an $L_1$ penalty on $\mathbf{f}(\cdot)$ in the training loss. An alternative is the Top-$K$ SAE~\citep{gao2024scaling}, which enforces sparsity by retaining only the $K$ largest activations. The JumpReLU variant~\citep{rajamanoharan2024jumping} replaces the standard ReLU with a learned threshold $\theta$, setting $\text{JumpReLU}(z) = z \cdot 1[z > \theta]$. By projecting activations into a higher-dimensional latent space, SAEs yield features that are significantly more interpretable than individual neurons, capturing specific syntactic patterns~\citep{huben2023sparse}, entity recognition signals~\citep{ferrando2024know}, and other semantically coherent concepts~\citep{paulo2024automatically}. 

Following~\citet{rajamanoharan2024jumping}, we write $W_{\mathrm{dec}} = [\mathbf{d}_1, \dots, \mathbf{d}_{d_{\text{sae}}}]$, where each column $\mathbf{d}_j \in \mathbb{R}^{d_{\text{res}}}$ is a feature direction. We refer to each such direction as a \emph{feature} throughout this paper.
We use residual-stream activations at layer \(l\) as the input $\mathbf{x}$, as they are more interpretable than other internal states~\citep{ferrando2024know, chanin2024absorption}. Most prior steering methods also operate on the residual stream~\citep{gurgurov2026clas, zhong2025language}, enabling a fair comparison.

\section{LangFIR: Language Feature Identification via Random-Token Filtering}
\label{sec:method}

We observed that encoding target-language text through an SAE produces a set of highly activated features, many of which are shared across all languages. 
Inspection of these features using Neuronpedia~\citep{neuronpedia} revealed that many do not increase the logits of target-language tokens. Instead, they appear to respond to language-universal properties (e.g., punctuation and whitespace) or non-linguistic symbol patterns (see examples in Appendix~\ref{sec:appendix_noise_features}).
Since these features correlate with generic textual structure rather than language identity, we hypothesize that random-token sequences should activate them consistently. 
Filtering out such features from the set of features activated consistently by target-language text should leave a set of features that genuinely encodes language identity. The method proceeds in three stages.

\textbf{Random-Token Sequence Generation.}
For each target language, we sample $N$ monolingual sentences from a target-language corpus. 
For each sentence, we generate a random-token sequence with a similar number of tokens by sampling uniformly from the tokenizer's vocabulary (excluding special tokens, control characters, and reserved entries). 

\textbf{Activation Extraction and SAE Encoding.}
Both target-language sentences and random-token sequences are traced through the LLM's forward pass, and residual-stream activations are collected at layer $l$. For each sample $i$, we encode each token's residual activation $\mathbf{x}_{i,t} \in \mathbb{R}^{d_{\text{res}}}$ with the SAE and average the resulting per-token SAE activations across the sequence to obtain a sample-level SAE activation vector $\mathbf{f}_i \in \mathbb{R}^{d_{\text{sae}}}$. This gives per-sample SAE activation vectors for target-language and random-token sequences, $\mathbf{f}_i^{\text{lang}}$ and $\mathbf{f}_i^{\text{rand}}$.
Stacking across \(N\) samples gives two latent activation matrices, $\mathbf{F}^{\text{lang}}, \mathbf{F}^{\text{rand}} \in \mathbb{R}^{N \times d_{\text{sae}}}$.

\textbf{Sample-wise Filtering and Feature Overlap Removal.}
A feature $j$ is considered high-frequency if it activates (i.e., has nonzero activation) for at least a fraction $\tau \in (0,1]$ of samples. We define sample-level activation frequencies:
\begin{equation}
\rho_j^{\text{lang}} = \frac{1}{N}\sum_{i=1}^{N} 1\!\left[\mathbf{F}^{\text{lang}}_{ij} > 0\right], \qquad \rho_j^{\text{rand}} = \frac{1}{N}\sum_{i=1}^{N} 1\!\left[\mathbf{F}^{\text{rand}}_{ij} > 0\right].
\end{equation}
Thresholding these frequencies gives:
\begin{itemize}
\item \emph{Language-consistent features}: consistently activated by target-language sentences, $S_\text{lang} = \{j : \rho_j^{\text{lang}} \ge \tau\}$.
\item \emph{Language-agnostic features}: consistently activated by random-token sequences, $S_\text{rand} = \{j : \rho_j^{\text{rand}} \ge \tau\}$.
\end{itemize}
The \emph{language-specific} feature set, $S_\text{spec} = S_\text{lang} \setminus S_\text{rand}$, is the language-consistent features with language-agnostic features removed.

\section{Analysis of Language-Specific Features}
\label{sec:analyses}

We present analyses on Llama 3.1 8B~\citep{grattafiori2024llama} using Llama Scope SAEs~\citep{he2024llama}, with data sampled from FLORES+~\citep{goyal2022flores, costa2022no}, a high-quality multilingual parallel translation dataset covering over 200 languages. 
Unless otherwise noted, we use a default configuration of sample size $N = 100$, layer percentile~$=~0.9$, and $\tau = 1.0$. Analyses focus on a set of five languages: English, Portuguese, German, Polish, and Chinese.

\subsection{Sparsity and Stability of Feature Identification}

\begin{figure}[t]
  \centering
  \begin{subfigure}[t]{0.49\columnwidth}
    \centering
    \includegraphics[width=\linewidth]{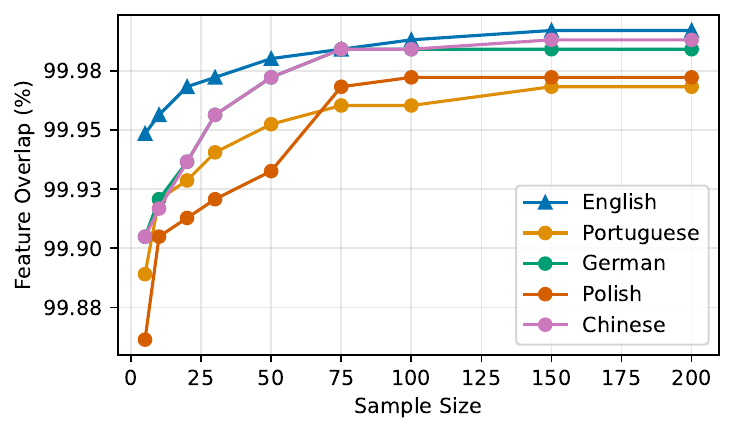}
    \caption{Feature set overlap vs sample size.}
    \label{llama:ls_features_overlap_vs_sentences}
  \end{subfigure}
  \hfill
  \begin{subfigure}[t]{0.49\columnwidth}
    \centering
    \includegraphics[width=\linewidth]{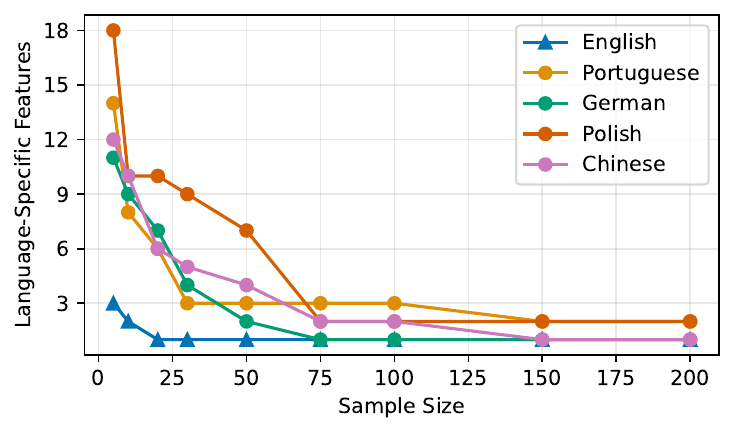}
    \caption{Language-specific feature count vs sample size.}
    \label{llama:ls_features_vs_sentences}
  \end{subfigure}
\vspace{-4pt}
  \caption{Analyses of language-specific feature identification across sample sizes on Llama~3.1~8B. (a)~Overlap between language-consistent and language-agnostic feature sets approaches 100\%. (b)~Fewer than 5 language-specific features remain per language. Both quantities stabilize by ${\sim}$100 sample size. }
 \vspace{-4pt} \label{llama:ls_overlap_and_count_vs_sentences}
\end{figure}

\textbf{Extreme sparsity and stability with minimal monolingual data.}
Figure~\ref{llama:ls_overlap_and_count_vs_sentences} shows the two key outcomes of LangFIR's filtering stage across sample sizes. The overlap between the language-consistent and language-agnostic feature sets approaches 100\% (Figure~\ref{llama:ls_features_overlap_vs_sentences}), meaning that random-token filtering removes the vast majority of candidate features. What remains is an extremely sparse set of language-specific features, typically fewer than five per language (Figure~\ref{llama:ls_features_vs_sentences}). 
Both quantities stabilize by around 100 sentences, providing a reliable identification signal. 
Convergence curves for the intermediate feature sets are provided in Figure~\ref{llama:la_lc_features_vs_sentences}.

\textbf{Stability with respect to threshold.}
The number of language-specific features stabilizes for $\tau \geq 0.8$ (Figure~\ref{llama:ls_features_vs_threshold}).
Extremely high thresholds may exclude weaker language-specific features (Appendix~\ref{sec:appendix_ablation_details}, Figure~\ref{llama:ablation_threshold}), but the method is robust across a broad range of high threshold values.

\textbf{English has few language-specific features.}
English consistently yields fewer language-specific features than other languages. We can attribute this to the English bias of multilingual LLMs~\citep{wendler-etal-2024-llamas}: general concepts may be entangled with English representations, making it harder for SAEs to isolate English-specific features.

\subsection{Properties of Language-Specific Features}

\begin{figure}[t]
  \centering
  \begin{subfigure}[t]{0.49\columnwidth}
    \centering
    \includegraphics[width=\linewidth]{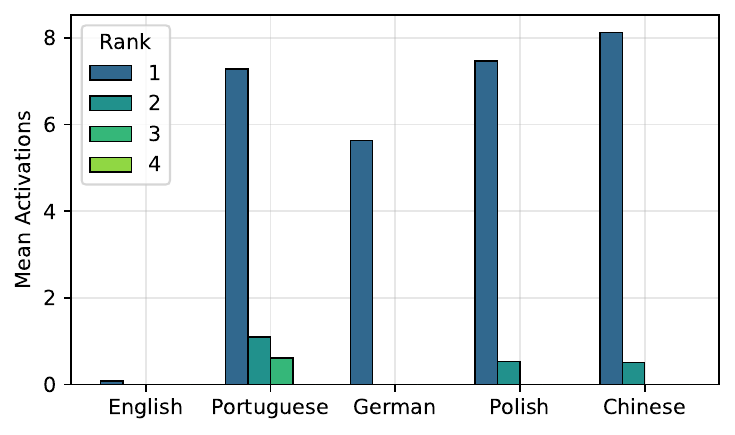}
    \caption{Activations of top-ranked language-specific features.}
    \label{llama:top_feature_activations}
  \end{subfigure}
  \hfill
  \begin{subfigure}[t]{0.49\columnwidth}
    \centering
    \includegraphics[width=\linewidth]{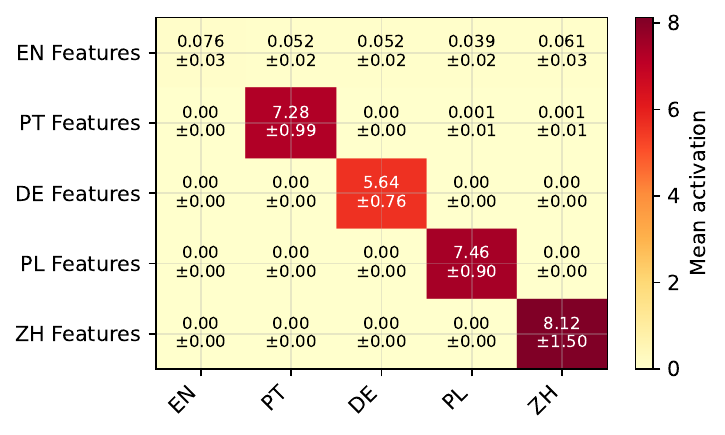}
    \caption{Language-specific feature activations across evaluation languages.}
    \label{llama:ls_feature_activations_heatmap}
  \end{subfigure}
  \caption{Analyses of language-specific feature properties on Llama~3.1~8B. (a)~Rank-1 features dominate in mean activation magnitude. (b)~Language-specific features activate selectively on their target language and remain near zero on others.}
\label{llama:ls_and_non_ls_feature_activations}
\end{figure}

\textbf{Dominance of the top feature.}
For all non-English languages, the rank-1 feature dominates in mean activation magnitude, the average activation over target-language samples (Figure~\ref{llama:top_feature_activations}). English is the sole exception: its rank-1 feature activates at near-zero magnitude.

\textbf{Language-selective activation.}
Given the dominance of the top-ranked feature, we visualize the mean activation of each language's rank-1 feature across texts in all five languages in Figure~\ref{llama:ls_feature_activations_heatmap}. Each feature activates strongly only on its corresponding target language, with near-zero activation for all others. In contrast, features not identified as language-specific for any language activate broadly and uniformly across languages, with English showing a notably higher mean activation (Figure~\ref{llama:non_ls_feature_activations}).

\textbf{Late-layer activation dominance.}
Language-specific feature counts peak at the earliest and latest layers, with moderate counts in between (Figure~\ref{llama:ls_features_vs_layers}), consistent with the finding that multilingual LLMs shift from English-aligned intermediate representations to target-language spaces in later layers~\citep{wendler-etal-2024-llamas}. Moreover, late-layer features activate significantly more strongly than early-layer ones (Figure~\ref{llama:ls_feature_activations_vs_layers}), suggesting that language-specific information concentrates in later layers~\citep{gurgurov2026clas}.

\subsection{Validating Language-Specific Features}
\label{sec:validation}

Following~\citet{deng-etal-2025-unveiling}, we examine whether the identified features causally influence model behavior by using directional ablation~\citep{arditi2024refusal, ferrando2024know} to ``zero out'' language-specific features. Given a feature direction $\mathbf{d} \in \mathbb{R}^{d_{\text{res}}}$, directional ablation removes its contribution from the residual stream activation $\mathbf{x}$ via $\mathbf{x}' \leftarrow \mathbf{x} - \hat{\mathbf{d}}\,\hat{\mathbf{d}}^\top \mathbf{x}$, where $\hat{\mathbf{d}}$ is the unit vector of $\mathbf{d}$. For each target language and layer in Llama~3.1~8B, we ablate the top-2 language-specific features and measure changes in cross-entropy (CE) loss for texts in the target language versus texts in other languages.

\begin{figure*}[t]
  \centering
  \begin{subfigure}[t]{0.32\linewidth}
    \centering
    \includegraphics[width=\linewidth]{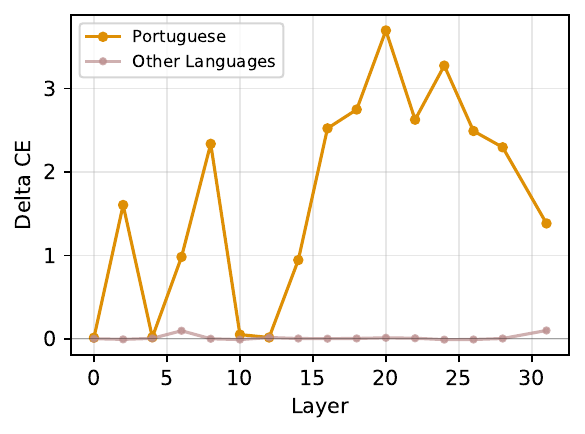}
    \caption{Ablate Portuguese features.}
  \end{subfigure}\hfill
  \begin{subfigure}[t]{0.32\linewidth}
    \centering
    \includegraphics[width=\linewidth]{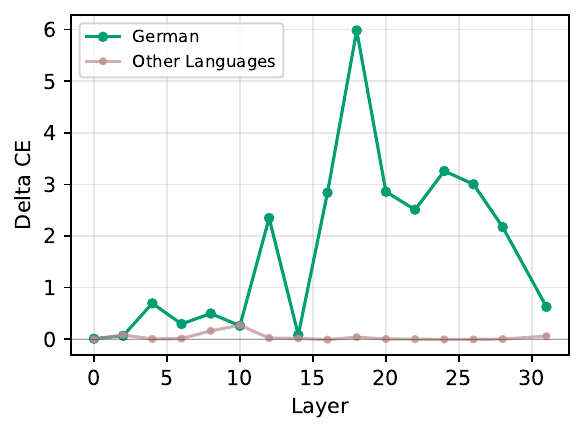}
    \caption{Ablate German features.}
  \end{subfigure}\hfill
  \begin{subfigure}[t]{0.32\linewidth}
    \centering
    \includegraphics[width=\linewidth]{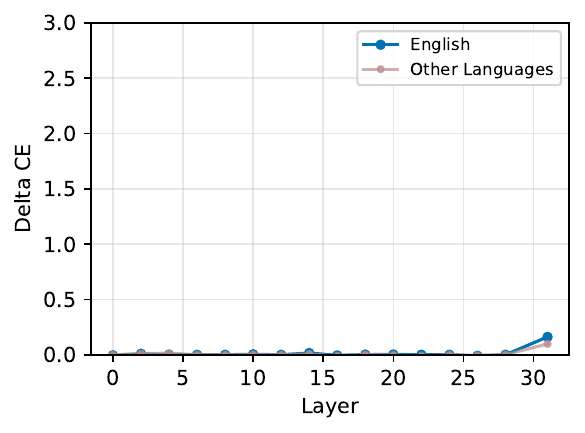}
    \caption{Ablate English features.}
  \end{subfigure}
  \caption{Change in cross-entropy after ablating the top-2 language-specific features per layer on Llama~3.1~8B. Ablation increases loss selectively for the target language. English is a notable exception.}
  \label{llama:directional_ablation}
\end{figure*}

The results reveal three key findings. First, ablating language-specific features substantially increases CE loss only for the target language, with negligible effects on other languages. This confirms that the features are causally important for language-specific processing. Second, changes in CE loss are generally larger in later layers, providing further evidence that language-specific information is concentrated in later model layers. Third, English is a notable exception: ablating its features produces near-zero change in CE across most layers, possibly because English-specific information is entangled with general concepts and thus diffused across many SAE features, so that the few features our method identifies have minimal impact when ablated. See Appendix~\ref{sec:appendix_directional_ablation_details} for additional results on other languages.

\section{Feature-Guided Language Steering}
\label{sec:steering_experiment}

\subsection{LangFIR-guided Steering}

Recall that $S_\text{spec}$ is the set of language-specific SAE feature indices identified in Section~\ref{sec:method}, and $\mathbf{f}_i^{\text{lang}} \in \mathbb{R}^{d_{\text{sae}}}$ is the SAE encoding of the $i$-th target-language sample.
We first compute the mean SAE activation across all $N$ target-language samples and select the top-$k$ most dominant features:
\begin{equation}
\bar{\mathbf{f}}^{\text{lang}} = \frac{1}{N}\sum_{i=1}^{N} \mathbf{f}_i^{\text{lang}}, \qquad
S_{\text{spec}}^{(k)} = \operatorname{TopK}\!\left(\left\{\bar{\mathbf{f}}^{\text{lang}}_j : j \in S_\text{spec}\right\}, k\right).
\end{equation}
We then form a sparse latent vector by masking all non-selected coordinates and decode it back into residual space:
\begin{equation}
\tilde{\mathbf{f}}^{\text{lang},(k)}_j =
\begin{cases}
\bar{\mathbf{f}}^{\text{lang}}_j, & j \in S_{\text{spec}}^{(k)}, \\
0, & \text{otherwise,}
\end{cases}
\qquad
\mathbf{v}_{\text{lang}}^{(k)} = W_{\mathrm{dec}}\,\tilde{\mathbf{f}}^{\text{lang},(k)} = \sum_{j \in S_{\text{spec}}^{(k)}} \bar{\mathbf{f}}^{\text{lang}}_j\,\mathbf{d}_j.
\end{equation}
The resulting steering vector is $\ell_2$-normalized and added to the residual stream at a chosen layer with scaling coefficient $\alpha$.

\subsection{Experimental Setup}
\label{sec:multilingual_experiment_setup}

\textbf{Task.}
Following~\citet{kojima2024multilingual}, we adopt the multilingual generation control task. The model receives a prompt of the form ``\texttt{Translate an English sentence into a target language. English: \{source\_text\} Target language:}'' without specifying the target language, isolating the steering method's ability to control output language.

\textbf{Models.}
We evaluate three open-weight LLMs for which pretrained SAEs covering all layers are available: Gemma 3 1B and Gemma 3 4B~\citep{Kamath2025Gemma3T} using Gemma Scope 2 SAEs~\citep{lieberum-etal-2024-gemma, mcdougall2025gemmascope2}, and Llama 3.1 8B~\citep{grattafiori2024llama} using Llama Scope SAEs~\citep{he2024llama}.

\textbf{Data and languages.}
We evaluate twelve English-to-X directions across three multilingual datasets: FLORES+~\citep{goyal2022flores, costa2022no}, WikiMatrix~\citep{schwenk-etal-2021-wikimatrix}, and Tatoeba~\citep{tiedemann-2020-tatoeba}. The target languages span twelve languages from multiple language families: Spanish (es), French (fr), German (de), Italian (it), Indonesian (id), Portuguese (pt), Vietnamese (vi), Polish (pl), Czech (cs), Turkish (tr), Chinese (zh), and Dutch (nl). Data is partitioned into three disjoint sets for steering vector construction, hyperparameter tuning, and final evaluation. More details are provided in Appendix~\ref{sec:appendix_datasets}.

\textbf{Evaluation metrics.}
We use three metrics. \textbf{Accuracy (ACC)} is the fraction of outputs in the target language as determined by fastText~\citep{joulin2017bag} with confidence threshold 0.5. \textbf{BLEU} measures translation quality on successful samples only. \textbf{ACC $\times$ BLEU} is a composite score reflecting both success rate and quality.

\textbf{Baselines.}
We compare against seven baselines. Five use parallel multilingual sentence pairs: LDA~\citep{balakrishnama1998linear}, Zhong's Parallel~\citep{zhong2025language}, SAE-DiffMean~\citep{gurgurov2026clas}, Gated-DiffMean~\citep{deng-etal-2025-unveiling}, and DiffMean~\citep{rimsky2024steering, marks2023geometry}. Two are monolingual: PCA~\citep{abdi2010principal} and Zhong's Monolingual. All methods produce a steering vector that is normalized and added to the residual stream at a chosen layer with a scaling coefficient $\alpha$ controlling the intervention strength. We also report two non-steering references: No Steering and Prompt Only. Details of all methods and references are provided in Appendix~\ref{sec:appendix_baselines}.

\subsection{Results}

\begin{table*}[t]
\centering
\caption{ACC$\times$BLEU averaged over twelve target languages and three datasets (FLORES+, WikiMatrix, Tatoeba). The best result among steering methods within each model is \textbf{bolded}.}
\label{tab:main_avg_accbleu}
\small
\setlength{\tabcolsep}{4pt}
\resizebox{\textwidth}{!}{%
\begin{tabular}{l cc @{\hspace{1.4em}} cccccccc}
\toprule
& \multicolumn{2}{c}{\textbf{Non-steering}} & \multicolumn{8}{c}{\textbf{Steering}} \\
\cmidrule(lr){2-3} \cmidrule(lr){4-11}
\textbf{Model} & No Steering & Prompt Only & PCA & LDA & Zhong (M) & Zhong (P) & SAE-DM & Gated-DM & DiffMean & LangFIR \\
\midrule
Gemma 3 1B   & 0.8 & 22.2 & 1.6 & 2.0 & 3.3 & 9.5 & 14.7 & 9.0 & 15.3 & \textbf{15.5} \\
Gemma 3 4B   & 2.6 & 35.5 & 3.3 & 15.2 & 10.5 & 13.9 & 23.2 & 10.8 & 23.9 & \textbf{26.6} \\
Llama 3.1 8B & 1.5 & 34.7 & 2.9 & 22.7 & 6.6 & 7.6 & 4.7 & 22.6 & 22.2 & \textbf{23.4} \\
\midrule
\textbf{Avg} & 1.6 & 30.8 & 2.6 & 13.3 & 6.8 & 10.3 & 14.2 & 14.1 & 20.5 & \textbf{21.8} \\
\bottomrule
\end{tabular}%
}
\end{table*}

We focus on ACC$\times$BLEU because a method is only useful if it succeeds on both axes at once: producing the target language (ACC) while preserving translation quality (BLEU). Table~\ref{tab:main_avg_accbleu} reports average ACC$\times$BLEU over twelve languages and three datasets. Per-language scores and the disaggregated ACC and BLEU tables are in Appendix~\ref{sec:appendix_steering_results}. Among steering methods, LangFIR achieves the highest average ACC$\times$BLEU on all three models (15.5, 26.6, and 23.4 on Gemma 3 1B, Gemma 3 4B, and Llama 3.1 8B, respectively), outperforming the next-best method, DiffMean (15.3, 23.9, and 22.2). The gain is largest on Gemma 3 4B, where LangFIR leads DiffMean by 2.7 points.

\textbf{Monolingual advantage.}
Despite requiring only monolingual target-language data, LangFIR slightly outperforms the strongest multilingual baseline (DiffMean) on average across all three models. Among the other monolingual methods (PCA, Zhong's Monolingual), none exceeds 10.5 average ACC$\times$BLEU on any model, while LangFIR achieves 15.5--26.6. On Gemma 3 1B, LangFIR surpasses the strongest monolingual baseline (Zhong's Monolingual) by $4.7\times$. LangFIR's effectiveness likely stems from SAE features being more disentangled by construction, which gives it a cleaner representational basis for isolating language-specific directions.

\textbf{Robustness across datasets.}
Per-dataset results show that LangFIR maintains its advantage across FLORES+, WikiMatrix, and Tatoeba (Tables~\ref{tab:accbleu_flores}--\ref{tab:accbleu_tatoeba}). The occasional weaker performance on certain languages can be attributed to the quality of the pretrained SAE and to hyperparameter tuning on a small subset of four languages, which may not optimally transfer to all twelve target languages.

\textbf{DiffMean-family baselines.}
Consistent with~\citet{gurgurov2026clas}, DiffMean on residual activations remains a strong baseline, outperforming both LDA and SAE-DiffMean on Gemma models. Gated-DiffMean~\citep{deng-etal-2025-unveiling} is competitive on Llama 3.1 8B but significantly weaker on the Gemma models.

\textbf{Prompt Only.}
Prompt Only attains the highest ACC$\times$BLEU on this task (22.2, 35.5, and 34.7). However, this advantage does not generalize: on the cross-lingual answer generation task (Section~\ref{sec:xquad}), Prompt Only underperforms the steering methods. Moreover, Prompt Only and steering operate at fundamentally different levels: the former relies on the model interpreting a language directive from the prompt, while the latter acts directly on residual-stream activations regardless of instruction-following behavior. We therefore treat Prompt Only as an informative reference rather than a competing baseline.

\subsection{Ablation Study}
\label{sec:ablation}

We ablate the following aspects of LangFIR on Llama~3.1~8B: number of top-$k$ features, sample size, feature-overlap removal, intervention layer, scaling coefficient, frequency threshold, random-token sequence length, random seed, and filter negative set. Each factor is varied independently while holding all others at their validation-selected values. Results on Gemma models are generally consistent. We highlight the most important findings below and report all setup details, additional ablations, and Gemma results in Appendix~\ref{sec:appendix_ablation_details}.

\begin{figure*}[t]
  \centering
  \begin{subfigure}[t]{0.32\textwidth}
    \centering
    \includegraphics[width=\linewidth]{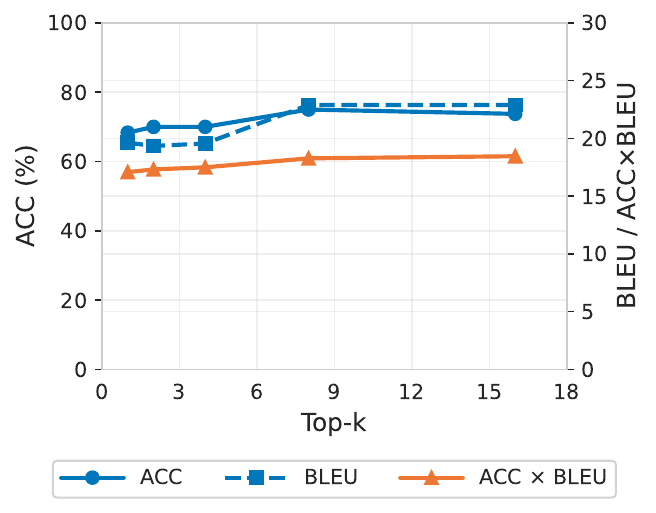}
    \caption{Top-$k$ features.}
    \label{llama:ablation_topk}
  \end{subfigure}
  \hfill
  \begin{subfigure}[t]{0.32\textwidth}
    \centering
    \includegraphics[width=\linewidth]{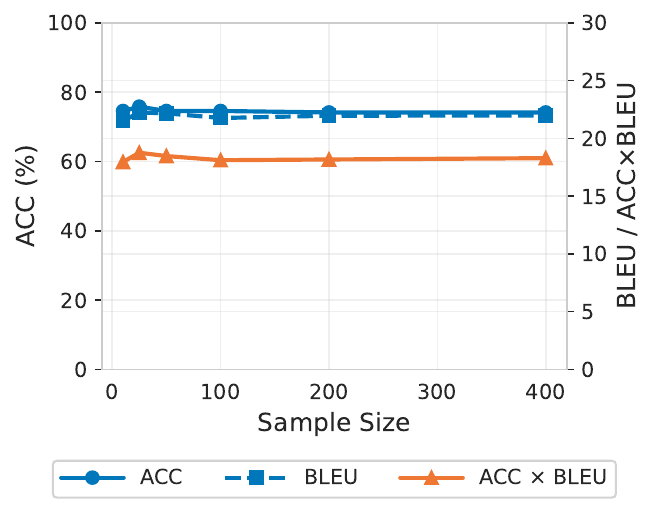}
    \caption{Sample size $N$.}
    \label{llama:ablation_samplesize}
  \end{subfigure}
  \hfill
  \begin{subfigure}[t]{0.32\textwidth}
    \centering
    \includegraphics[width=\linewidth]{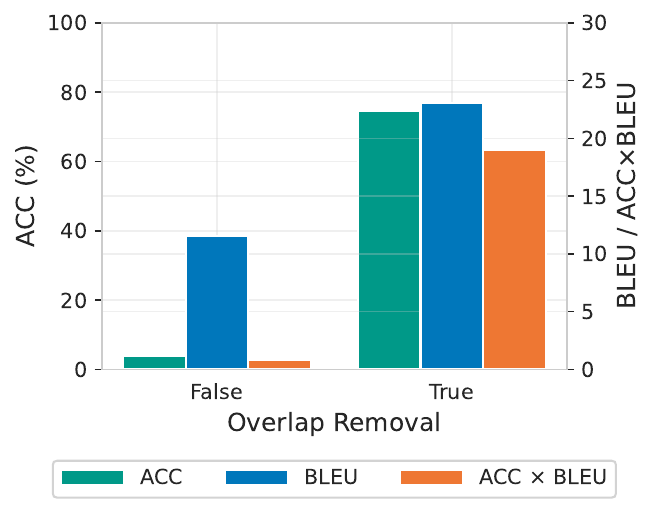}
    \caption{Feature overlap removal.}
    \label{llama:ablation_ignore_shared}
  \end{subfigure}
  \caption{Ablation of number of top-$k$ language-specific features, sample size, and feature-overlap removal on Llama~3.1~8B. (a)~Even $k{=}1$ achieves strong performance. (b)~Steering metrics are stable with as few as 10 sentences. (c)~Removing feature-overlap removal degrades all steering metrics dramatically.}
  \label{llama:ablation_topk_sample_shared}
\end{figure*}

\textbf{Number of features.}
Figure~\ref{llama:ablation_topk} shows that even at $k = 1$, all three metrics are already high, consistent with the top-feature dominance observed in Section~\ref{sec:analyses}. Including additional language-specific features yields modest further improvement.

\textbf{Sample size.}
Figure~\ref{llama:ablation_samplesize} shows that all three metrics remain stable across sample sizes. LangFIR achieves comparable performance with as few as 10 monolingual sentences, demonstrating that the method is extremely data-efficient.

\textbf{Feature overlap removal.}
The feature-overlap-removal step is essential to LangFIR. Omitting it (using $S_\text{lang}$ directly instead of $S_\text{spec} = S_\text{lang} \setminus S_\text{rand}$) degrades ACC by approximately $18\times$, BLEU by $2\times$, and ACC$\times$BLEU by $19\times$, even after retuning hyperparameters (Figure~\ref{llama:ablation_ignore_shared}). Without this step, the steering vector is dominated by features unrelated to language identity.

\subsection{Cross-Lingual Answer Generation}
\label{sec:xquad}

\begin{table*}[t]
\centering
\caption{Cross-lingual answer generation results: F1 averaged over five target languages (es, de, vi, tr, zh). The best result among steering methods within each model is \textbf{bolded}.}
\label{tab:xquad_f1}
\small
\setlength{\tabcolsep}{4pt}
\resizebox{\textwidth}{!}{%
\begin{tabular}{l cc @{\hspace{1.4em}} cccccccc}
\toprule
& \multicolumn{2}{c}{\textbf{Non-steering}} & \multicolumn{8}{c}{\textbf{Steering}} \\
\cmidrule(lr){2-3} \cmidrule(lr){4-11}
\textbf{Model} & No Steering & Prompt Only & PCA & LDA & Zhong (M) & Zhong (P) & SAE-DM & Gated-DM & DiffMean & LangFIR \\
\midrule
Gemma 3 1B   & 12.7 & 12.3 & 10.6 & 12.7 & 11.1 & 15.1 & 13.3 & 13.6 & 14.4 & \textbf{16.7} \\
Gemma 3 4B   & 14.6 & 7.9 & 15.6 & 14.7 & 11.3 & 13.3 & \textbf{22.5} & 19.3 & 21.7 & 20.0 \\
Llama 3.1 8B & 12.3 & 12.2 & 11.6 & 11.8 & 10.6 & 11.2 & 11.1 & 20.8 & 20.7 & \textbf{21.6} \\
\midrule
\textbf{Avg} & 13.2 & 10.8 & 12.6 & 13.0 & 11.0 & 13.2 & 15.6 & 17.9 & 18.9 & \textbf{19.4} \\
\bottomrule
\end{tabular}%
}
\end{table*}

To test whether LangFIR generalizes beyond sentence translation, we evaluate it on an additional cross-lingual answer generation task using XQuAD~\citep{artetxe2020cross}, covering the five languages that are both in the multilingual generation control experiment and available in XQuAD (es, de, vi, tr, zh). Given an English context and question, the model must generate the answer in the target language. The setup follows our multilingual generation control task, with steering vectors constructed from FLORES+. The prompt is ``\texttt{Read the following passage and answer the question [in \{target language\}]. Context: ... Question: ... Answer:}'', where the bracketed clause is used only for Prompt Only. We focus on F1, since exact match (EM) may penalize valid paraphrases. The full setup, EM, and per-language results are in Appendix~\ref{sec:appendix_xquad}.

On cross-lingual answer generation, Table~\ref{tab:xquad_f1} shows that the picture inverts relative to sentence translation: all steering methods beat Prompt Only on average (11.0--19.4 vs.\ 10.8). Among steering methods, LangFIR is the strongest on average despite being monolingual (19.4), slightly ahead of DiffMean (18.9). We speculate that Prompt Only performs poorly here because the large English context creates stronger English-generation momentum than the sentence-translation task does, a momentum that Prompt Only does not reliably overcome. Inspecting Prompt Only's generated outputs confirms this: it almost always generates in English rather than the target language. Activation-level steering methods, by operating directly on the residual stream rather than relying on the model to interpret a language directive mid-context, appear better suited to overcoming this momentum.

\section{Conclusion}

We introduced LangFIR, a method for identifying sparse language-specific SAE features using minimal monolingual data and random-token sequences. The key insight is that random-token sequences serve as a simple proxy for surfacing language-agnostic features, and removing them from the set of language-consistent features isolates a clean, sparse set of features that encode language identity.
Our analyses confirm that the identified features are language-selective, causally meaningful under directional ablation, and concentrated in later model layers. Steering vectors built from LangFIR features consistently outperform strong residual-stream baselines across three models, three datasets, and twelve target languages on the multilingual generation control task, and the advantage carries over to cross-lingual answer generation on XQuAD.

These results show that language steering can be achieved without parallel or multilingual data, enabling steering in settings where such data is unavailable, and that language identity in multilingual LLMs is concentrated in a small number of sparse, interpretable, and manipulable feature directions.
Future work could extend the underlying framework, contrasting a target-concept corpus against a concept-agnostic proxy, to non-language concepts. A deeper analysis of how language-specific features organize across language families and model layers could further illuminate how language information is structured in multilingual models.

\section*{Limitations}

Our method depends on the availability and quality of pretrained SAEs, which restricts our evaluation to three models, all under 10B parameters and from two English-centric model families (Gemma and Llama). The languages we evaluate are also primarily mid- to high-resource. Whether our findings hold for larger models, for models pretrained with a more balanced language mix, or for low-resource languages remains open.


\section*{Acknowledgments}
This work was supported in part by the National Science Foundation (NSF) under Grant IIS-2401685 and by the Humanities and AI Virtual Institute (HAVI) program of Schmidt Sciences.
This work used the Delta system at the National Center for Supercomputing Applications [award OAC 2005572] through allocation CIS250532 from the Advanced Cyberinfrastructure Coordination Ecosystem: Services \& Support (ACCESS) program~\citep{10.1145/3569951.3597559}, which is supported by National Science Foundation grants \#2138259, \#2138286, \#2138307, \#2137603, and \#2138296.
We also thank the National Deep Inference Fabric (NDIF)~\citep{fiotto-kaufman2025nnsight} for providing remote inference access through its pilot program.


\section*{LLM Usage Disclosure}
We disclose that LLMs were used for editing and polishing the language of this manuscript. All scientific content, experimental design, and analysis were conducted by the authors.

\bibliography{references}
\bibliographystyle{colm2026_conference}

\clearpage
\appendix
\section{Experimental Details}
\label{sec:appendix_experimental_details}

\subsection{Models and SAEs}
\label{sec:appendix_models_saes}

We experiment with three pretrained language models paired with publicly available SAEs that cover all residual-stream layers. All SAEs are loaded from the SAELens library~\citep{bloom2024saetrainingcodebase}. All model interventions are implemented using NNsight~\citep{fiotto-kaufman2025nnsight}. Table~\ref{tab:models_saes} summarizes the configurations.

\begin{table}[h]
\centering
\caption{Models and SAEs used in our experiments.}
\label{tab:models_saes}
\resizebox{\linewidth}{!}{%
\begin{tabular}{lll}
    \toprule
    \textbf{Model ID} & \textbf{SAE Release} & \textbf{SAE ID} \\
    \midrule
    \texttt{google/gemma-3-1b-pt} & \texttt{gemma-scope-2-1b-pt-res-all} & \texttt{layer\_\{layer\}\_width\_16k\_l0\_small} \\
    \texttt{google/gemma-3-4b-pt} & \texttt{gemma-scope-2-4b-pt-res-all} & \texttt{layer\_\{layer\}\_width\_16k\_l0\_small} \\
    \texttt{meta-llama/Llama-3.1-8B} & \texttt{llama\_scope\_lxr\_8x} & l\{layer\}r\_8x\\
    \bottomrule
\end{tabular}%
}
\end{table}

\subsection{Datasets and Preprocessing}
\label{sec:appendix_datasets}

We evaluate on three multilingual datasets. \textbf{FLORES+}~\citep{goyal2022flores, costa2022no} is a high-quality parallel translation dataset covering 200+ languages. \textbf{WikiMatrix}~\citep{schwenk-etal-2021-wikimatrix} is a Wikipedia-mined corpus providing aligned sentence pairs across 1,600+ language pairs. \textbf{Tatoeba}~\citep{tiedemann-2020-tatoeba} is a community-contributed multi-parallel corpus of short translated sentences spanning 400+ languages.

For Tatoeba, the validation split is used for vector construction and the test split for evaluation. For WikiMatrix, the train split is used for vector construction and the dev split for evaluation. For FLORES+, the dev split is used for construction and disjoint indices from the devtest split are used for evaluation.

Sentences are filtered by length (30 to 300 characters). 
For each language, 100 parallel sentence pairs are used to construct steering vectors, 20 samples form a validation set for hyperparameter tuning, and 100 test samples are used for final evaluation. 
All methods use the same splits.

\subsection{Steering Setup}
\label{sec:appendix_steering_setup}

All steering interventions are applied at the last token position of each generation step. We use greedy decoding (no sampling) for all experiments.

\subsection{Baselines and Non-Steering References}
\label{sec:appendix_baselines}

All addition-based methods normalize their steering vectors and add them to the residual stream with a scaling factor $\alpha$. Hyperparameter tuning details are provided in Appendix~\ref{sec:appendix_hyperparams}.

\textbf{PCA.} Applies Principal Component Analysis~\citep{abdi2010principal} to residual activations from the target language, projecting hidden states onto a low-dimensional principal subspace.

\textbf{LDA.} Uses Linear Discriminant Analysis~\citep{balakrishnama1998linear} to find a direction maximizing separation between target and non-target language representations.

\textbf{Zhong's Monolingual.} Identifies language-specific dimensions by comparing target-language representations between an intermediate and the final layer, then overwrites these dimensions during inference~\citep{zhong2025language}.

\textbf{Zhong's Parallel.} Identifies language-specific dimensions by comparing parallel English/target sentence pairs at the final layer, then overwrites these dimensions during inference~\citep{zhong2025language}.

\textbf{SAE-DiffMean.} Extends DiffMean to the SAE latent space, computing the steering direction as the difference between mean latent activations of the target and source languages, then decoding back to model space.

\textbf{Gated-DiffMean.} \citet{deng-etal-2025-unveiling} extend DiffMean with conditional application: the steering vector is added only at generation steps where the source language's top-2 language-specific features are active. The original method intervenes on three layers. For fair comparison, we apply it at a single layer.

\textbf{DiffMean.} Computes a steering direction as the difference between average residual-stream activations of the target and source languages~\citep{rimsky2024steering, marks2023geometry}. This is the strongest baseline on CLaS-Bench.

We also report two non-steering references that apply no activation-level intervention. \textbf{No Steering} runs the base model on the task prompt as is. \textbf{Prompt Only} augments the task prompt with an explicit instruction to generate in the target language.

\subsection{Hyperparameter Search Space}
\label{sec:appendix_hyperparams}

Hyperparameters are tuned as fairly as possible across methods using a subset of four languages (Chinese, Spanish, French, and German) from the twelve evaluated. Per dataset and model combination, a validation run is conducted. For all addition-based steering methods (PCA, LDA, SAE-DiffMean, Gated-DiffMean, DiffMean, LangFIR), we sweep over the same set of late layers, motivated by the finding that language-specific structure emerges predominantly in later layers~\citep{gurgurov2026clas}, and a model-specific set of intervention scaling coefficients $\alpha$. For Zhong's replacement-based methods, we use the hyperparameter search space recommended by the original paper~\citep{zhong2025language}. We select hyperparameters based on ACC$\times$BLEU averaged across the four languages on a 20-sample validation set. Table~\ref{tab:hyperparams} lists the full search space for each method.

\begin{table}[h]
\centering
\caption{Hyperparameter search space. Tuning is conducted on Chinese, Spanish, French, and German for every dataset and model combination.}
\label{tab:hyperparams}
\begin{tabular}{ll}
    \toprule
    \textbf{Hyperparameter} & \textbf{Values} \\
    \midrule
    Intervention layer percentile & 0.4, 0.5, 0.6, 0.7, 0.8, 0.9, 1.0 \\
    Steering scalar $\alpha$ (Zhong Mono/Para) & 0.2, 0.6, 1.0, 1.4, 1.8, 2.0 \\
    Steering scalar $\alpha$ (Other methods, Llama) & 0.1, 0.5, 1, 2.5, 5, 10 \\
    Steering scalar $\alpha$ (Other methods, Gemma) & 0.1, 1, 10, 100, 1000, 10000, 100000 \\
    PCA top-$k$ components & 1, 4, 8, 16, 32 \\
    Zhong top-$k$ dimensions & 200, 400, 600 \\
    Zhong anchor layer percentile & 0.375, 0.5, 0.625 \\
    LangFIR activation threshold $\tau$ & 0.8, 0.9, 1.0 \\
    LangFIR top-$k$ SAE features & 1, 2, 100 (effectively selects all features) \\
    \bottomrule
\end{tabular}%
\end{table}

\FloatBarrier
\subsection{Validation-Tuned Hyperparameters}
\label{sec:appendix_tuned_hyperparams}

Tables~\ref{tab:tuned_hp_gemma1b}--\ref{tab:tuned_hp_llama} report the hyperparameters selected by validation for each method, dataset, and model.

\begin{table}[h]
\centering
\caption{Selected hyperparameters for Gemma-3-1B.}
\label{tab:tuned_hp_gemma1b}
\begin{tabular}{llp{6cm}}
\toprule
\textbf{Dataset} & \textbf{Method} & \textbf{Selected hyperparameters} \\
\midrule
\multirow{8}{*}{FLORES+}
  & PCA            & layer 25, $\alpha$=1000, top-$k$=1 \\
  & LDA            & layer 22, $\alpha$=1000 \\
  & Zhong (Mono)   & layer 22, $\alpha$=0.6, top-$k$=200, anchor=15 \\
  & Zhong (Para)   & layer 20, $\alpha$=0.6, top-$k$=200 \\
  & SAE-DM         & layer 17, $\alpha$=1000 \\
  & Gated-DM       & layer 10, $\alpha$=1000 \\
  & DiffMean       & layer 17, $\alpha$=1000 \\
  & LangFIR        & layer 17, $\alpha$=1000, $\tau$=0.8, top-$k$=100 \\
\midrule
\multirow{8}{*}{WikiMatrix}
  & PCA            & layer 20, $\alpha$=10000, top-$k$=16 \\
  & LDA            & layer 17, $\alpha$=1000 \\
  & Zhong (Mono)   & layer 20, $\alpha$=0.6, top-$k$=200, anchor=9 \\
  & Zhong (Para)   & layer 20, $\alpha$=0.6, top-$k$=200 \\
  & SAE-DM         & layer 17, $\alpha$=1000 \\
  & Gated-DM       & layer 10, $\alpha$=1000 \\
  & DiffMean       & layer 17, $\alpha$=1000 \\
  & LangFIR        & layer 17, $\alpha$=1000, $\tau$=0.9, top-$k$=100 \\
\midrule
\multirow{8}{*}{Tatoeba}
  & PCA            & layer 22, $\alpha$=10000, top-$k$=8 \\
  & LDA            & layer 12, $\alpha$=100 \\
  & Zhong (Mono)   & layer 22, $\alpha$=0.6, top-$k$=200, anchor=9 \\
  & Zhong (Para)   & layer 20, $\alpha$=0.6, top-$k$=200 \\
  & SAE-DM         & layer 17, $\alpha$=1000 \\
  & Gated-DM       & layer 10, $\alpha$=1000 \\
  & DiffMean       & layer 17, $\alpha$=1000 \\
  & LangFIR        & layer 17, $\alpha$=1000, $\tau$=0.8, top-$k$=1 \\
\bottomrule
\end{tabular}
\end{table}

\begin{table}[h]
\centering
\caption{Selected hyperparameters for Gemma-3-4B.}
\label{tab:tuned_hp_gemma4b}
\begin{tabular}{llp{6cm}}
\toprule
\textbf{Dataset} & \textbf{Method} & \textbf{Selected hyperparameters} \\
\midrule
\multirow{8}{*}{FLORES+}
  & PCA            & layer 29, $\alpha$=100000, top-$k$=16 \\
  & LDA            & layer 19, $\alpha$=1000 \\
  & Zhong (Mono)   & layer 29, $\alpha$=0.6, top-$k$=200, anchor=12 \\
  & Zhong (Para)   & layer 29, $\alpha$=0.6, top-$k$=200 \\
  & SAE-DM         & layer 29, $\alpha$=10000 \\
  & Gated-DM       & layer 13, $\alpha$=10000 \\
  & DiffMean       & layer 23, $\alpha$=10000 \\
  & LangFIR        & layer 19, $\alpha$=1000, $\tau$=0.8, top-$k$=100 \\
\midrule
\multirow{8}{*}{WikiMatrix}
  & PCA            & layer 29, $\alpha$=100000, top-$k$=8 \\
  & LDA            & layer 19, $\alpha$=1000 \\
  & Zhong (Mono)   & layer 29, $\alpha$=1.0, top-$k$=200, anchor=12 \\
  & Zhong (Para)   & layer 26, $\alpha$=0.6, top-$k$=200 \\
  & SAE-DM         & layer 23, $\alpha$=10000 \\
  & Gated-DM       & layer 16, $\alpha$=10000 \\
  & DiffMean       & layer 23, $\alpha$=10000 \\
  & LangFIR        & layer 19, $\alpha$=1000, $\tau$=0.8, top-$k$=2 \\
\midrule
\multirow{8}{*}{Tatoeba}
  & PCA            & layer 26, $\alpha$=10000, top-$k$=16 \\
  & LDA            & layer 13, $\alpha$=100 \\
  & Zhong (Mono)   & layer 29, $\alpha$=1.0, top-$k$=200, anchor=16 \\
  & Zhong (Para)   & layer 29, $\alpha$=1.0, top-$k$=200 \\
  & SAE-DM         & layer 13, $\alpha$=10000 \\
  & Gated-DM       & layer 16, $\alpha$=10000 \\
  & DiffMean       & layer 23, $\alpha$=10000 \\
  & LangFIR        & layer 19, $\alpha$=1000, $\tau$=0.9, top-$k$=2 \\
\bottomrule
\end{tabular}
\end{table}

\begin{table}[h]
\centering
\caption{Selected hyperparameters for Llama-3.1-8B.}
\label{tab:tuned_hp_llama}
\begin{tabular}{llp{6cm}}
\toprule
\textbf{Dataset} & \textbf{Method} & \textbf{Selected hyperparameters} \\
\midrule
\multirow{8}{*}{FLORES+}
  & PCA            & layer 31, $\alpha$=10, top-$k$=1 \\
  & LDA            & layer 15, $\alpha$=2.5 \\
  & Zhong (Mono)   & layer 31, $\alpha$=1.0, top-$k$=200, anchor=15 \\
  & Zhong (Para)   & layer 31, $\alpha$=1.0, top-$k$=200 \\
  & SAE-DM         & layer 24, $\alpha$=5 \\
  & Gated-DM       & layer 21, $\alpha$=10 \\
  & DiffMean       & layer 18, $\alpha$=10 \\
  & LangFIR        & layer 18, $\alpha$=5, $\tau$=0.8, top-$k$=100 \\
\midrule
\multirow{8}{*}{WikiMatrix}
  & PCA            & layer 31, $\alpha$=10, top-$k$=16 \\
  & LDA            & layer 12, $\alpha$=2.5 \\
  & Zhong (Mono)   & layer 27, $\alpha$=0.2, top-$k$=400, anchor=19 \\
  & Zhong (Para)   & layer 21, $\alpha$=0.2, top-$k$=400 \\
  & SAE-DM         & layer 31, $\alpha$=10 \\
  & Gated-DM       & layer 21, $\alpha$=10 \\
  & DiffMean       & layer 18, $\alpha$=10 \\
  & LangFIR        & layer 18, $\alpha$=5, $\tau$=0.8, top-$k$=1 \\
\midrule
\multirow{8}{*}{Tatoeba}
  & PCA            & layer 31, $\alpha$=10, top-$k$=16 \\
  & LDA            & layer 15, $\alpha$=2.5 \\
  & Zhong (Mono)   & layer 21, $\alpha$=0.2, top-$k$=200, anchor=19 \\
  & Zhong (Para)   & layer 21, $\alpha$=0.2, top-$k$=200 \\
  & SAE-DM         & layer 31, $\alpha$=10 \\
  & Gated-DM       & layer 24, $\alpha$=10 \\
  & DiffMean       & layer 21, $\alpha$=10 \\
  & LangFIR        & layer 12, $\alpha$=2.5, $\tau$=1.0, top-$k$=1 \\
\bottomrule
\end{tabular}
\end{table}

\FloatBarrier
\section{Examples of Language-Agnostic Features}
\label{sec:appendix_noise_features}

\begin{figure}[ht!]
  \centering
  \begin{subfigure}[t]{\columnwidth}
    \centering
    \includegraphics[width=\linewidth]{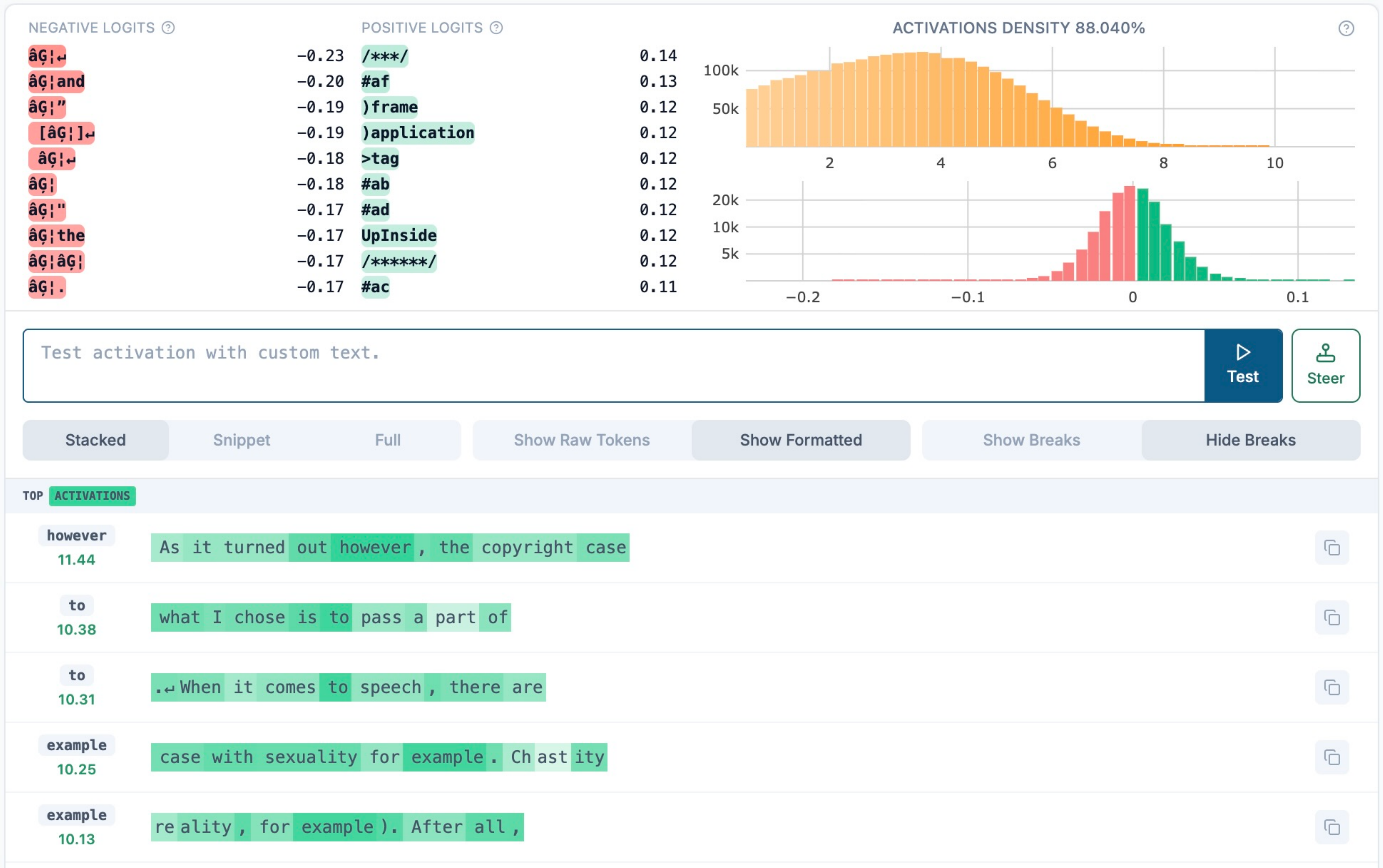}
    \caption{Feature 2618: activates near-indiscriminately at very high density (88.0\%).}
    \label{llama:noise_feature_2618}
  \end{subfigure}

  \vspace{1ex}
  \begin{subfigure}[t]{\columnwidth}
    \centering
    \includegraphics[width=\linewidth]{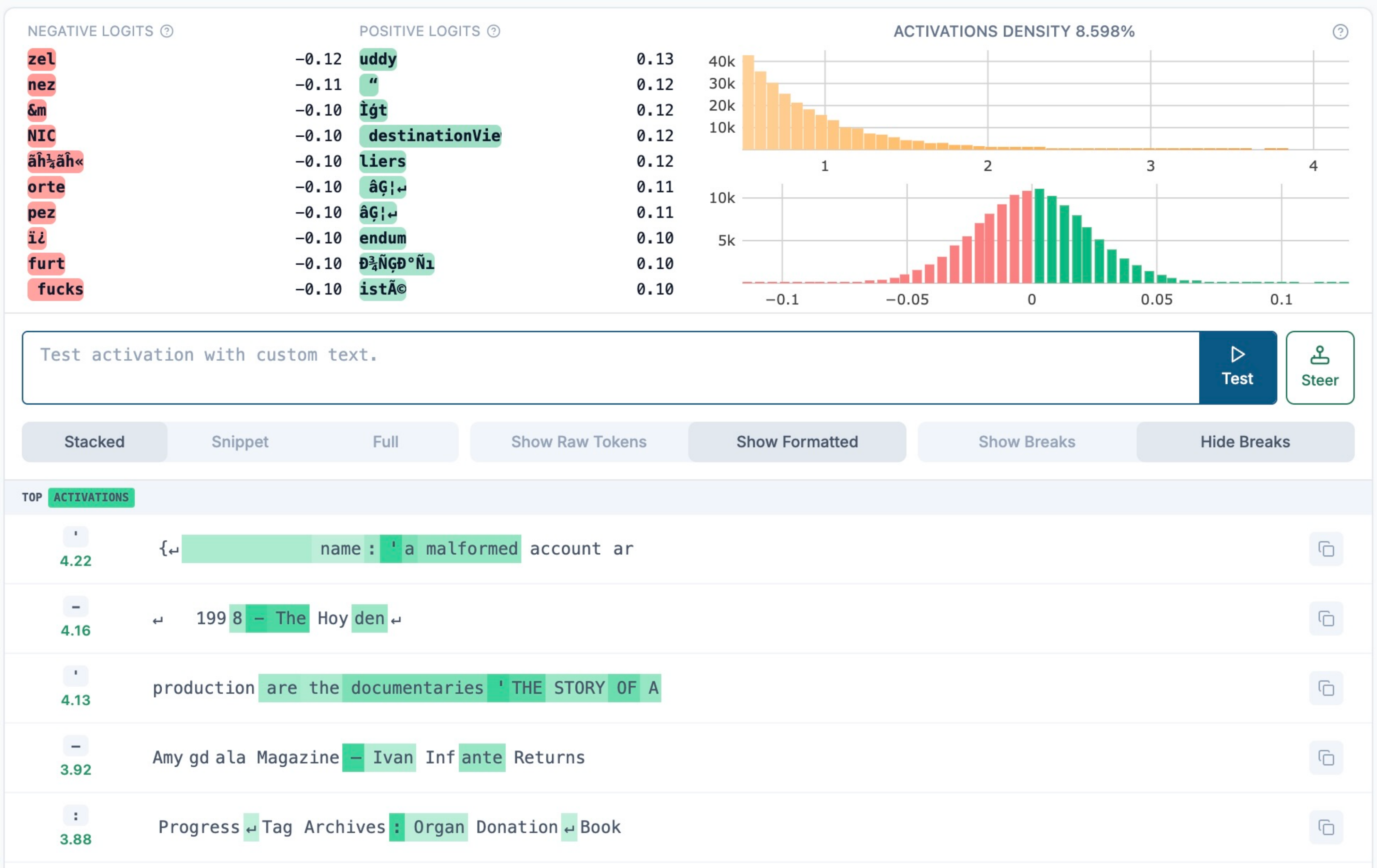}
    \caption{Feature 3157: activates on surface patterns including formatted headings and punctuations.}
    \label{llama:noise_feature_3157}
  \end{subfigure}
  \caption{Two language-agnostic features from Llama~3.1~8B at layer 18 using the \texttt{llamascope-res-32k} SAE. Panels show logit effects, activation density, and top activating examples from Neuronpedia~\citep{neuronpedia}. Neither increases target-language token logits, and both track generic textual structure rather than language identity. Feature pages: \href{https://www.neuronpedia.org/llama3.1-8b/18-llamascope-res-32k/2618}{2618}, \href{https://www.neuronpedia.org/llama3.1-8b/18-llamascope-res-32k/3157}{3157}.}
  \label{llama:noise_features}
\end{figure}

\FloatBarrier
\section{Analysis Details}
\label{sec:appendix_analysis_details}

\begin{figure}[ht!]
  \centering
  \includegraphics[width=0.49\columnwidth]{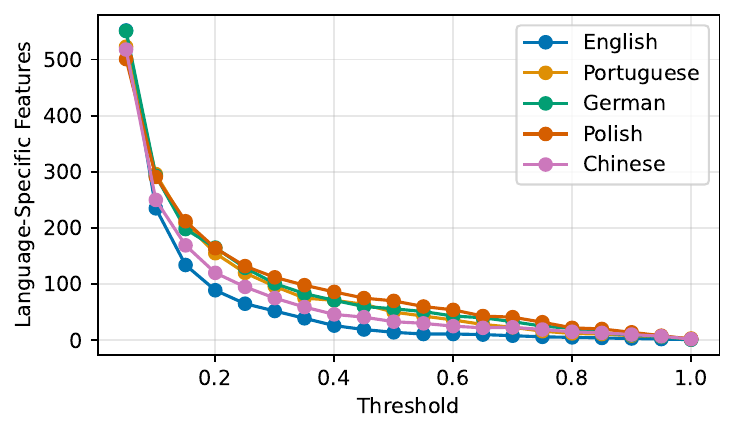}
  \caption{Language-specific feature count versus frequency threshold $\tau$. Feature counts stabilize for $\tau \geq 0.8$.}
  \label{llama:ls_features_vs_threshold}
\end{figure}

\begin{figure}[ht!]
  \centering
  \includegraphics[width=0.49\columnwidth]{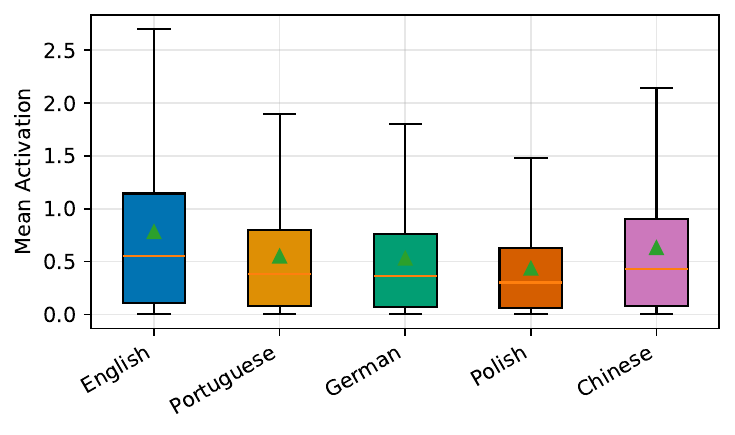}
  \caption{Non-language-specific feature activations across evaluation languages. These features activate broadly and uniformly across languages, with English slightly elevated.}
  \label{llama:non_ls_feature_activations}
\end{figure}

\begin{figure*}[ht!]
  \centering
  \begin{subfigure}[t]{0.49\linewidth}
    \centering
    \includegraphics[width=\linewidth]{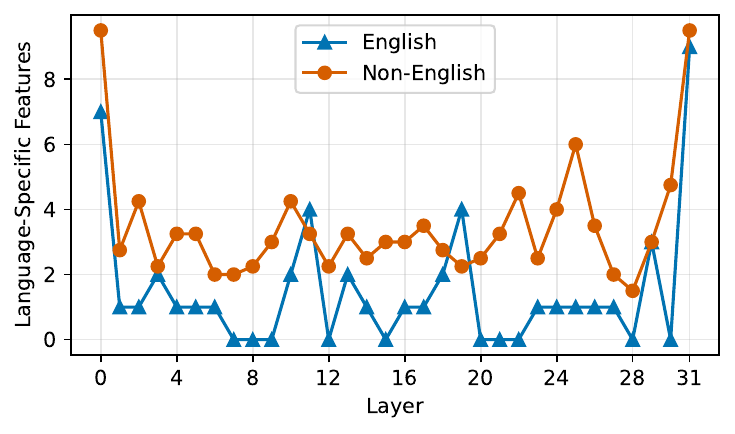}
    \caption{Feature count across layers.}
    \label{llama:ls_features_vs_layers}
  \end{subfigure}
  \hfill
  \begin{subfigure}[t]{0.49\linewidth}
    \centering
    \includegraphics[width=\linewidth]{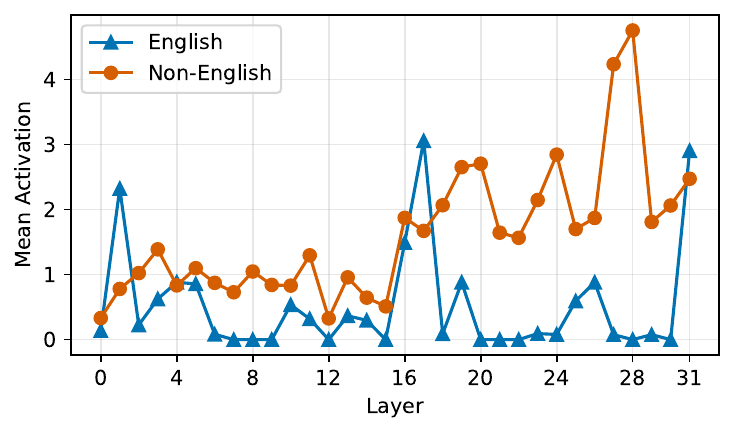}
    \caption{Mean feature activation across layers.}
    \label{llama:ls_feature_activations_vs_layers}
  \end{subfigure}
  \caption{Layer-wise analyses on Llama~3.1~8B. (a)~Feature density dips in middle layers and recovers in later layers. (b)~Activation magnitudes peak in later layers, indicating stronger language specialization near the output.}
  \label{llama:layerwise_counts_and_activations}
\end{figure*}

\begin{figure}[ht!]
  \centering
  \begin{subfigure}[t]{0.49\columnwidth}
    \centering
    \includegraphics[width=\linewidth]{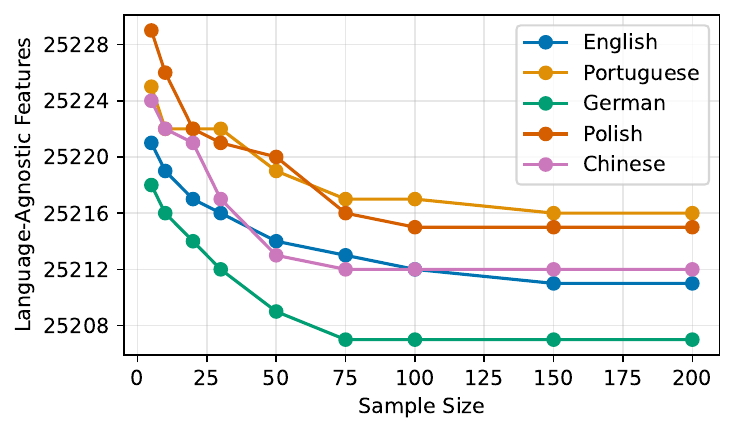}
    \caption{Language-agnostic feature count vs sample size.}
    \label{llama:la_features_vs_sentences}
  \end{subfigure}
  \hfill
  \begin{subfigure}[t]{0.49\columnwidth}
    \centering
    \includegraphics[width=\linewidth]{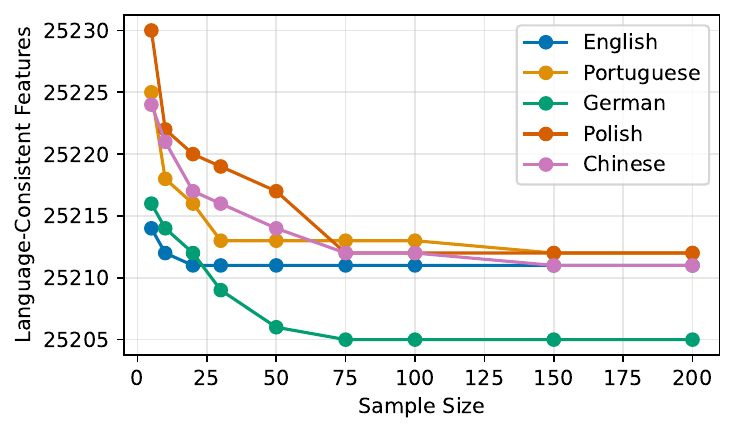}
    \caption{Language-consistent feature count vs sample size.}
    \label{llama:lc_features_vs_sentences}
  \end{subfigure}

  \caption{Convergence of intermediate feature sets (before feature overlap removal) with increasing sample size on Llama~3.1~8B. (a)~Language-agnostic features (identified from random-token sequences) and (b)~language-consistent features (identified from target-language sentences) both stabilize by ${\sim}$100 sentences, confirming that LangFIR's inputs to the filtering step are themselves stable with limited data.}
  \label{llama:la_lc_features_vs_sentences}
\end{figure}

\FloatBarrier
\section{Directional Ablation Details}
\label{sec:appendix_directional_ablation_details}

\begin{figure*}[h]
  \centering
  \begin{subfigure}[t]{0.32\textwidth}
    \centering
    \includegraphics[width=\linewidth]{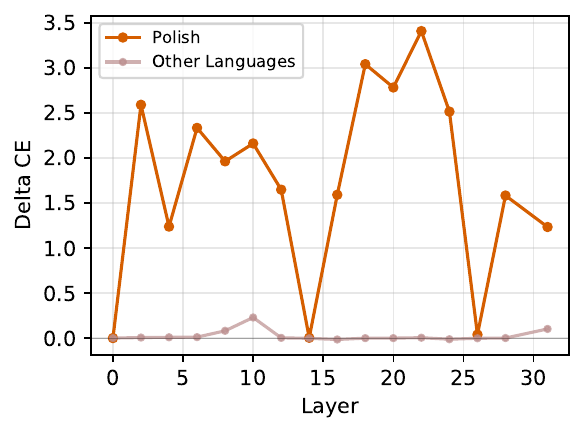}
    \caption{Ablate Polish features.}
    \label{fig:directional_ablation_pl}
  \end{subfigure}
  \quad
  \begin{subfigure}[t]{0.32\textwidth}
    \centering
    \includegraphics[width=\linewidth]{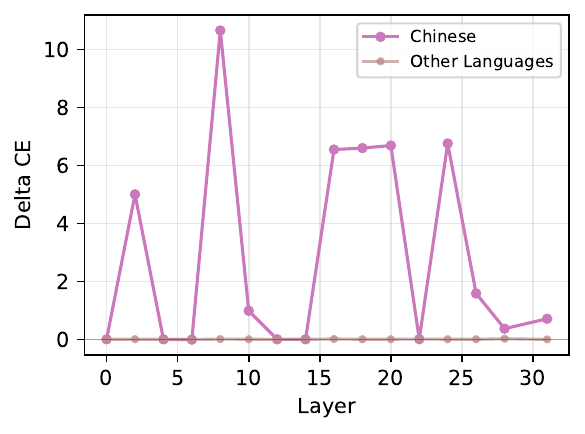}
    \caption{Ablate Chinese features.}
    \label{fig:directional_ablation_zh}
  \end{subfigure}
  \caption{Change in cross-entropy after ablating the top-2 language-specific features per layer on Llama~3.1~8B. Consistent with Figure~\ref{llama:directional_ablation}, ablation selectively increases loss for the target language.}
  \label{fig:directional_ablation_pl_zh}
\end{figure*}

\textbf{Additional results.} Figure~\ref{fig:directional_ablation_pl_zh} shows additional directional ablation results for Polish and Chinese on Llama~3.1~8B. Both languages exhibit similar patterns to Portuguese and German in Figure~\ref{llama:directional_ablation}, though the late-layer effect is less pronounced for Chinese.

\textbf{Setup.} We use 100 evaluation samples per language from the FLORES+ validation split, disjoint from the 100 samples used for feature identification. When multiple features are ablated, their projections are subtracted from the residual stream sequentially.

\FloatBarrier
\section{Results}
\label{sec:appendix_steering_results}

This appendix provides the full per-language steering results of the multilingual generation control experiment, organized by metric. For each metric we report an average over the three datasets, followed by per-dataset breakdowns.

\FloatBarrier
\subsection{\texorpdfstring{ACC$\times$BLEU}{ACC x BLEU}}
Table~\ref{tab:accbleu_alldatasets} gives the per-language ACC$\times$BLEU averaged over the three datasets, and Tables~\ref{tab:accbleu_flores}--\ref{tab:accbleu_tatoeba} give the per-dataset breakdowns. Per language, LangFIR achieves the best ACC$\times$BLEU among steering methods on 7, 10, and 8 of 12 languages on Gemma 3 1B, Gemma 3 4B, and Llama 3.1 8B, respectively.

\begin{table*}[t]
\centering
\caption{ACC$\times$BLEU averaged over FLORES+, WikiMatrix, and Tatoeba for Gemma~3~1B, Gemma~3~4B, Llama~3.1~8B. The best result among steering methods per language within each model is \textbf{bolded}.}
\small
\setlength{\tabcolsep}{3pt}
\resizebox{\textwidth}{!}{
\begin{tabular}{llrrrrrrrrrrrrr}
\toprule
\textbf{Model} & \textbf{Method} & es & fr & de & it & id & pt & vi & pl & cs & tr & zh & nl & \textbf{Avg} \\
\midrule
\multirow{10}{*}{Gemma 3 1B} & No Steering & 0.9 & 1.1 & 4.9 & 0.8 & 0.1 & 0.3 & 0.4 & 0.2 & 0.4 & 0.5 & 0.2 & 0.3 & 0.8 \\
 & Prompt Only & 20.3 & 23.7 & 23.8 & 27.0 & 29.1 & 36.8 & 25.6 & 14.7 & 15.9 & 7.8 & 15.5 & 26.4 & 22.2 \\
\cmidrule{2-15}
 & PCA & 1.0 & 1.5 & 8.3 & 1.3 & 0.4 & 0.2 & 0.6 & 1.7 & 2.1 & 0.1 & 0.6 & 0.9 & 1.6 \\
 & LDA & 3.2 & 5.0 & 8.3 & 1.8 & 1.1 & 1.7 & 0.3 & 0.2 & 0.9 & 0.7 & 0.1 & 1.1 & 2.0 \\
 & Zhong (Mono) & 2.3 & 4.2 & 6.5 & 3.4 & 1.8 & 6.5 & 4.5 & 0.7 & 2.1 & 0.9 & 2.4 & 4.2 & 3.3 \\
 & Zhong (Para) & 14.2 & 14.9 & 12.7 & 9.1 & 10.2 & 21.1 & 12.0 & 0.7 & 2.9 & 1.2 & 4.1 & 10.6 & 9.5 \\
 & SAE-DM & 27.0 & 26.9 & 21.9 & 12.4 & 23.3 & 25.2 & \textbf{17.8} & 1.5 & 3.0 & 2.2 & 11.4 & 3.6 & 14.7 \\
 & Gated-DM & 18.0 & 10.5 & 16.7 & 8.6 & 14.5 & 13.8 & 11.6 & 1.3 & 1.6 & 2.9 & 3.6 & 5.2 & 9.0 \\
 & DiffMean & \textbf{28.2} & 28.3 & 22.0 & 13.9 & \textbf{24.6} & \textbf{26.1} & 17.4 & 1.5 & 2.0 & 4.4 & \textbf{11.9} & 3.8 & 15.3 \\
 & LangFIR & 25.3 & \textbf{30.4} & \textbf{22.8} & \textbf{16.8} & 19.1 & 15.8 & 16.8 & \textbf{3.6} & \textbf{6.0} & \textbf{4.8} & 6.4 & \textbf{17.8} & \textbf{15.5} \\
\midrule
\multirow{10}{*}{Gemma 3 4B} & No Steering & 14.2 & 6.6 & 2.0 & 1.1 & 0.9 & 0.8 & 0.5 & 0.5 & 1.2 & 0.3 & 2.2 & 0.5 & 2.6 \\
 & Prompt Only & 41.1 & 46.9 & 35.3 & 42.2 & 34.4 & 50.6 & 36.1 & 25.5 & 28.6 & 19.4 & 27.7 & 37.9 & 35.5 \\
\cmidrule{2-15}
 & PCA & 12.5 & 9.8 & 2.8 & 1.1 & 1.5 & 2.4 & 0.8 & 2.3 & 2.8 & 0.1 & 2.3 & 0.6 & 3.3 \\
 & LDA & 31.5 & 27.7 & 21.0 & 18.5 & 17.3 & 25.3 & 17.4 & 2.3 & 4.7 & 1.8 & 5.2 & 9.8 & 15.2 \\
 & Zhong (Mono) & 14.0 & 15.9 & 12.0 & 11.8 & 11.6 & 21.7 & 7.4 & 4.3 & 4.4 & 2.8 & 9.8 & 10.7 & 10.5 \\
 & Zhong (Para) & 12.6 & 22.5 & 21.6 & 7.7 & 20.5 & 22.5 & 8.5 & 7.2 & 8.9 & 6.5 & 11.9 & 16.8 & 13.9 \\
 & SAE-DM & 32.2 & 36.0 & 27.4 & 28.6 & 22.6 & 35.7 & \textbf{29.0} & 12.5 & 12.0 & 10.6 & 9.7 & 22.4 & 23.2 \\
 & Gated-DM & 22.0 & 16.3 & 13.5 & 10.8 & 12.6 & 14.0 & 11.0 & 4.6 & 4.8 & 5.2 & 10.4 & 4.8 & 10.8 \\
 & DiffMean & 32.5 & 38.9 & 27.7 & 32.8 & 23.3 & 38.9 & 26.9 & 14.0 & \textbf{15.2} & 8.8 & 6.3 & 21.5 & 23.9 \\
 & LangFIR & \textbf{38.3} & \textbf{40.9} & \textbf{30.7} & \textbf{33.1} & \textbf{26.0} & \textbf{40.3} & 27.0 & \textbf{14.9} & 9.7 & \textbf{11.9} & \textbf{16.4} & \textbf{30.5} & \textbf{26.6} \\
\midrule
\multirow{10}{*}{Llama 3.1 8B} & No Steering & 3.3 & 4.3 & 1.9 & 0.8 & 0.0 & 1.9 & 0.5 & 1.3 & 2.1 & 0.6 & 0.6 & 0.2 & 1.5 \\
 & Prompt Only & 42.1 & 46.0 & 35.5 & 42.1 & 33.9 & 48.8 & 33.2 & 24.4 & 29.2 & 17.5 & 27.2 & 36.3 & 34.7 \\
\cmidrule{2-15}
 & PCA & 5.1 & 7.1 & 3.7 & 1.9 & 0.2 & 5.0 & 0.7 & 2.3 & 3.8 & 2.6 & 1.1 & 1.2 & 2.9 \\
 & LDA & 37.5 & 39.6 & \textbf{30.9} & 29.8 & 19.5 & 29.0 & 15.8 & 12.5 & 13.4 & 7.4 & 12.0 & \textbf{25.0} & 22.7 \\
 & Zhong (Mono) & 9.4 & 14.0 & 3.7 & 5.2 & 1.3 & 12.5 & 4.5 & 4.6 & 4.0 & 1.4 & 8.2 & 10.5 & 6.6 \\
 & Zhong (Para) & 10.8 & 14.4 & 6.6 & 5.7 & 1.1 & 13.0 & 3.4 & 4.6 & 3.6 & 1.5 & 7.1 & 19.2 & 7.6 \\
 & SAE-DM & 9.2 & 12.4 & 7.7 & 2.3 & 1.2 & 9.2 & 0.7 & 3.0 & 4.5 & 3.0 & 2.1 & 1.2 & 4.7 \\
 & Gated-DM & 33.2 & 36.0 & 28.1 & 26.9 & 16.7 & \textbf{34.2} & \textbf{21.2} & 6.8 & 12.8 & 5.6 & 25.6 & 24.5 & 22.6 \\
 & DiffMean & 36.0 & 37.3 & 23.6 & 28.5 & 15.6 & 33.7 & 19.7 & 6.9 & 12.0 & 5.2 & 24.4 & 23.7 & 22.2 \\
 & LangFIR & \textbf{38.4} & \textbf{42.6} & 30.8 & \textbf{32.3} & \textbf{24.1} & 30.6 & 4.5 & \textbf{16.1} & \textbf{19.9} & \textbf{9.3} & \textbf{29.2} & 2.7 & \textbf{23.4} \\
\bottomrule
\end{tabular}
}
\label{tab:accbleu_alldatasets}
\end{table*}

\begin{table*}[h!]
\centering
\caption{ACC$\times$BLEU scores on FLORES+ for Gemma 3 1B, Gemma 3 4B, Llama 3.1 8B. The best result among steering methods per language within each model is \textbf{bolded}.}
\small
\setlength{\tabcolsep}{3pt}
\resizebox{\textwidth}{!}{
\begin{tabular}{llrrrrrrrrrrrrr}
\toprule
\textbf{Model} & \textbf{Method} & es & fr & de & it & id & pt & vi & pl & cs & tr & zh & nl & \textbf{Avg} \\
\midrule
\multirow{10}{*}{Gemma 3 1B} & No Steering & 0.0 & 0.0 & 0.0 & 0.0 & 0.0 & 0.0 & 0.0 & 0.0 & 0.0 & 0.0 & 0.0 & 0.0 & 0.0 \\
 & Prompt Only & 17.9 & 25.3 & 16.8 & 14.1 & 29.1 & 26.8 & 26.3 & 7.5 & 10.4 & 7.3 & 18.0 & 15.1 & 17.9 \\
\cmidrule{2-15}
 & PCA & 0.0 & 0.0 & 0.0 & 0.0 & 0.0 & 0.0 & 0.0 & 0.0 & 0.0 & 0.0 & 0.0 & 0.0 & 0.0 \\
 & LDA & 0.0 & 0.4 & 0.0 & 0.0 & 0.0 & 0.0 & 0.0 & 0.0 & 0.0 & 0.0 & 0.0 & 0.0 & 0.0 \\
 & Zhong (Mono) & 2.4 & 3.6 & 1.8 & 1.7 & 0.9 & 5.5 & 5.1 & 0.7 & 0.7 & 0.3 & 3.0 & 2.3 & 2.3 \\
 & Zhong (Para) & 7.4 & 10.2 & 9.1 & 6.4 & 7.3 & 15.9 & 9.4 & 0.3 & 1.4 & 0.6 & 4.2 & 4.7 & 6.4 \\
 & SAE-DM & 14.4 & 23.6 & 13.8 & 5.9 & 18.6 & 17.9 & 18.4 & 0.0 & 0.4 & 1.0 & \textbf{14.3} & 1.4 & 10.8 \\
 & Gated-DM & 10.4 & 10.1 & 10.2 & 5.5 & 12.1 & 11.8 & 16.5 & 0.5 & 0.3 & \textbf{2.2} & 9.9 & 5.6 & 7.9 \\
 & DiffMean & 15.8 & 23.9 & 13.8 & 7.0 & \textbf{20.4} & 18.5 & \textbf{18.4} & 0.0 & 0.4 & 1.2 & 14.2 & 1.5 & 11.3 \\
 & LangFIR & \textbf{15.8} & \textbf{25.2} & \textbf{16.2} & \textbf{12.2} & 14.3 & \textbf{19.2} & 12.4 & \textbf{3.3} & \textbf{3.1} & 1.9 & 9.2 & \textbf{10.1} & \textbf{11.9} \\
\midrule
\multirow{10}{*}{Gemma 3 4B} & No Steering & 11.0 & 5.3 & 0.5 & 0.3 & 0.1 & 0.0 & 0.0 & 0.0 & 0.0 & 0.0 & 1.1 & 0.0 & 1.5 \\
 & Prompt Only & 25.4 & 41.5 & 31.4 & 27.8 & 37.2 & 45.0 & 38.2 & 17.5 & 25.6 & 17.6 & 34.3 & 22.4 & 30.3 \\
\cmidrule{2-15}
 & PCA & 10.2 & 9.4 & 0.7 & 1.3 & 0.6 & 0.0 & 0.0 & 0.0 & 0.0 & 0.0 & 1.5 & 0.0 & 2.0 \\
 & LDA & 18.2 & 25.8 & 21.9 & 15.1 & 21.1 & 27.7 & 25.6 & 0.7 & 4.0 & 0.6 & 8.5 & 6.2 & 14.6 \\
 & Zhong (Mono) & 12.9 & 19.2 & 10.0 & 11.6 & 8.2 & 21.2 & 8.4 & 1.4 & 2.1 & 2.1 & 8.3 & 3.8 & 9.1 \\
 & Zhong (Para) & 14.1 & 14.7 & 19.1 & 9.1 & 15.1 & 17.4 & 4.9 & 2.8 & 7.0 & 3.9 & 10.1 & 10.4 & 10.7 \\
 & SAE-DM & 21.4 & 30.2 & 20.6 & 18.6 & 25.3 & 31.2 & 26.1 & 8.3 & 6.4 & 7.2 & 19.8 & \textbf{14.9} & 19.2 \\
 & Gated-DM & 20.0 & 22.5 & 19.6 & 13.6 & 22.2 & 25.8 & 25.5 & 8.6 & 8.0 & 9.8 & 9.1 & 8.2 & 16.1 \\
 & DiffMean & 22.6 & 35.2 & 20.3 & \textbf{20.7} & 26.5 & \textbf{31.4} & 24.0 & \textbf{9.6} & \textbf{12.1} & 8.2 & 4.9 & 10.7 & 18.8 \\
 & LangFIR & \textbf{23.9} & \textbf{36.7} & \textbf{24.7} & 20.3 & \textbf{35.9} & 24.3 & \textbf{36.3} & 1.3 & 0.3 & \textbf{10.2} & \textbf{27.2} & 11.3 & \textbf{21.0} \\
\midrule
\multirow{10}{*}{Llama 3.1 8B} & No Steering & 2.5 & 1.3 & 0.2 & 0.0 & 0.0 & 0.0 & 0.0 & 0.0 & 0.0 & 0.0 & 0.2 & 0.1 & 0.4 \\
 & Prompt Only & 25.3 & 42.2 & 30.3 & 27.6 & 36.3 & 41.9 & 35.7 & 17.6 & 24.4 & 16.9 & 32.6 & 22.7 & 29.5 \\
\cmidrule{2-15}
 & PCA & 3.2 & 2.7 & 0.0 & 0.2 & 0.0 & 0.0 & 0.0 & 0.0 & 0.0 & 0.0 & 0.6 & 0.0 & 0.6 \\
 & LDA & \textbf{24.0} & 37.3 & 26.3 & 20.7 & 12.4 & 19.4 & 17.2 & 6.3 & 4.4 & 4.6 & 10.9 & 12.9 & 16.4 \\
 & Zhong (Mono) & 3.9 & 4.0 & 1.0 & 1.2 & 1.6 & 5.5 & 12.3 & 1.2 & 2.2 & 0.2 & 23.6 & 1.6 & 4.9 \\
 & Zhong (Para) & 4.6 & 4.8 & 1.1 & 0.3 & 2.0 & 3.4 & 8.8 & 1.9 & 2.2 & 0.5 & 20.5 & 1.4 & 4.3 \\
 & SAE-DM & 4.8 & 5.6 & 0.0 & 0.7 & 0.0 & 0.0 & 0.0 & 0.0 & 0.0 & 0.0 & 0.0 & 0.3 & 0.9 \\
 & Gated-DM & 20.7 & 34.3 & 21.7 & 18.1 & 24.4 & 25.8 & \textbf{29.4} & 3.5 & 11.2 & 5.0 & \textbf{30.6} & 14.0 & 19.9 \\
 & DiffMean & 23.6 & 30.0 & 20.2 & 22.2 & 21.7 & 33.3 & 27.5 & 5.5 & 12.0 & 3.4 & 26.1 & \textbf{14.7} & 20.0 \\
 & LangFIR & 23.7 & \textbf{40.0} & \textbf{28.8} & \textbf{25.5} & \textbf{27.8} & \textbf{37.9} & 11.8 & \textbf{10.7} & \textbf{15.9} & \textbf{8.0} & 30.2 & 5.2 & \textbf{22.1} \\
\bottomrule
\end{tabular}
}
\label{tab:accbleu_flores}
\end{table*}

\begin{table*}[h!]
\centering
\caption{ACC$\times$BLEU scores on WikiMatrix for Gemma 3 1B, Gemma 3 4B, Llama~3.1~8B. The best result among steering methods per language within each model is \textbf{bolded}.}
\small
\setlength{\tabcolsep}{3pt}
\resizebox{\textwidth}{!}{
\begin{tabular}{llrrrrrrrrrrrrr}
\toprule
\textbf{Model} & \textbf{Method} & es & fr & de & it & id & pt & vi & pl & cs & tr & zh & nl & \textbf{Avg} \\
\midrule
\multirow{10}{*}{Gemma 3 1B} & No Steering & 2.6 & 2.4 & 9.4 & 2.3 & 0.4 & 0.8 & 0.6 & 0.6 & 1.2 & 1.5 & 0.6 & 0.9 & 1.9 \\
 & Prompt Only & 34.4 & 34.1 & 29.6 & 28.8 & 32.3 & 39.0 & 25.4 & 17.4 & 22.2 & 8.2 & 15.3 & 33.1 & 26.6 \\
\cmidrule{2-15}
 & PCA & 1.8 & 2.3 & 9.8 & 1.3 & 0.4 & 0.2 & 0.0 & 1.7 & 2.1 & 0.1 & 0.6 & 0.9 & 1.8 \\
 & LDA & 1.3 & 2.1 & 3.7 & 2.1 & 1.4 & 0.1 & 0.0 & 0.2 & 0.9 & 0.7 & 0.1 & 1.7 & 1.2 \\
 & Zhong (Mono) & 2.8 & 5.7 & 9.2 & 3.4 & 3.5 & 8.6 & 3.0 & 0.8 & 5.5 & 2.1 & 3.0 & 5.1 & 4.4 \\
 & Zhong (Para) & 16.0 & 13.6 & 13.6 & 9.4 & 9.1 & 22.6 & 14.0 & 0.8 & 5.8 & 1.8 & 5.8 & 9.5 & 10.2 \\
 & SAE-DM & 34.8 & 25.7 & 27.7 & 18.6 & 29.0 & 21.9 & 16.2 & \textbf{2.4} & 5.6 & 3.6 & 11.8 & 4.5 & 16.8 \\
 & Gated-DM & 15.1 & 11.5 & 21.0 & 11.0 & 18.4 & 13.7 & 2.4 & 1.0 & 3.9 & 2.8 & 0.6 & 3.4 & 8.7 \\
 & DiffMean & \textbf{34.9} & 26.9 & \textbf{28.1} & 18.7 & \textbf{29.9} & 22.3 & 15.7 & \textbf{2.4} & 5.6 & 4.5 & \textbf{12.5} & 4.8 & 17.2 \\
 & LangFIR & 32.5 & \textbf{30.5} & 27.4 & \textbf{23.4} & 21.5 & \textbf{28.0} & \textbf{17.8} & 1.3 & \textbf{11.4} & \textbf{4.8} & 10.0 & \textbf{20.4} & \textbf{19.1} \\
\midrule
\multirow{10}{*}{Gemma 3 4B} & No Steering & 11.7 & 8.2 & 5.0 & 2.5 & 2.7 & 1.8 & 0.6 & 1.5 & 3.7 & 1.0 & 2.9 & 1.4 & 3.6 \\
 & Prompt Only & 45.0 & 48.4 & 37.5 & 45.3 & 37.0 & 53.4 & 36.6 & 30.7 & 34.6 & 19.8 & 28.6 & 48.2 & 38.8 \\
\cmidrule{2-15}
 & PCA & 10.1 & 13.5 & 6.6 & 1.1 & 2.5 & 4.0 & 0.6 & 2.3 & 2.8 & 0.1 & 3.0 & 0.6 & 3.9 \\
 & LDA & 36.0 & 31.2 & 18.9 & 26.0 & 21.7 & 38.9 & 17.6 & 5.9 & 9.4 & 4.3 & 3.3 & 22.2 & 19.6 \\
 & Zhong (Mono) & 18.7 & 21.7 & 14.0 & 15.4 & 17.2 & 28.0 & 7.9 & 9.1 & 6.4 & 4.7 & 13.9 & 18.9 & 14.7 \\
 & Zhong (Para) & 12.1 & 26.5 & 28.5 & 11.0 & 27.2 & 20.0 & 12.1 & 11.8 & 14.3 & 9.5 & \textbf{19.7} & 22.7 & 18.0 \\
 & SAE-DM & 40.2 & 39.0 & 30.5 & 26.2 & 29.6 & 44.1 & 29.6 & 15.7 & 19.7 & 12.3 & 9.1 & 20.9 & 26.4 \\
 & Gated-DM & 18.6 & 12.4 & 12.4 & 9.1 & 10.7 & 7.0 & 3.0 & 4.5 & 3.9 & 2.5 & 16.8 & 4.2 & 8.8 \\
 & DiffMean & 38.2 & 37.8 & 30.7 & \textbf{36.1} & 29.1 & 47.1 & \textbf{30.5} & 17.0 & \textbf{23.3} & 10.6 & 10.8 & 24.1 & 27.9 \\
 & LangFIR & \textbf{40.6} & \textbf{40.1} & \textbf{31.9} & 33.2 & \textbf{31.8} & \textbf{50.7} & 24.0 & \textbf{22.7} & 18.3 & \textbf{15.4} & 19.3 & \textbf{42.2} & \textbf{30.9} \\
\midrule
\multirow{10}{*}{Llama 3.1 8B} & No Steering & 4.5 & 6.9 & 5.5 & 2.2 & 0.1 & 3.7 & 0.6 & 3.7 & 6.3 & 1.9 & 0.5 & 0.4 & 3.0 \\
 & Prompt Only & 47.1 & 48.9 & 37.9 & 49.0 & 34.7 & 54.9 & 32.6 & 29.8 & 37.0 & 15.3 & 26.8 & 46.7 & 38.4 \\
\cmidrule{2-15}
 & PCA & 5.7 & 10.4 & 6.5 & 3.9 & 0.2 & 4.6 & 0.6 & 3.6 & 7.4 & 2.6 & 1.1 & 1.2 & 4.0 \\
 & LDA & 39.5 & 42.6 & 31.7 & 26.7 & 24.4 & 30.9 & 7.1 & 16.5 & \textbf{25.5} & \textbf{13.7} & 12.1 & \textbf{36.0} & 25.6 \\
 & Zhong (Mono) & 10.6 & 13.8 & 9.3 & 3.1 & 2.1 & 10.9 & 0.6 & 4.7 & 9.3 & 2.6 & 0.3 & 1.9 & 5.8 \\
 & Zhong (Para) & 11.0 & 19.6 & 11.0 & 8.0 & 1.0 & 17.4 & 0.6 & 6.1 & 8.2 & 2.6 & 0.6 & 28.7 & 9.6 \\
 & SAE-DM & 9.7 & 15.6 & 11.4 & 3.0 & 1.7 & 9.0 & 0.6 & 4.8 & 8.4 & 3.0 & 2.0 & 1.5 & 5.9 \\
 & Gated-DM & 37.7 & 39.7 & \textbf{32.8} & 28.0 & 12.8 & 38.7 & \textbf{15.8} & 10.2 & 20.3 & 10.4 & \textbf{28.5} & 34.8 & 25.8 \\
 & DiffMean & \textbf{44.2} & 43.1 & 25.3 & 31.5 & 13.1 & 32.6 & 12.3 & 6.9 & 17.6 & 9.3 & 26.6 & 28.4 & 24.2 \\
 & LangFIR & 42.5 & \textbf{48.7} & 32.2 & \textbf{41.2} & \textbf{26.9} & \textbf{52.9} & 1.0 & \textbf{18.9} & 24.0 & 10.6 & 28.2 & 0.2 & \textbf{27.3} \\
\bottomrule
\end{tabular}
}
\label{tab:accbleu_wikimatrix}
\end{table*}

\begin{table*}[h!]
\centering
\caption{ACC$\times$BLEU scores on Tatoeba for Gemma 3 1B, Gemma 3 4B, Llama 3.1 8B. The best result among steering methods per language within each model is \textbf{bolded}.}
\small
\setlength{\tabcolsep}{3pt}
\resizebox{\textwidth}{!}{
\begin{tabular}{llrrrrrrrrrrrrr}
\toprule
\textbf{Model} & \textbf{Method} & es & fr & de & it & id & pt & vi & pl & cs & tr & zh & nl & \textbf{Avg} \\
\midrule
\multirow{10}{*}{Gemma 3 1B} & No Steering & 0.0 & 1.0 & 5.4 & 0.0 & 0.0 & 0.0 & 0.7 & 0.0 & 0.0 & 0.0 & 0.0 & 0.0 & 0.6 \\
 & Prompt Only & 8.5 & 11.7 & 24.9 & 38.0 & 25.7 & 44.6 & 25.1 & 19.2 & 15.2 & 7.9 & 13.3 & 30.9 & 22.1 \\
\cmidrule{2-15}
 & PCA & 0.1 & 0.8 & 6.8 & 0.0 & 0.0 & 0.0 & 0.6 & 0.0 & 0.0 & 0.0 & 0.0 & 0.0 & 0.7 \\
 & LDA & 5.0 & 12.6 & 13.0 & 1.4 & 0.7 & 3.4 & 0.6 & 0.0 & 0.0 & 0.0 & 0.0 & 0.6 & 3.1 \\
 & Zhong (Mono) & 1.7 & 3.4 & 8.6 & 5.1 & 1.1 & 5.3 & 5.4 & 0.5 & 0.1 & 0.2 & 1.1 & 5.1 & 3.1 \\
 & Zhong (Para) & 19.2 & 20.9 & 15.4 & 11.4 & 14.1 & 24.7 & 12.5 & 1.1 & 1.6 & 1.1 & 2.3 & 17.7 & 11.8 \\
 & SAE-DM & 32.0 & 31.3 & 24.2 & 12.8 & 22.2 & 35.8 & 18.9 & 0.7 & 0.0 & 2.0 & 8.1 & 4.8 & 16.1 \\
 & Gated-DM & 28.6 & 9.8 & 19.1 & 9.2 & 13.0 & 15.9 & 16.0 & 2.3 & 0.6 & 3.8 & 0.3 & 6.5 & 10.4 \\
 & DiffMean & \textbf{33.9} & 34.0 & 24.0 & \textbf{16.1} & \textbf{23.5} & \textbf{37.4} & 17.9 & 0.7 & 0.1 & 7.5 & \textbf{9.1} & 5.2 & \textbf{17.4} \\
 & LangFIR & 27.7 & \textbf{35.5} & \textbf{24.7} & 14.9 & 21.4 & 0.3 & \textbf{20.2} & \textbf{6.2} & \textbf{3.3} & \textbf{7.8} & 0.1 & \textbf{23.0} & 15.4 \\
\midrule
\multirow{10}{*}{Gemma 3 4B} & No Steering & 19.9 & 6.4 & 0.4 & 0.6 & 0.0 & 0.7 & 1.0 & 0.0 & 0.0 & 0.0 & 2.7 & 0.0 & 2.6 \\
 & Prompt Only & 52.9 & 50.9 & 37.0 & 53.3 & 29.1 & 53.5 & 33.6 & 28.4 & 25.5 & 20.9 & 20.3 & 43.0 & 37.4 \\
\cmidrule{2-15}
 & PCA & 17.4 & 6.5 & 1.0 & 0.8 & 0.0 & 0.7 & 1.0 & 0.0 & 0.0 & 0.0 & 2.4 & 0.0 & 2.5 \\
 & LDA & 40.3 & 26.2 & 22.2 & 14.4 & 9.0 & 9.3 & 9.0 & 0.2 & 0.6 & 0.6 & 3.8 & 0.9 & 11.4 \\
 & Zhong (Mono) & 10.3 & 6.8 & 11.9 & 8.4 & 9.6 & 16.0 & 5.8 & 2.5 & 4.7 & 1.5 & \textbf{7.1} & 9.3 & 7.8 \\
 & Zhong (Para) & 11.6 & 26.4 & 17.2 & 3.2 & \textbf{19.2} & 30.1 & 8.6 & 7.0 & 5.4 & 6.1 & 5.8 & 17.3 & 13.2 \\
 & SAE-DM & 35.0 & 38.7 & 31.0 & 40.8 & 12.8 & 31.7 & \textbf{31.2} & 13.4 & 9.9 & \textbf{12.3} & 0.0 & 31.5 & 24.0 \\
 & Gated-DM & 27.3 & 13.9 & 8.4 & 9.7 & 4.8 & 9.3 & 4.3 & 0.5 & 2.6 & 3.4 & 5.2 & 2.1 & 7.6 \\
 & DiffMean & 36.6 & 43.7 & 32.2 & 41.5 & 14.3 & 38.2 & 26.4 & 15.4 & 10.3 & 7.5 & 3.1 & 29.8 & 24.9 \\
 & LangFIR & \textbf{50.3} & \textbf{46.0} & \textbf{35.5} & \textbf{45.9} & 10.3 & \textbf{46.1} & 20.7 & \textbf{20.7} & \textbf{10.5} & 10.1 & 2.6 & \textbf{38.0} & \textbf{28.0} \\
\midrule
\multirow{10}{*}{Llama 3.1 8B} & No Steering & 2.9 & 4.7 & 0.1 & 0.2 & 0.0 & 2.0 & 0.8 & 0.2 & 0.1 & 0.0 & 1.2 & 0.0 & 1.0 \\
 & Prompt Only & 53.9 & 46.7 & 38.2 & 49.7 & 30.8 & 49.7 & 31.3 & 25.7 & 26.2 & 20.3 & 22.2 & 39.6 & 36.2 \\
\cmidrule{2-15}
 & PCA & 6.4 & 8.2 & 0.8 & 1.5 & 0.0 & 5.4 & 0.8 & 0.9 & 0.1 & 0.0 & 1.4 & 0.0 & 2.1 \\
 & LDA & 49.0 & 38.9 & \textbf{34.7} & \textbf{42.1} & \textbf{21.6} & 36.7 & \textbf{22.9} & 14.7 & \textbf{10.3} & \textbf{3.9} & 12.9 & 26.3 & \textbf{26.2} \\
 & Zhong (Mono) & 13.7 & 24.1 & 0.8 & 11.3 & 0.3 & 21.0 & 0.8 & 7.9 & 0.6 & 0.0 & 0.6 & 28.0 & 9.1 \\
 & Zhong (Para) & 16.9 & 18.9 & 7.6 & 8.9 & 0.3 & 18.2 & 0.9 & 5.8 & 0.4 & 0.0 & 0.2 & 27.4 & 8.8 \\
 & SAE-DM & 13.0 & 15.9 & 4.0 & 3.1 & 0.7 & 9.3 & 0.9 & 1.2 & 0.6 & 0.0 & 4.2 & 1.7 & 4.6 \\
 & Gated-DM & 41.3 & 33.9 & 29.8 & 34.6 & 12.8 & \textbf{38.1} & 18.5 & 6.6 & 6.7 & 1.4 & 17.8 & 24.6 & 22.2 \\
 & DiffMean & 40.1 & 38.9 & 25.2 & 31.8 & 12.0 & 35.3 & 19.4 & 8.4 & 6.3 & 2.8 & \textbf{20.5} & \textbf{28.2} & 22.4 \\
 & LangFIR & \textbf{49.1} & \textbf{39.0} & 31.5 & 30.3 & 17.6 & 1.0 & 0.8 & \textbf{18.8} & 0.0 & 0.0 & 0.0 & 0.0 & 15.7 \\
\bottomrule
\end{tabular}
}
\label{tab:accbleu_tatoeba}
\end{table*}

\clearpage

\FloatBarrier
\subsection{ACC}
We analyze the dataset-averaged table (Table~\ref{tab:acc_alldatasets}). The per-dataset tables (Tables~\ref{tab:acc_flores}--\ref{tab:acc_tatoeba}) are mostly consistent with it. Prompt Only achieves substantially higher ACC than all steering methods across all models (82.4\%, 94.0\%, and 95.2\% on Gemma 3 1B, Gemma 3 4B, and Llama 3.1 8B, respectively). Among steering methods, LangFIR achieves the best accuracy on Gemma 3 1B (62.2\%), but trails DiffMean on Gemma 3 4B (72.2\% vs.\ 85.3\%) and Llama 3.1 8B (67.7\% vs.\ 74.2\%). PCA has the lowest ACC across all models (4.2\%, 9.7\%, and 5.8\%).

\begin{table*}[h!]
\centering
\caption{ACC scores (\%) averaged over FLORES+, WikiMatrix, and Tatoeba for Gemma 3 1B, Gemma 3 4B, Llama 3.1 8B. The best result among steering methods per language within each model is \textbf{bolded}.}
\small
\setlength{\tabcolsep}{3pt}
\resizebox{\textwidth}{!}{
\begin{tabular}{llrrrrrrrrrrrrr}
\toprule
\textbf{Model} & \textbf{Method} & es & fr & de & it & id & pt & vi & pl & cs & tr & zh & nl & \textbf{Avg} \\
\midrule
\multirow{10}{*}{Gemma 3 1B} & No Steering & 1.7 & 2.7 & 13.7 & 1.3 & 0.7 & 0.3 & 1.0 & 0.7 & 1.0 & 1.0 & 0.3 & 0.3 & 2.1 \\
 & Prompt Only & 69.3 & 71.0 & 92.3 & 81.7 & 84.3 & 88.0 & 94.3 & 87.3 & 77.3 & 79.3 & 74.7 & 89.7 & 82.4 \\
\cmidrule{2-15}
 & PCA & 3.0 & 3.0 & 24.0 & 3.0 & 2.0 & 1.0 & 2.0 & 4.0 & 4.0 & 2.0 & 1.0 & 1.0 & 4.2 \\
 & LDA & 14.0 & 12.7 & 43.5 & 4.5 & 3.5 & 5.5 & 3.0 & 2.0 & 3.0 & 3.0 & 1.0 & 6.0 & 8.5 \\
 & Zhong (Mono) & 35.0 & 33.7 & 51.7 & 19.3 & 14.0 & 37.7 & 35.7 & 23.3 & 12.7 & 14.3 & 30.3 & 39.0 & 28.9 \\
 & Zhong (Para) & 67.3 & 72.7 & 73.0 & 54.0 & 51.3 & \textbf{69.7} & 69.0 & 19.3 & \textbf{39.3} & 24.3 & 37.0 & 61.0 & 53.2 \\
 & SAE-DM & 84.7 & 77.0 & 83.7 & 45.0 & 75.7 & 68.3 & 75.7 & 5.5 & 5.5 & 22.7 & 59.0 & 9.3 & 51.0 \\
 & Gated-DM & 64.7 & 34.7 & 80.3 & 34.0 & 56.3 & 49.0 & 47.3 & 11.7 & 4.7 & 26.3 & 27.3 & 25.7 & 38.5 \\
 & DiffMean & 85.3 & 80.0 & 84.7 & 53.0 & \textbf{79.0} & \textbf{69.7} & \textbf{77.3} & 5.5 & 4.0 & 31.3 & \textbf{60.7} & 12.0 & 53.5 \\
 & LangFIR & \textbf{87.7} & \textbf{90.0} & \textbf{90.0} & \textbf{70.7} & 59.3 & 51.3 & 70.0 & \textbf{40.0} & 29.0 & \textbf{57.3} & 31.0 & \textbf{70.3} & \textbf{62.2} \\
\midrule
\multirow{10}{*}{Gemma 3 4B} & No Steering & 34.0 & 13.7 & 5.7 & 3.3 & 2.3 & 1.3 & 1.0 & 0.7 & 1.7 & 1.7 & 7.0 & 0.7 & 6.1 \\
 & Prompt Only & 94.3 & 97.7 & 97.0 & 96.0 & 85.3 & 95.7 & 95.7 & 93.3 & 89.3 & 92.0 & 94.7 & 97.0 & 94.0 \\
\cmidrule{2-15}
 & PCA & 40.0 & 28.0 & 10.7 & 5.0 & 3.0 & 4.0 & 1.5 & 5.0 & 5.0 & 3.0 & 8.7 & 2.0 & 9.7 \\
 & LDA & 77.3 & 65.3 & 69.7 & 50.7 & 49.0 & 56.3 & 50.3 & 12.7 & 16.0 & 14.3 & 19.0 & 35.0 & 43.0 \\
 & Zhong (Mono) & 45.3 & 47.7 & 48.7 & 43.7 & 36.0 & 58.0 & 38.7 & 22.0 & 17.3 & 34.0 & 53.3 & 38.3 & 40.2 \\
 & Zhong (Para) & 54.3 & 62.3 & 87.0 & 53.7 & 68.3 & 57.0 & 55.0 & 40.3 & 40.0 & 55.7 & 60.7 & 72.7 & 58.9 \\
 & SAE-DM & 94.7 & 91.0 & \textbf{96.0} & 87.0 & 78.3 & 87.0 & 92.7 & 68.7 & 49.3 & 80.7 & \textbf{67.0} & 82.0 & 81.2 \\
 & Gated-DM & 61.7 & 44.0 & 45.0 & 34.0 & 40.3 & 35.7 & 37.7 & 25.3 & 16.7 & 33.7 & 57.7 & 20.3 & 37.7 \\
 & DiffMean & 94.7 & \textbf{95.3} & 95.7 & \textbf{93.0} & \textbf{88.0} & \textbf{91.7} & \textbf{96.0} & \textbf{81.0} & \textbf{66.3} & \textbf{84.0} & 54.7 & \textbf{83.0} & \textbf{85.3} \\
 & LangFIR & \textbf{95.3} & 92.3 & 92.0 & 85.0 & 62.0 & 85.3 & 72.3 & 55.3 & 30.0 & 67.3 & 50.3 & 79.3 & 72.2 \\
\midrule
\multirow{10}{*}{Llama 3.1 8B} & No Steering & 8.3 & 8.0 & 4.3 & 1.3 & 0.3 & 2.7 & 1.0 & 3.0 & 3.0 & 1.0 & 2.0 & 0.7 & 3.0 \\
 & Prompt Only & 96.3 & 98.7 & 97.0 & 98.0 & 86.0 & 96.7 & 96.3 & 98.0 & 96.0 & 96.3 & 89.0 & 94.3 & 95.2 \\
\cmidrule{2-15}
 & PCA & 12.3 & 13.7 & 9.5 & 3.7 & 1.0 & 8.5 & 1.5 & 5.0 & 5.0 & 4.0 & 2.7 & 3.0 & 5.8 \\
 & LDA & 87.3 & 90.7 & 88.3 & 71.3 & 53.0 & 64.0 & 48.0 & 54.3 & 43.0 & 49.7 & 34.0 & 64.0 & 62.3 \\
 & Zhong (Mono) & 32.0 & 35.7 & 16.7 & 17.7 & 5.3 & 33.7 & 30.0 & 29.3 & 14.0 & 12.0 & 30.7 & 35.7 & 24.4 \\
 & Zhong (Para) & 35.3 & 39.3 & 20.7 & 17.7 & 5.7 & 32.3 & 28.0 & 33.3 & 13.7 & 12.0 & 28.0 & 52.3 & 26.5 \\
 & SAE-DM & 26.3 & 26.0 & 21.5 & 6.7 & 4.0 & 18.5 & 2.0 & 7.5 & 7.0 & 5.0 & 6.7 & 2.7 & 11.2 \\
 & Gated-DM & 87.7 & 88.7 & 83.0 & 71.7 & 47.3 & 79.7 & 69.0 & 32.7 & 50.3 & 41.3 & 84.0 & 75.7 & 67.6 \\
 & DiffMean & 90.3 & 91.0 & 86.3 & 81.7 & 50.7 & \textbf{81.0} & \textbf{73.7} & 47.7 & 57.3 & \textbf{61.3} & \textbf{88.3} & \textbf{80.7} & \textbf{74.2} \\
 & LangFIR & \textbf{95.3} & \textbf{94.3} & \textbf{94.0} & \textbf{86.0} & \textbf{65.0} & 64.3 & 12.7 & \textbf{70.3} & \textbf{75.0} & 61.0 & 81.5 & 12.5 & 67.7 \\
\bottomrule
\end{tabular}
}
\label{tab:acc_alldatasets}
\end{table*}

\begin{table*}[h!]
\centering
\caption{ACC scores (\%) on FLORES+ for Gemma 3 1B, Gemma 3 4B, Llama 3.1 8B. The best result among steering methods per language within each model is \textbf{bolded}.}
\small
\setlength{\tabcolsep}{3pt}
\resizebox{\textwidth}{!}{
\begin{tabular}{llrrrrrrrrrrrrr}
\toprule
\textbf{Model} & \textbf{Method} & es & fr & de & it & id & pt & vi & pl & cs & tr & zh & nl & \textbf{Avg} \\
\midrule
\multirow{10}{*}{Gemma 3 1B} & No Steering & 0.0 & 0.0 & 0.0 & 0.0 & 0.0 & 0.0 & 0.0 & 0.0 & 0.0 & 0.0 & 0.0 & 0.0 & 0.0 \\
 & Prompt Only & 92.0 & 93.0 & 87.0 & 76.0 & 90.0 & 86.0 & 95.0 & 83.0 & 80.0 & 76.0 & 78.0 & 90.0 & 85.5 \\
\cmidrule{2-15}
 & PCA & 0.0 & 0.0 & 0.0 & 0.0 & 0.0 & 0.0 & 0.0 & 0.0 & 0.0 & 0.0 & 0.0 & 0.0 & 0.0 \\
 & LDA & 0.0 & 1.0 & 0.0 & 0.0 & 0.0 & 0.0 & 0.0 & 0.0 & 0.0 & 0.0 & 0.0 & 0.0 & 0.1 \\
 & Zhong (Mono) & 59.0 & 34.0 & 20.0 & 30.0 & 15.0 & 47.0 & 36.0 & 42.0 & 24.0 & 24.0 & 46.0 & 34.0 & 34.2 \\
 & Zhong (Para) & 77.0 & 62.0 & 66.0 & 57.0 & 44.0 & 72.0 & 53.0 & 23.0 & \textbf{41.0} & 32.0 & 39.0 & 54.0 & 51.7 \\
 & SAE-DM & 80.0 & 72.0 & 68.0 & 33.0 & 69.0 & 69.0 & 78.0 & 0.0 & 1.0 & 12.0 & 68.0 & 6.0 & 46.3 \\
 & Gated-DM & 59.0 & 46.0 & 68.0 & 38.0 & 62.0 & 53.0 & 72.0 & 10.0 & 4.0 & 23.0 & \textbf{79.0} & 34.0 & 45.7 \\
 & DiffMean & 82.0 & 74.0 & 69.0 & 39.0 & \textbf{75.0} & 71.0 & \textbf{80.0} & 0.0 & 1.0 & 21.0 & 68.0 & 11.0 & 49.2 \\
 & LangFIR & \textbf{90.0} & \textbf{89.0} & \textbf{84.0} & \textbf{66.0} & 49.0 & \textbf{77.0} & 52.0 & \textbf{43.0} & 34.0 & \textbf{40.0} & 49.0 & \textbf{69.0} & \textbf{61.8} \\
\midrule
\multirow{10}{*}{Gemma 3 4B} & No Steering & 44.0 & 13.0 & 2.0 & 3.0 & 1.0 & 0.0 & 0.0 & 0.0 & 0.0 & 0.0 & 2.0 & 0.0 & 5.4 \\
 & Prompt Only & 99.0 & 99.0 & 100.0 & 100.0 & 92.0 & 100.0 & 98.0 & 96.0 & 98.0 & 91.0 & 99.0 & 100.0 & 97.7 \\
\cmidrule{2-15}
 & PCA & 54.0 & 39.0 & 4.0 & 6.0 & 1.0 & 0.0 & 0.0 & 0.0 & 0.0 & 0.0 & 6.0 & 0.0 & 9.2 \\
 & LDA & 86.0 & 77.0 & 83.0 & 63.0 & 56.0 & 78.0 & 80.0 & 13.0 & 22.0 & 13.0 & 28.0 & 44.0 & 53.6 \\
 & Zhong (Mono) & 59.0 & 63.0 & 45.0 & 60.0 & 30.0 & 57.0 & 47.0 & 16.0 & 14.0 & 36.0 & 52.0 & 25.0 & 42.0 \\
 & Zhong (Para) & 82.0 & 49.0 & 85.0 & 57.0 & 60.0 & 54.0 & 61.0 & 21.0 & 46.0 & 58.0 & 48.0 & 75.0 & 58.0 \\
 & SAE-DM & \textbf{99.0} & \textbf{97.0} & \textbf{99.0} & 90.0 & 85.0 & \textbf{94.0} & 92.0 & 82.0 & 49.0 & 80.0 & \textbf{88.0} & \textbf{93.0} & \textbf{87.3} \\
 & Gated-DM & 84.0 & 63.0 & 66.0 & 61.0 & 64.0 & 64.0 & 76.0 & 64.0 & 36.0 & 65.0 & 62.0 & 46.0 & 62.6 \\
 & DiffMean & \textbf{99.0} & \textbf{97.0} & 95.0 & \textbf{95.0} & 89.0 & 89.0 & 96.0 & \textbf{83.0} & \textbf{69.0} & \textbf{83.0} & 42.0 & 82.0 & 84.9 \\
 & LangFIR & 98.0 & 94.0 & 87.0 & 83.0 & \textbf{90.0} & 69.0 & \textbf{97.0} & 5.0 & 2.0 & 62.0 & 81.0 & 58.0 & 68.8 \\
\midrule
\multirow{10}{*}{Llama 3.1 8B} & No Steering & 9.0 & 3.0 & 1.0 & 0.0 & 0.0 & 0.0 & 0.0 & 0.0 & 0.0 & 0.0 & 1.0 & 1.0 & 1.3 \\
 & Prompt Only & 99.0 & 100.0 & 99.0 & 100.0 & 95.0 & 100.0 & 100.0 & 100.0 & 100.0 & 96.0 & 97.0 & 98.0 & 98.7 \\
\cmidrule{2-15}
 & PCA & 12.0 & 5.0 & 0.0 & 1.0 & 0.0 & 0.0 & 0.0 & 0.0 & 0.0 & 0.0 & 2.0 & 0.0 & 1.7 \\
 & LDA & 90.0 & 91.0 & 85.0 & 77.0 & 38.0 & 57.0 & 45.0 & 41.0 & 21.0 & 33.0 & 30.0 & 57.0 & 55.4 \\
 & Zhong (Mono) & 47.0 & 32.0 & 22.0 & 19.0 & 10.0 & 32.0 & 87.0 & 44.0 & 28.0 & 19.0 & 89.0 & 31.0 & 38.3 \\
 & Zhong (Para) & 54.0 & 33.0 & 11.0 & 17.0 & 11.0 & 28.0 & 81.0 & 57.0 & 29.0 & 19.0 & 82.0 & 21.0 & 36.9 \\
 & SAE-DM & 20.0 & 14.0 & 0.0 & 4.0 & 0.0 & 0.0 & 0.0 & 0.0 & 0.0 & 0.0 & 1.0 & 1.0 & 3.3 \\
 & Gated-DM & 98.0 & 98.0 & 84.0 & 81.0 & 74.0 & 91.0 & 95.0 & 40.0 & 67.0 & 63.0 & \textbf{99.0} & 91.0 & 81.7 \\
 & DiffMean & 99.0 & 98.0 & 98.0 & \textbf{100.0} & \textbf{81.0} & 95.0 & \textbf{98.0} & \textbf{71.0} & \textbf{83.0} & \textbf{86.0} & 94.0 & \textbf{95.0} & \textbf{91.5} \\
 & LangFIR & \textbf{100.0} & \textbf{100.0} & \textbf{100.0} & 96.0 & 80.0 & \textbf{97.0} & 34.0 & 67.0 & 73.0 & 57.0 & 90.0 & 24.0 & 76.5 \\
\bottomrule
\end{tabular}
}
\label{tab:acc_flores}
\end{table*}

\begin{table*}[h!]
\centering
\caption{ACC scores (\%) on WikiMatrix for Gemma 3 1B, Gemma 3 4B, Llama 3.1 8B. The best result among steering methods per language within each model is \textbf{bolded}.}
\small
\setlength{\tabcolsep}{3pt}
\resizebox{\textwidth}{!}{
\begin{tabular}{llrrrrrrrrrrrrr}
\toprule
\textbf{Model} & \textbf{Method} & es & fr & de & it & id & pt & vi & pl & cs & tr & zh & nl & \textbf{Avg} \\
\midrule
\multirow{10}{*}{Gemma 3 1B} & No Steering & 5.0 & 5.0 & 21.0 & 4.0 & 2.0 & 1.0 & 1.0 & 2.0 & 3.0 & 3.0 & 1.0 & 1.0 & 4.1 \\
 & Prompt Only & 85.0 & 86.0 & 91.0 & 77.0 & 84.0 & 84.0 & 90.0 & 81.0 & 71.0 & 72.0 & 65.0 & 85.0 & 80.9 \\
\cmidrule{2-15}
 & PCA & 5.0 & 4.0 & 23.0 & 3.0 & 2.0 & 1.0 & 0.0 & 4.0 & 4.0 & 2.0 & 1.0 & 1.0 & 4.2 \\
 & LDA & 15.0 & 5.0 & 32.0 & 6.0 & 4.0 & 4.0 & 1.0 & 2.0 & 3.0 & 3.0 & 1.0 & 9.0 & 7.1 \\
 & Zhong (Mono) & 17.0 & 28.0 & 61.0 & 12.0 & 13.0 & 26.0 & 14.0 & 13.0 & 12.0 & 15.0 & 20.0 & 24.0 & 21.3 \\
 & Zhong (Para) & 58.0 & 68.0 & 67.0 & 47.0 & 49.0 & 62.0 & \textbf{66.0} & \textbf{17.0} & \textbf{38.0} & 19.0 & 33.0 & 55.0 & 48.2 \\
 & SAE-DM & 88.0 & 68.0 & 88.0 & 46.0 & 74.0 & 51.0 & 57.0 & 7.0 & 10.0 & 17.0 & 53.0 & 9.0 & 47.3 \\
 & Gated-DM & 61.0 & 33.0 & 80.0 & 32.0 & 53.0 & 36.0 & 6.0 & 9.0 & 7.0 & 7.0 & 1.0 & 16.0 & 28.4 \\
 & DiffMean & 88.0 & 73.0 & \textbf{90.0} & 52.0 & \textbf{77.0} & 52.0 & 58.0 & 7.0 & 10.0 & 22.0 & \textbf{55.0} & 10.0 & 49.5 \\
 & LangFIR & \textbf{90.0} & \textbf{84.0} & \textbf{90.0} & \textbf{74.0} & 56.0 & \textbf{75.0} & \textbf{66.0} & 3.0 & 31.0 & \textbf{51.0} & 42.0 & \textbf{65.0} & \textbf{60.6} \\
\midrule
\multirow{10}{*}{Gemma 3 4B} & No Steering & 22.0 & 13.0 & 11.0 & 6.0 & 6.0 & 3.0 & 1.0 & 2.0 & 5.0 & 5.0 & 7.0 & 2.0 & 6.9 \\
 & Prompt Only & 88.0 & 95.0 & 92.0 & 92.0 & 82.0 & 91.0 & 92.0 & 93.0 & 86.0 & 90.0 & 89.0 & 95.0 & 90.4 \\
\cmidrule{2-15}
 & PCA & 30.0 & 29.0 & 22.0 & 7.0 & 5.0 & 7.0 & 1.0 & 5.0 & 5.0 & 3.0 & 9.0 & 2.0 & 10.4 \\
 & LDA & 71.0 & 66.0 & 56.0 & 60.0 & 65.0 & 74.0 & 47.0 & 24.0 & 23.0 & 24.0 & 12.0 & 56.0 & 48.2 \\
 & Zhong (Mono) & 46.0 & 60.0 & 56.0 & 44.0 & 45.0 & 66.0 & 39.0 & 42.0 & 22.0 & 43.0 & 52.0 & 45.0 & 46.7 \\
 & Zhong (Para) & 50.0 & 66.0 & 86.0 & 53.0 & 84.0 & 50.0 & 57.0 & 63.0 & 54.0 & 69.0 & \textbf{73.0} & 69.0 & 64.5 \\
 & SAE-DM & \textbf{92.0} & 85.0 & 89.0 & 74.0 & 81.0 & 80.0 & 87.0 & 64.0 & 49.0 & 69.0 & 61.0 & 71.0 & 75.2 \\
 & Gated-DM & 45.0 & 33.0 & 38.0 & 20.0 & 31.0 & 17.0 & 14.0 & 10.0 & 8.0 & 13.0 & 66.0 & 8.0 & 25.2 \\
 & DiffMean & \textbf{92.0} & \textbf{91.0} & \textbf{93.0} & \textbf{86.0} & \textbf{87.0} & 90.0 & \textbf{92.0} & 72.0 & \textbf{63.0} & \textbf{76.0} & 70.0 & 81.0 & \textbf{82.8} \\
 & LangFIR & 89.0 & 85.0 & 89.0 & 79.0 & 70.0 & \textbf{94.0} & 62.0 & \textbf{80.0} & 43.0 & 70.0 & 58.0 & \textbf{88.0} & 75.6 \\
\midrule
\multirow{10}{*}{Llama 3.1 8B} & No Steering & 8.0 & 11.0 & 11.0 & 3.0 & 1.0 & 6.0 & 1.0 & 8.0 & 8.0 & 3.0 & 1.0 & 1.0 & 5.2 \\
 & Prompt Only & 90.0 & 97.0 & 93.0 & 94.0 & 79.0 & 94.0 & 91.0 & 94.0 & 91.0 & 93.0 & 75.0 & 91.0 & 90.2 \\
\cmidrule{2-15}
 & PCA & 11.0 & 19.0 & 14.0 & 5.0 & 1.0 & 8.0 & 1.0 & 8.0 & 9.0 & 4.0 & 3.0 & 3.0 & 7.2 \\
 & LDA & 79.0 & 87.0 & 82.0 & 55.0 & 53.0 & 56.0 & 21.0 & 60.0 & 63.0 & \textbf{65.0} & 31.0 & 72.0 & 60.3 \\
 & Zhong (Mono) & 23.0 & 26.0 & 23.0 & 6.0 & 4.0 & 19.0 & 1.0 & 9.0 & 12.0 & 5.0 & 1.0 & 5.0 & 11.2 \\
 & Zhong (Para) & 21.0 & 43.0 & 30.0 & 17.0 & 4.0 & 32.0 & 1.0 & 20.0 & 10.0 & 5.0 & 1.0 & 64.0 & 20.7 \\
 & SAE-DM & 27.0 & 31.0 & 28.0 & 6.0 & 4.0 & 16.0 & 1.0 & 11.0 & 11.0 & 5.0 & 7.0 & 4.0 & 12.6 \\
 & Gated-DM & 79.0 & 86.0 & 82.0 & 61.0 & 36.0 & 67.0 & \textbf{47.0} & 33.0 & 50.0 & 42.0 & \textbf{80.0} & \textbf{76.0} & 61.6 \\
 & DiffMean & 86.0 & 89.0 & 79.0 & 77.0 & 37.0 & 59.0 & 43.0 & 40.0 & 44.0 & 47.0 & 78.0 & 74.0 & 62.8 \\
 & LangFIR & \textbf{91.0} & \textbf{97.0} & \textbf{92.0} & \textbf{95.0} & \textbf{71.0} & \textbf{95.0} & 2.0 & \textbf{67.0} & \textbf{77.0} & \textbf{65.0} & 73.0 & 1.0 & \textbf{68.8} \\
\bottomrule
\end{tabular}
}
\label{tab:acc_wikimatrix}
\end{table*}

\begin{table*}[h!]
\centering
\caption{ACC scores (\%) on Tatoeba for Gemma 3 1B, Gemma 3 4B, Llama 3.1 8B. The best result among steering methods per language within each model is \textbf{bolded}.}
\small
\setlength{\tabcolsep}{3pt}
\resizebox{\textwidth}{!}{
\begin{tabular}{llrrrrrrrrrrrrr}
\toprule
\textbf{Model} & \textbf{Method} & es & fr & de & it & id & pt & vi & pl & cs & tr & zh & nl & \textbf{Avg} \\
\midrule
\multirow{10}{*}{Gemma 3 1B} & No Steering & 0.0 & 3.0 & 20.0 & 0.0 & 0.0 & 0.0 & 2.0 & 0.0 & 0.0 & 0.0 & 0.0 & 0.0 & 2.1 \\
 & Prompt Only & 31.0 & 34.0 & 99.0 & 92.0 & 79.0 & 94.0 & 98.0 & 98.0 & 81.0 & 90.0 & 81.0 & 94.0 & 80.9 \\
\cmidrule{2-15}
 & PCA & 1.0 & 2.0 & 25.0 & 0.0 & 0.0 & 0.0 & 2.0 & 0.0 & 0.0 & 0.0 & 0.0 & 0.0 & 2.5 \\
 & LDA & 13.0 & 32.0 & 55.0 & 3.0 & 3.0 & 7.0 & 5.0 & 0.0 & 0.0 & 0.0 & 0.0 & 3.0 & 10.1 \\
 & Zhong (Mono) & 29.0 & 39.0 & 74.0 & 16.0 & 14.0 & 40.0 & 57.0 & 15.0 & 2.0 & 4.0 & 25.0 & 59.0 & 31.2 \\
 & Zhong (Para) & 67.0 & 88.0 & 86.0 & 58.0 & 61.0 & 75.0 & 88.0 & 18.0 & \textbf{39.0} & 22.0 & 39.0 & 74.0 & 59.6 \\
 & SAE-DM & \textbf{86.0} & 91.0 & 95.0 & 56.0 & 84.0 & 85.0 & 92.0 & 4.0 & 0.0 & 39.0 & 56.0 & 13.0 & 58.4 \\
 & Gated-DM & 74.0 & 25.0 & 93.0 & 32.0 & 54.0 & 58.0 & 64.0 & 16.0 & 3.0 & 49.0 & 2.0 & 27.0 & 41.4 \\
 & DiffMean & \textbf{86.0} & 93.0 & 95.0 & 68.0 & \textbf{85.0} & \textbf{86.0} & \textbf{94.0} & 4.0 & 1.0 & 51.0 & \textbf{59.0} & 15.0 & 61.4 \\
 & LangFIR & 83.0 & \textbf{97.0} & \textbf{96.0} & \textbf{72.0} & 73.0 & 2.0 & 92.0 & \textbf{74.0} & 22.0 & \textbf{81.0} & 2.0 & \textbf{77.0} & \textbf{64.2} \\
\midrule
\multirow{10}{*}{Gemma 3 4B} & No Steering & 36.0 & 15.0 & 4.0 & 1.0 & 0.0 & 1.0 & 2.0 & 0.0 & 0.0 & 0.0 & 12.0 & 0.0 & 5.9 \\
 & Prompt Only & 96.0 & 99.0 & 99.0 & 96.0 & 82.0 & 96.0 & 97.0 & 91.0 & 84.0 & 95.0 & 96.0 & 96.0 & 93.9 \\
\cmidrule{2-15}
 & PCA & 36.0 & 16.0 & 6.0 & 2.0 & 0.0 & 1.0 & 2.0 & 0.0 & 0.0 & 0.0 & 11.0 & 0.0 & 6.2 \\
 & LDA & 75.0 & 53.0 & 70.0 & 29.0 & 26.0 & 17.0 & 24.0 & 1.0 & 3.0 & 6.0 & 17.0 & 5.0 & 27.2 \\
 & Zhong (Mono) & 31.0 & 20.0 & 45.0 & 27.0 & 33.0 & 51.0 & 30.0 & 8.0 & 16.0 & 23.0 & 56.0 & 45.0 & 32.1 \\
 & Zhong (Para) & 31.0 & 72.0 & 90.0 & 51.0 & 61.0 & 67.0 & 47.0 & 37.0 & 20.0 & 40.0 & \textbf{61.0} & 74.0 & 54.3 \\
 & SAE-DM & 93.0 & 91.0 & \textbf{100.0} & 97.0 & 69.0 & 87.0 & 99.0 & 60.0 & 50.0 & \textbf{93.0} & 52.0 & 82.0 & 81.1 \\
 & Gated-DM & 56.0 & 36.0 & 31.0 & 21.0 & 26.0 & 26.0 & 23.0 & 2.0 & 6.0 & 23.0 & 45.0 & 7.0 & 25.2 \\
 & DiffMean & 93.0 & \textbf{98.0} & 99.0 & \textbf{98.0} & \textbf{88.0} & \textbf{96.0} & \textbf{100.0} & \textbf{88.0} & \textbf{67.0} & \textbf{93.0} & 52.0 & 86.0 & \textbf{88.2} \\
 & LangFIR & \textbf{99.0} & \textbf{98.0} & \textbf{100.0} & 93.0 & 26.0 & 93.0 & 58.0 & 81.0 & 45.0 & 70.0 & 12.0 & \textbf{92.0} & 72.3 \\
\midrule
\multirow{10}{*}{Llama 3.1 8B} & No Steering & 8.0 & 10.0 & 1.0 & 1.0 & 0.0 & 2.0 & 2.0 & 1.0 & 1.0 & 0.0 & 4.0 & 0.0 & 2.5 \\
 & Prompt Only & 100.0 & 99.0 & 99.0 & 100.0 & 84.0 & 96.0 & 98.0 & 100.0 & 97.0 & 100.0 & 95.0 & 94.0 & 96.8 \\
\cmidrule{2-15}
 & PCA & 14.0 & 17.0 & 5.0 & 5.0 & 0.0 & 9.0 & 2.0 & 2.0 & 1.0 & 0.0 & 3.0 & 0.0 & 4.8 \\
 & LDA & 93.0 & \textbf{94.0} & \textbf{98.0} & \textbf{82.0} & \textbf{68.0} & 79.0 & 78.0 & 62.0 & \textbf{45.0} & \textbf{51.0} & 41.0 & 63.0 & \textbf{71.2} \\
 & Zhong (Mono) & 26.0 & 49.0 & 5.0 & 28.0 & 2.0 & 50.0 & 2.0 & 35.0 & 2.0 & 0.0 & 2.0 & 71.0 & 22.7 \\
 & Zhong (Para) & 31.0 & 42.0 & 21.0 & 19.0 & 2.0 & 37.0 & 2.0 & 23.0 & 2.0 & 0.0 & 1.0 & 72.0 & 21.0 \\
 & SAE-DM & 32.0 & 33.0 & 15.0 & 10.0 & 4.0 & 21.0 & 3.0 & 4.0 & 3.0 & 0.0 & 12.0 & 3.0 & 11.7 \\
 & Gated-DM & 86.0 & 82.0 & 83.0 & 73.0 & 32.0 & 81.0 & 65.0 & 25.0 & 34.0 & 19.0 & 73.0 & 60.0 & 59.4 \\
 & DiffMean & 86.0 & 86.0 & 82.0 & 68.0 & 34.0 & \textbf{89.0} & \textbf{80.0} & 32.0 & \textbf{45.0} & \textbf{51.0} & \textbf{93.0} & \textbf{73.0} & 68.2 \\
 & LangFIR & \textbf{95.0} & 86.0 & 90.0 & 67.0 & 44.0 & 1.0 & 2.0 & \textbf{77.0} & 0.0 & 0.0 & 0.0 & 0.0 & 38.5 \\
\bottomrule
\end{tabular}
}
\label{tab:acc_tatoeba}
\end{table*}

\clearpage

\FloatBarrier
\subsection{BLEU}
We analyze the dataset-averaged table (Table~\ref{tab:bleu_alldatasets}). The per-dataset tables (Tables~\ref{tab:bleu_flores}--\ref{tab:bleu_tatoeba}) are mostly consistent with it. PCA anomalously leads BLEU across all models (38.9, 37.7, and 44.4). This is a small-sample artifact: BLEU is computed only on successfully steered samples, and PCA's extremely low ACC produces a small, unrepresentative sample. Comparing the methods that steer reliably, LangFIR's BLEU exceeds DiffMean's on Gemma 3 4B (34.9 vs.\ 27.1) and Llama 3.1 8B (35.3 vs.\ 29.8), but is lower on Gemma 3 1B (23.5 vs.\ 28.1).

\begin{table*}[h!]
\centering
\caption{BLEU scores averaged over FLORES+, WikiMatrix, and Tatoeba for Gemma 3 1B, Gemma 3 4B, Llama 3.1 8B. The best result among steering methods per language within each model is \textbf{bolded}.}
\small
\setlength{\tabcolsep}{3pt}
\resizebox{\textwidth}{!}{
\begin{tabular}{llrrrrrrrrrrrrr}
\toprule
\textbf{Model} & \textbf{Method} & es & fr & de & it & id & pt & vi & pl & cs & tr & zh & nl & \textbf{Avg} \\
\midrule
\multirow{10}{*}{Gemma 3 1B} & No Steering & 17.5 & 27.1 & 23.8 & 19.1 & 7.0 & 27.0 & 31.2 & 9.3 & 13.8 & 16.2 & 19.5 & 29.6 & 20.1 \\
 & Prompt Only & 29.1 & 33.7 & 25.7 & 32.4 & 34.5 & 41.7 & 27.2 & 16.7 & 21.0 & 9.9 & 21.0 & 29.5 & 26.9 \\
\cmidrule{2-15}
 & PCA & 21.7 & \textbf{48.3} & \textbf{35.0} & \textbf{43.1} & 21.1 & 21.4 & 28.7 & \textbf{41.7} & \textbf{51.7} & 6.6 & \textbf{58.6} & \textbf{88.7} & \textbf{38.9} \\
 & LDA & 23.9 & 39.4 & 17.6 & 40.9 & 29.5 & 25.0 & 7.1 & 11.2 & 30.7 & \textbf{23.4} & 5.3 & 18.6 & 22.7 \\
 & Zhong (Mono) & 8.9 & 13.2 & 11.9 & 21.9 & 13.5 & 19.4 & 15.0 & 3.8 & 18.5 & 6.4 & 8.6 & 12.3 & 12.8 \\
 & Zhong (Para) & 22.0 & 20.1 & 17.3 & 17.0 & 19.4 & 30.5 & 17.7 & 4.1 & 7.6 & 5.4 & 11.4 & 16.6 & 15.8 \\
 & SAE-DM & 31.5 & 35.0 & 25.8 & 27.0 & 30.9 & 37.0 & 24.2 & 25.5 & 48.8 & 11.4 & 19.2 & 36.6 & 29.4 \\
 & Gated-DM & 27.0 & 32.0 & 20.6 & 25.8 & 26.1 & 29.3 & \textbf{29.3} & 10.2 & 27.7 & 19.2 & 28.6 & 20.6 & 24.7 \\
 & DiffMean & \textbf{32.8} & 35.2 & 25.5 & 25.9 & 31.2 & \textbf{37.5} & 23.1 & 25.5 & 34.7 & 13.7 & 19.7 & 32.2 & 28.1 \\
 & LangFIR & 29.0 & 33.7 & 25.2 & 23.6 & \textbf{32.3} & 25.2 & 24.3 & 19.9 & 20.4 & 7.9 & 15.5 & 25.3 & 23.5 \\
\midrule
\multirow{10}{*}{Gemma 3 4B} & No Steering & 44.4 & 48.7 & 26.6 & 37.0 & 18.3 & 42.1 & 36.2 & 24.8 & 24.5 & 6.5 & 39.3 & 22.8 & 30.9 \\
 & Prompt Only & 43.9 & 48.1 & 36.5 & 44.2 & 40.3 & 53.1 & 37.8 & 27.5 & 32.2 & 21.1 & 29.3 & 39.3 & 37.8 \\
\cmidrule{2-15}
 & PCA & 33.6 & 37.3 & 21.8 & 25.0 & \textbf{55.3} & \textbf{62.5} & \textbf{54.4} & \textbf{46.8} & \textbf{56.2} & 3.6 & 26.5 & 29.2 & \textbf{37.7} \\
 & LDA & \textbf{41.9} & 43.4 & 30.6 & \textbf{38.9} & 35.2 & 47.6 & 35.7 & 16.8 & 26.4 & 10.7 & 26.7 & 24.1 & 31.5 \\
 & Zhong (Mono) & 31.9 & 33.5 & 24.5 & 28.5 & 31.5 & 37.0 & 19.2 & 20.6 & 24.3 & 7.8 & 18.5 & 25.9 & 25.3 \\
 & Zhong (Para) & 26.3 & 35.6 & 24.9 & 14.3 & 29.7 & 39.1 & 15.8 & 17.1 & 22.8 & 11.9 & 19.2 & 23.4 & 23.3 \\
 & SAE-DM & 34.3 & 39.9 & 28.7 & 32.7 & 28.3 & 41.6 & 31.3 & 19.0 & 24.4 & 13.3 & 12.5 & 27.9 & 27.8 \\
 & Gated-DM & 37.9 & 37.3 & 29.8 & 38.0 & 29.3 & 39.0 & 24.6 & 28.6 & 38.1 & 16.3 & 17.2 & 33.3 & 30.8 \\
 & DiffMean & 34.6 & 40.8 & 29.0 & 35.4 & 26.5 & 42.5 & 28.2 & 17.5 & 23.3 & 10.6 & 11.0 & 25.8 & 27.1 \\
 & LangFIR & 40.3 & \textbf{44.4} & \textbf{33.2} & 38.6 & 41.7 & 46.2 & 37.3 & 26.7 & 26.5 & \textbf{17.6} & \textbf{29.5} & \textbf{36.3} & 34.9 \\
\midrule
\multirow{10}{*}{Llama 3.1 8B} & No Steering & 40.3 & 51.5 & 26.4 & 32.5 & 4.4 & 53.8 & 32.6 & 21.0 & 28.6 & 21.2 & 33.4 & 18.9 & 30.4 \\
 & Prompt Only & 43.9 & 46.6 & 36.6 & 43.2 & 39.6 & 50.7 & 34.5 & 25.0 & 30.7 & 18.1 & 30.9 & 38.9 & 36.6 \\
\cmidrule{2-15}
 & PCA & 41.3 & \textbf{52.3} & 31.4 & 42.2 & 23.4 & 58.4 & \textbf{48.9} & \textbf{46.4} & 44.8 & \textbf{66.2} & \textbf{38.8} & 38.4 & \textbf{44.4} \\
 & LDA & \textbf{43.1} & 43.8 & \textbf{35.0} & \textbf{42.3} & 36.8 & 45.2 & 33.8 & 22.2 & 28.1 & 14.3 & 35.6 & 38.1 & 34.9 \\
 & Zhong (Mono) & 35.6 & 38.2 & 20.0 & 32.6 & 27.4 & 38.8 & 36.5 & 25.9 & 38.9 & 26.9 & 28.1 & 27.6 & 31.4 \\
 & Zhong (Para) & 38.5 & 35.0 & 27.6 & 31.9 & 19.4 & 38.6 & 37.6 & 19.7 & 35.6 & 26.7 & 34.6 & 29.9 & 31.3 \\
 & SAE-DM & 33.6 & 46.1 & 33.6 & 32.7 & 30.4 & 50.3 & 43.9 & 36.3 & \textbf{48.4} & 59.4 & 22.2 & \textbf{41.9} & 39.9 \\
 & Gated-DM & 38.9 & 40.8 & 33.9 & 38.6 & 36.3 & 44.4 & 31.0 & 22.0 & 25.7 & 13.3 & 30.3 & 34.0 & 32.4 \\
 & DiffMean & 40.6 & 41.4 & 27.8 & 36.6 & 32.6 & 43.3 & 27.0 & 17.0 & 22.8 & 9.8 & 28.0 & 30.8 & 29.8 \\
 & LangFIR & 40.7 & 45.2 & 32.9 & 38.4 & \textbf{37.5} & \textbf{64.9} & 41.2 & 22.8 & 26.5 & 15.2 & 36.1 & 22.4 & 35.3 \\
\bottomrule
\end{tabular}
}
\label{tab:bleu_alldatasets}
\end{table*}

\begin{table*}[h!]
\centering
\caption{BLEU scores on FLORES+ for Gemma 3 1B, Gemma 3 4B, Llama 3.1 8B. The best result among steering methods per language within each model is \textbf{bolded}.}
\small
\setlength{\tabcolsep}{3pt}
\resizebox{\textwidth}{!}{
\begin{tabular}{llrrrrrrrrrrrrr}
\toprule
\textbf{Model} & \textbf{Method} & es & fr & de & it & id & pt & vi & pl & cs & tr & zh & nl & \textbf{Avg} \\
\midrule
\multirow{10}{*}{Gemma 3 1B} & No Steering & 0.0 & 0.0 & 0.0 & 0.0 & 0.0 & 0.0 & 0.0 & 0.0 & 0.0 & 0.0 & 0.0 & 0.0 & 0.0 \\
 & Prompt Only & 19.5 & 27.2 & 19.3 & 18.6 & 32.4 & 31.2 & 27.7 & 9.0 & 13.0 & 9.6 & 23.1 & 16.8 & 20.6 \\
\cmidrule{2-15}
 & PCA & 0.0 & 0.0 & 0.0 & 0.0 & 0.0 & 0.0 & 0.0 & 0.0 & 0.0 & 0.0 & 0.0 & 0.0 & 0.0 \\
 & LDA & 0.0 & \textbf{38.0} & 0.0 & 0.0 & 0.0 & 0.0 & 0.0 & 0.0 & 0.0 & 0.0 & 0.0 & 0.0 & 3.2 \\
 & Zhong (Mono) & 4.1 & 10.7 & 9.1 & 5.5 & 6.0 & 11.8 & 14.2 & 1.6 & 2.8 & 1.3 & 6.4 & 6.8 & 6.7 \\
 & Zhong (Para) & 9.6 & 16.5 & 13.8 & 11.3 & 16.7 & 22.1 & 17.7 & 1.3 & 3.4 & 1.9 & 10.8 & 8.7 & 11.1 \\
 & SAE-DM & 18.0 & 32.7 & \textbf{20.4} & 17.8 & 27.0 & 25.9 & 23.6 & 0.0 & \textbf{42.0} & 8.1 & \textbf{21.0} & \textbf{22.8} & \textbf{21.6} \\
 & Gated-DM & 17.6 & 22.0 & 15.0 & 14.4 & 19.5 & 22.3 & 22.9 & 5.2 & 6.6 & \textbf{9.4} & 12.6 & 16.5 & 15.3 \\
 & DiffMean & \textbf{19.3} & 32.3 & 20.0 & 17.9 & 27.2 & \textbf{26.0} & 23.1 & 0.0 & \textbf{42.0} & 5.8 & 20.9 & 13.6 & 20.7 \\
 & LangFIR & 17.6 & 28.4 & 19.3 & \textbf{18.4} & \textbf{29.1} & 24.9 & \textbf{23.9} & \textbf{7.7} & 9.2 & 4.8 & 18.8 & 14.7 & 18.1 \\
\midrule
\multirow{10}{*}{Gemma 3 4B} & No Steering & 25.0 & 40.4 & 25.1 & 10.8 & 10.0 & 0.0 & 0.0 & 0.0 & 0.0 & 0.0 & 54.5 & 0.0 & 13.8 \\
 & Prompt Only & 25.6 & 41.9 & 31.4 & 27.8 & 40.4 & 45.0 & 39.0 & 18.3 & 26.1 & 19.3 & 34.6 & 22.4 & 31.0 \\
\cmidrule{2-15}
 & PCA & 18.8 & 24.2 & 18.3 & 21.6 & \textbf{61.1} & 0.0 & 0.0 & 0.0 & 0.0 & 0.0 & 24.3 & 0.0 & 14.0 \\
 & LDA & 21.2 & 33.5 & 26.4 & 23.9 & 37.7 & 35.5 & 32.0 & 5.1 & 18.4 & 4.3 & 30.5 & 14.1 & 23.5 \\
 & Zhong (Mono) & 21.9 & 30.5 & 22.2 & 19.3 & 27.2 & 37.1 & 17.8 & 8.8 & 14.7 & 5.7 & 16.0 & 15.1 & 19.7 \\
 & Zhong (Para) & 17.2 & 30.1 & 22.5 & 15.9 & 25.2 & 32.2 & 8.1 & 13.5 & 15.1 & 6.7 & 21.0 & 13.9 & 18.4 \\
 & SAE-DM & 21.6 & 31.1 & 20.8 & 20.7 & 29.7 & 33.2 & 28.3 & 10.2 & 13.0 & 9.0 & 22.5 & 16.0 & 21.3 \\
 & Gated-DM & 23.8 & 35.7 & \textbf{29.7} & 22.3 & 34.7 & \textbf{40.4} & 33.6 & 13.5 & \textbf{22.3} & 15.1 & 14.7 & 17.7 & 25.3 \\
 & DiffMean & 22.8 & 36.2 & 21.4 & 21.8 & 29.7 & 35.3 & 25.0 & 11.6 & 17.5 & 9.8 & 11.7 & 13.0 & 21.3 \\
 & LangFIR & \textbf{24.4} & \textbf{39.1} & 28.4 & \textbf{24.5} & 39.9 & 35.2 & \textbf{37.4} & \textbf{26.3} & 13.6 & \textbf{16.4} & \textbf{33.5} & \textbf{19.5} & \textbf{28.2} \\
\midrule
\multirow{10}{*}{Llama 3.1 8B} & No Steering & 27.9 & 44.3 & 18.7 & 0.0 & 0.0 & 0.0 & 0.0 & 0.0 & 0.0 & 0.0 & 24.5 & 12.6 & 10.7 \\
 & Prompt Only & 25.5 & 42.2 & 30.6 & 27.6 & 38.2 & 41.9 & 35.7 & 17.6 & 24.4 & 17.6 & 33.6 & 23.2 & 29.9 \\
\cmidrule{2-15}
 & PCA & 26.4 & \textbf{53.9} & 0.0 & 17.7 & 0.0 & 0.0 & 0.0 & 0.0 & 0.0 & 0.0 & 32.4 & 0.0 & 10.9 \\
 & LDA & \textbf{26.6} & 41.0 & \textbf{31.0} & \textbf{26.9} & 32.6 & 34.0 & \textbf{38.3} & 15.5 & 20.8 & 14.0 & \textbf{36.4} & 22.5 & \textbf{28.3} \\
 & Zhong (Mono) & 8.4 & 12.4 & 4.4 & 6.2 & 16.2 & 17.1 & 14.1 & 2.7 & 7.9 & 1.0 & 26.5 & 5.2 & 10.2 \\
 & Zhong (Para) & 8.5 & 14.7 & 10.2 & 1.6 & 17.8 & 12.0 & 10.9 & 3.3 & 7.5 & 2.5 & 25.0 & 6.9 & 10.1 \\
 & SAE-DM & 24.0 & 39.7 & 0.0 & 17.2 & 0.0 & 0.0 & 0.0 & 0.0 & 0.0 & 0.0 & 3.5 & \textbf{29.2} & 9.5 \\
 & Gated-DM & 21.1 & 35.0 & 25.8 & 22.3 & 33.0 & 28.4 & 30.9 & 8.7 & 16.8 & 8.0 & 30.9 & 15.4 & 23.0 \\
 & DiffMean & 23.8 & 30.6 & 20.6 & 22.2 & 26.8 & 35.0 & 28.0 & 7.7 & 14.5 & 4.0 & 27.8 & 15.4 & 21.4 \\
 & LangFIR & 23.7 & 40.0 & 28.8 & 26.6 & \textbf{34.7} & \textbf{39.1} & 34.6 & \textbf{15.9} & \textbf{21.7} & \textbf{14.1} & 33.6 & 21.5 & 27.9 \\
\bottomrule
\end{tabular}
}
\label{tab:bleu_flores}
\end{table*}

\begin{table*}[h!]
\centering
\caption{BLEU scores on WikiMatrix for Gemma 3 1B, Gemma 3 4B, Llama 3.1 8B. The best result among steering methods per language within each model is \textbf{bolded}.}
\small
\setlength{\tabcolsep}{3pt}
\resizebox{\textwidth}{!}{
\begin{tabular}{llrrrrrrrrrrrrr}
\toprule
\textbf{Model} & \textbf{Method} & es & fr & de & it & id & pt & vi & pl & cs & tr & zh & nl & \textbf{Avg} \\
\midrule
\multirow{10}{*}{Gemma 3 1B} & No Steering & 52.6 & 47.6 & 44.7 & 57.3 & 21.1 & 80.9 & 57.6 & 27.9 & 41.3 & 48.6 & 58.6 & 88.7 & 52.2 \\
 & Prompt Only & 40.5 & 39.6 & 32.6 & 37.4 & 38.5 & 46.4 & 28.2 & 21.5 & 31.3 & 11.3 & 23.5 & 38.9 & 32.5 \\
\cmidrule{2-15}
 & PCA & 36.7 & \textbf{56.9} & \textbf{42.7} & \textbf{43.1} & 21.1 & 21.4 & 0.0 & 41.7 & 51.7 & 6.6 & \textbf{58.6} & \textbf{88.7} & \textbf{39.1} \\
 & LDA & 9.0 & 41.0 & 11.5 & 35.3 & 34.7 & 2.1 & 1.5 & 11.2 & 30.7 & 23.4 & 5.3 & 18.4 & 18.7 \\
 & Zhong (Mono) & 16.5 & 20.3 & 15.2 & 28.3 & 27.0 & 33.3 & 21.3 & 6.2 & 45.8 & 14.0 & 14.8 & 21.3 & 22.0 \\
 & Zhong (Para) & 27.7 & 20.1 & 20.3 & 20.1 & 18.5 & 36.5 & 21.2 & 4.9 & 15.2 & 9.4 & 17.5 & 17.2 & 19.0 \\
 & SAE-DM & 39.5 & 37.8 & 31.5 & 40.4 & \textbf{39.2} & 42.9 & 28.4 & 34.2 & \textbf{55.6} & 20.9 & 22.3 & 50.3 & 36.9 \\
 & Gated-DM & 24.8 & 35.0 & 26.2 & 34.4 & 34.6 & 38.2 & \textbf{39.8} & 10.8 & 55.0 & \textbf{40.3} & \textbf{58.6} & 21.0 & 34.9 \\
 & DiffMean & \textbf{39.7} & 36.9 & 31.2 & 36.0 & 38.9 & \textbf{42.9} & 27.1 & 34.2 & \textbf{55.6} & 20.6 & 22.8 & 47.9 & 36.1 \\
 & LangFIR & 36.1 & 36.3 & 30.5 & 31.7 & 38.4 & 37.3 & 27.0 & \textbf{43.7} & 36.8 & 9.4 & 23.7 & 31.4 & 31.9 \\
\midrule
\multirow{10}{*}{Gemma 3 4B} & No Steering & 53.1 & 63.0 & 45.5 & 41.5 & 45.0 & 58.9 & 57.6 & 74.4 & 73.6 & 19.5 & 41.4 & 68.5 & 53.5 \\
 & Prompt Only & 51.2 & 50.9 & 40.8 & 49.3 & 45.1 & 58.7 & 39.8 & 33.0 & 40.3 & 22.0 & 32.1 & 50.7 & 42.8 \\
\cmidrule{2-15}
 & PCA & 33.7 & 46.7 & 30.2 & 15.8 & \textbf{49.5} & \textbf{57.8} & \textbf{57.6} & \textbf{46.8} & \textbf{56.2} & 3.6 & \textbf{33.5} & 29.2 & 38.4 \\
 & LDA & \textbf{50.7} & \textbf{47.3} & 33.7 & 43.3 & 33.3 & 52.6 & 37.4 & 24.8 & 40.9 & 18.1 & 27.1 & 39.6 & 37.4 \\
 & Zhong (Mono) & 40.7 & 36.2 & 25.0 & 35.0 & 38.2 & 42.4 & 20.2 & 21.7 & 29.1 & 11.0 & 26.8 & 42.0 & 30.7 \\
 & Zhong (Para) & 24.2 & 40.2 & 33.1 & 20.8 & 32.4 & 40.0 & 21.1 & 18.8 & 26.4 & 13.8 & 27.0 & 33.0 & 27.6 \\
 & SAE-DM & 43.7 & 45.9 & 34.3 & 35.4 & 36.6 & 55.1 & 34.1 & 24.5 & 40.3 & 17.8 & 14.9 & 29.4 & 34.3 \\
 & Gated-DM & 41.3 & 37.6 & 32.7 & \textbf{45.5} & 34.6 & 41.1 & 21.3 & 45.2 & 49.2 & 19.0 & 25.4 & \textbf{52.8} & 37.1 \\
 & DiffMean & 41.6 & 41.5 & 33.0 & 42.0 & 33.4 & 52.3 & 33.1 & 23.6 & 37.0 & 14.0 & 15.4 & 29.8 & 33.1 \\
 & LangFIR & 45.6 & 47.2 & \textbf{35.8} & 42.0 & 45.5 & 53.9 & 38.7 & 28.4 & 42.6 & \textbf{22.0} & 33.4 & 48.0 & \textbf{40.3} \\
\midrule
\multirow{10}{*}{Llama 3.1 8B} & No Steering & 56.7 & 63.1 & 50.1 & 74.0 & 13.1 & 61.5 & 57.6 & 45.8 & 78.9 & 63.7 & 46.2 & 44.1 & 54.6 \\
 & Prompt Only & 52.3 & 50.4 & 40.7 & 52.1 & 44.0 & 58.4 & 35.8 & 31.7 & 40.7 & 16.5 & 35.7 & 51.3 & 42.5 \\
\cmidrule{2-15}
 & PCA & 51.8 & \textbf{54.6} & \textbf{46.7} & \textbf{78.5} & 23.4 & 57.0 & \textbf{57.6} & 45.3 & \textbf{82.8} & \textbf{66.2} & 37.6 & 38.4 & \textbf{53.3} \\
 & LDA & 50.0 & 49.0 & 38.7 & 48.6 & 46.0 & 55.3 & 33.8 & 27.5 & 40.5 & 21.0 & 38.9 & \textbf{49.9} & 41.6 \\
 & Zhong (Mono) & 45.9 & 53.0 & 40.4 & 50.9 & \textbf{52.9} & 57.4 & \textbf{57.6} & \textbf{52.5} & 77.2 & 52.7 & 29.2 & 38.2 & 50.7 \\
 & Zhong (Para) & \textbf{52.6} & 45.5 & 36.6 & 47.2 & 26.0 & 54.4 & \textbf{57.6} & 30.6 & 81.5 & 51.0 & \textbf{61.8} & 44.8 & 49.2 \\
 & SAE-DM & 35.9 & 50.3 & 40.7 & 50.0 & 42.3 & 56.3 & \textbf{57.6} & 43.6 & 76.1 & 59.4 & 28.2 & 38.4 & 48.3 \\
 & Gated-DM & 47.8 & 46.1 & 40.0 & 45.9 & 35.6 & \textbf{57.7} & 33.6 & 30.8 & 40.6 & 24.7 & 35.7 & 45.8 & 40.4 \\
 & DiffMean & 51.3 & 48.4 & 32.1 & 40.9 & 35.4 & 55.2 & 28.6 & 17.3 & 40.0 & 19.8 & 34.1 & 38.3 & 36.8 \\
 & LangFIR & 46.7 & 50.2 & 35.0 & 43.4 & 37.9 & 55.7 & 48.8 & 28.1 & 31.2 & 16.4 & 38.6 & 23.2 & 37.9 \\
\bottomrule
\end{tabular}
}
\label{tab:bleu_wikimatrix}
\end{table*}

\begin{table*}[h!]
\centering
\caption{BLEU scores on Tatoeba for Gemma 3 1B, Gemma 3 4B, Llama 3.1 8B. The best result among steering methods per language within each model is \textbf{bolded}.}
\small
\setlength{\tabcolsep}{3pt}
\resizebox{\textwidth}{!}{
\begin{tabular}{llrrrrrrrrrrrrr}
\toprule
\textbf{Model} & \textbf{Method} & es & fr & de & it & id & pt & vi & pl & cs & tr & zh & nl & \textbf{Avg} \\
\midrule
\multirow{10}{*}{Gemma 3 1B} & No Steering & 0.0 & 33.7 & 26.9 & 0.0 & 0.0 & 0.0 & 35.9 & 0.0 & 0.0 & 0.0 & 0.0 & 0.0 & 8.0 \\
 & Prompt Only & 27.4 & 34.4 & 25.1 & 41.3 & 32.6 & 47.4 & 25.6 & 19.5 & 18.8 & 8.8 & 16.4 & 32.9 & 27.5 \\
\cmidrule{2-15}
 & PCA & 6.6 & \textbf{39.7} & \textbf{27.2} & 0.0 & 0.0 & 0.0 & \textbf{28.7} & 0.0 & 0.0 & 0.0 & 0.0 & 0.0 & 8.5 \\
 & LDA & 38.8 & 39.2 & 23.6 & \textbf{46.5} & 24.3 & \textbf{47.9} & 12.8 & 0.0 & 0.0 & 0.0 & 0.0 & 18.7 & 21.0 \\
 & Zhong (Mono) & 6.0 & 8.7 & 11.6 & 31.9 & 7.5 & 13.2 & 9.5 & 3.4 & 6.8 & 4.0 & 4.6 & 8.7 & 9.7 \\
 & Zhong (Para) & 28.7 & 23.7 & 17.9 & 19.7 & 23.0 & 32.9 & 14.2 & 6.1 & 4.2 & 4.8 & 5.9 & 23.9 & 17.1 \\
 & SAE-DM & 37.2 & 34.4 & 25.5 & 22.8 & 26.5 & 42.2 & 20.6 & \textbf{16.8} & 0.0 & 5.1 & 14.4 & \textbf{36.8} & 23.5 \\
 & Gated-DM & 38.6 & 39.1 & 20.5 & 28.6 & 24.1 & 27.4 & 25.1 & 14.6 & \textbf{21.5} & 7.7 & 14.5 & 24.2 & 23.8 \\
 & DiffMean & \textbf{39.4} & 36.6 & 25.3 & 23.6 & 27.6 & 43.5 & 19.1 & \textbf{16.8} & 6.4 & \textbf{14.7} & \textbf{15.4} & 35.0 & \textbf{25.3} \\
 & LangFIR & 33.3 & 36.6 & 25.7 & 20.6 & \textbf{29.4} & 13.4 & 22.0 & 8.4 & 15.1 & 9.6 & 3.8 & 29.9 & 20.7 \\
\midrule
\multirow{10}{*}{Gemma 3 4B} & No Steering & 55.2 & 42.6 & 9.4 & 58.7 & 0.0 & 67.3 & 51.1 & 0.0 & 0.0 & 0.0 & 22.1 & 0.0 & 25.5 \\
 & Prompt Only & 55.1 & 51.4 & 37.4 & 55.6 & 35.5 & 55.7 & 34.6 & 31.2 & 30.3 & 22.0 & 21.2 & 44.8 & 39.6 \\
\cmidrule{2-15}
 & PCA & 48.2 & 40.9 & 16.9 & 37.7 & 0.0 & \textbf{67.3} & \textbf{51.1} & 0.0 & 0.0 & 0.0 & 21.6 & 0.0 & 23.6 \\
 & LDA & \textbf{53.8} & \textbf{49.4} & 31.8 & \textbf{49.6} & 34.7 & 54.6 & 37.6 & 20.6 & 19.8 & 9.7 & \textbf{22.5} & 18.6 & 33.5 \\
 & Zhong (Mono) & 33.1 & 34.0 & 26.4 & 31.2 & 29.1 & 31.3 & 19.5 & \textbf{31.4} & 29.2 & 6.5 & 12.6 & 20.7 & 25.4 \\
 & Zhong (Para) & 37.5 & 36.6 & 19.1 & 6.2 & 31.5 & 45.0 & 18.2 & 18.9 & 26.9 & \textbf{15.3} & 9.5 & 23.4 & 24.0 \\
 & SAE-DM & 37.7 & 42.6 & 31.0 & 42.1 & 18.5 & 36.5 & 31.5 & 22.3 & 19.8 & 13.2 & 0.1 & 38.4 & 27.8 \\
 & Gated-DM & 48.7 & 38.5 & 27.1 & 46.3 & 18.5 & 35.6 & 18.9 & 27.1 & \textbf{42.9} & 14.7 & 11.5 & 29.5 & 29.9 \\
 & DiffMean & 39.3 & 44.6 & 32.6 & 42.3 & 16.2 & 39.8 & 26.4 & 17.5 & 15.3 & 8.0 & 5.9 & 34.7 & 26.9 \\
 & LangFIR & 50.8 & 46.9 & \textbf{35.5} & 49.3 & \textbf{39.7} & 49.5 & 35.7 & 25.5 & 23.3 & 14.4 & 21.7 & \textbf{41.3} & \textbf{36.1} \\
\midrule
\multirow{10}{*}{Llama 3.1 8B} & No Steering & 36.3 & 47.0 & 10.4 & 23.5 & 0.0 & 100.0 & 40.1 & 17.3 & 6.9 & 0.0 & 29.5 & 0.0 & 25.9 \\
 & Prompt Only & 53.9 & 47.2 & 38.6 & 49.7 & 36.6 & 51.8 & 32.0 & 25.7 & 27.1 & 20.3 & 23.4 & 42.2 & 37.4 \\
\cmidrule{2-15}
 & PCA & 45.7 & 48.3 & 16.0 & 30.5 & 0.0 & 59.8 & 40.1 & \textbf{47.4} & 6.9 & 0.0 & \textbf{46.5} & 0.0 & 28.4 \\
 & LDA & 52.7 & 41.4 & 35.4 & \textbf{51.3} & 31.8 & 46.5 & 29.4 & 23.7 & 23.0 & \textbf{7.7} & 31.5 & 41.7 & \textbf{34.7} \\
 & Zhong (Mono) & 52.5 & \textbf{49.3} & 15.1 & 40.5 & 13.0 & 42.0 & 37.7 & 22.5 & \textbf{31.5} & 0.0 & 28.7 & 39.4 & 31.0 \\
 & Zhong (Para) & \textbf{54.6} & 44.9 & \textbf{36.0} & 47.0 & 14.4 & 49.3 & \textbf{44.3} & 25.1 & 17.7 & 0.0 & 17.1 & 38.0 & 32.4 \\
 & SAE-DM & 40.8 & 48.2 & 26.4 & 30.8 & 18.5 & 44.4 & 30.2 & 29.0 & 20.7 & 0.0 & 35.0 & \textbf{58.1} & 31.8 \\
 & Gated-DM & 48.0 & 41.3 & 35.9 & 47.4 & \textbf{40.1} & 47.0 & 28.4 & 26.5 & 19.8 & 7.2 & 24.4 & 40.9 & 33.9 \\
 & DiffMean & 46.7 & 45.2 & 30.7 & 46.7 & 35.4 & 39.6 & 24.3 & 26.1 & 13.9 & 5.5 & 22.0 & 38.6 & 31.2 \\
 & LangFIR & 51.7 & 45.3 & 35.0 & 45.3 & 40.0 & \textbf{100.0} & 40.1 & 24.4 & 0.0 & 0.0 & 0.0 & 0.0 & 31.8 \\
\bottomrule
\end{tabular}
}
\label{tab:bleu_tatoeba}
\end{table*}

\FloatBarrier
\section{Ablation Details}
\label{sec:appendix_ablation_details}

\subsection{Ablation Setup}
\label{sec:appendix_ablation_setup}

All ablations are conducted using the same twelve languages from the FLORES+ validation split used in the steering experiment. For each hyperparameter under study, all other hyperparameters are held fixed at the validation-selected configuration from the steering experiment, so that the effect of each factor can be measured independently.

\subsection{Ablation on Llama~3.1~8B}

\begin{figure*}[ht!]
  \centering
  \begin{subfigure}[t]{0.32\textwidth}
    \centering
    \includegraphics[width=\linewidth]{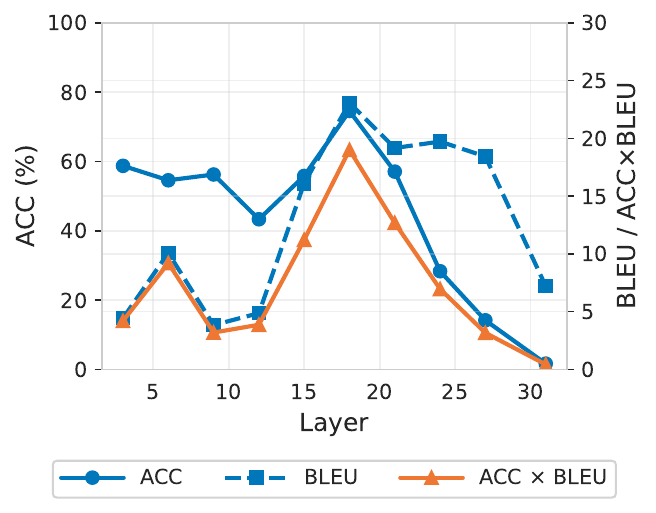}
    \caption{Intervention layer.}
    \label{llama:ablation_layer}
  \end{subfigure}
  \quad
  \begin{subfigure}[t]{0.32\textwidth}
    \centering
    \includegraphics[width=\linewidth]{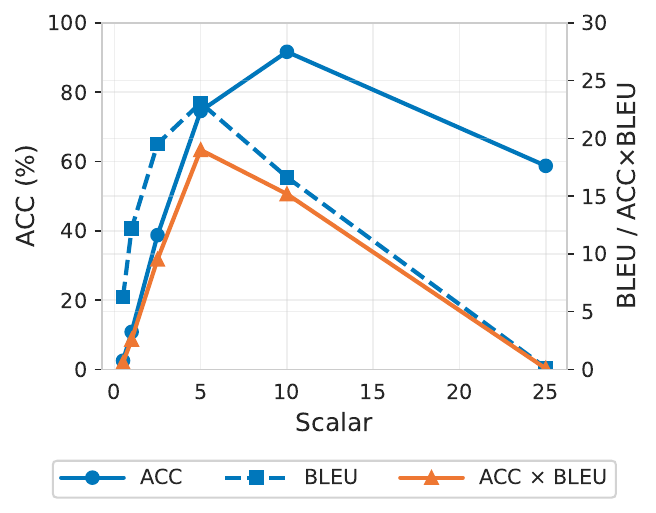}
    \caption{Scaling coefficient $\alpha$.}
    \label{llama:ablation_scalar}
  \end{subfigure}
  \caption{Ablation of intervention layer and scaling coefficient $\alpha$ on Llama~3.1~8B. (a)~ACC$\times$BLEU peaks in middle layers, where steering redirects language without disrupting content. (b)~Intermediate $\alpha$ balances steering strength and translation quality.}
  \label{llama:ablation_layer_scalar}
\end{figure*}

\textbf{Layer.}
Figure~\ref{llama:ablation_layer} shows steering metrics across intervention layers. ACC is highest in the early-to-middle layers, while BLEU increases in the middle-to-late layers, possibly because richer semantic representations in these layers~\citep{wendler-etal-2024-llamas} allow steered outputs to better preserve translation quality. The composite ACC$\times$BLEU peaks in the middle layers, where steering successfully redirects language without disrupting content fidelity. In the very late layers, both metrics decline, suggesting that the model has largely committed to its output distribution.

\textbf{Scaling coefficient.}
Figure~\ref{llama:ablation_scalar} shows a clear peak in all three metrics at an intermediate scalar value. When the scalar is too small, the steering vector has minimal effect on the model's output. When the scalar is too large, BLEU drops sharply while ACC remains relatively high. This indicates that an overly strong steering signal causes the model to produce tokens in the target language that are incoherent, so language classification succeeds but translation quality deteriorates.

\FloatBarrier

\begin{figure*}[ht!]
  \centering
  \begin{subfigure}[t]{0.32\textwidth}
    \centering
    \includegraphics[width=\linewidth]{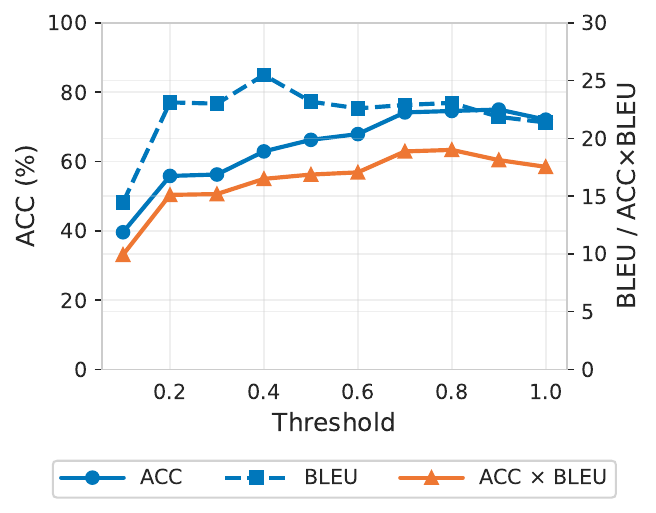}
    \caption{Frequency threshold $\tau$.}
    \label{llama:ablation_threshold}
  \end{subfigure}
  \quad
  \begin{subfigure}[t]{0.32\textwidth}
    \centering
    \includegraphics[width=\linewidth]{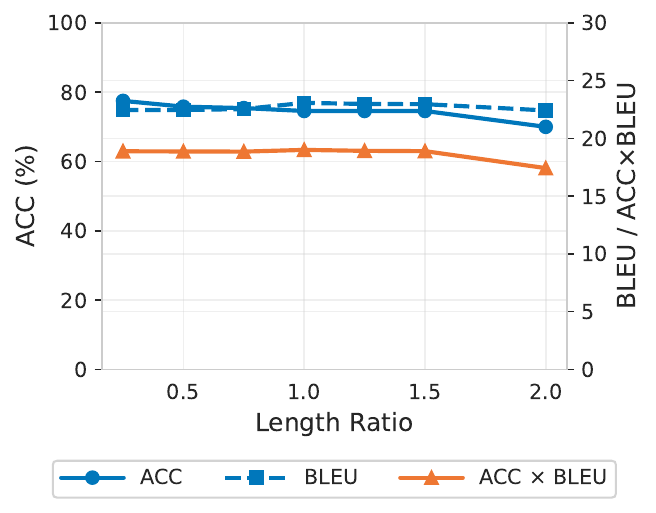}
    \caption{Sequence length ratio.}
    \label{llama:ablation_randlenratio}
  \end{subfigure}
  \caption{Ablation of frequency threshold $\tau$ and length ratio between random-token sequences and target-language sentences on Llama~3.1~8B. (a)~Performance trends upward with increasing $\tau$ and stabilizes around $\tau = 0.8$. (b)~Steering metrics remain stable across length ratios, showing that exact length matching is not required.}
  \label{llama:ablation_threshold_randlen}
\end{figure*}

\textbf{Frequency threshold.}
Figure~\ref{llama:ablation_threshold} shows that performance generally trends upward with increasing $\tau$ and stabilizes around $\tau = 0.8$, consistent with the threshold stability reported in Section~\ref{sec:analyses}. Performance drops drastically when $\tau < 0.2$, demonstrating that a sufficiently high threshold is needed to isolate a reliable set of language-consistent and language-agnostic features. A slight decline at very high thresholds ($\tau > 0.9$) suggests some weaker language-specific features are excluded.

\textbf{Random-token sequence length.}
Figure~\ref{llama:ablation_randlenratio} shows steering metrics as a function of the length ratios between random-token sequences and target-language sentences. Metrics remain stable across ratios from 0.25 to 1.5, with only a slight drop at ratio 2.0. This shows that random-token sequences need only be of similar length to the target-language sentences, and exact length matching is not required, though it is a convenient default.

\FloatBarrier

\textbf{Random seed.}
Table~\ref{tab:seed_stability_llama} reports mean and standard deviation of steering metrics across three independent runs with different random seeds used for random-token sequence generation. The standard deviation is near zero for all metrics, demonstrating that LangFIR's performance is robust to random seeds.

\begin{table}[h]
  \centering
  \caption{Steering metrics stability across 3 random seeds on Llama~3.1~8B. Near-zero standard deviations indicate robustness to random seeds.}
  \label{tab:seed_stability_llama}
  \begin{tabular}{lcc}
    \toprule
    Metric & Mean & Std \\
    \midrule
    ACC (\%) & 74.58 & 0.000 \\
    BLEU & 23.06 & 0.036 \\
    ACC $\times$ BLEU & 18.99 & 0.025 \\
    \bottomrule
  \end{tabular}
\end{table}

\FloatBarrier
\subsection{Ablation on Gemma~3~4B}

\begin{figure*}[ht!]
  \centering
  \begin{subfigure}[t]{0.32\textwidth}
    \centering
    \includegraphics[width=\linewidth]{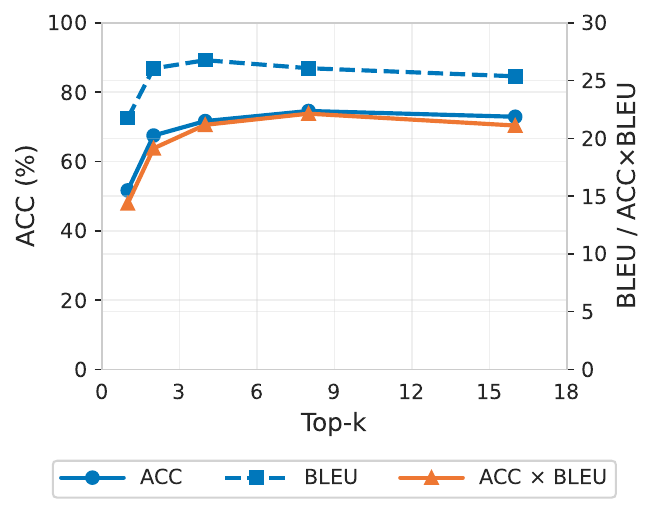}
    \caption{Top-$k$ features.}
    \label{gemma4b:ablation_topk}
  \end{subfigure}
  \hfill
  \begin{subfigure}[t]{0.32\textwidth}
    \centering
    \includegraphics[width=\linewidth]{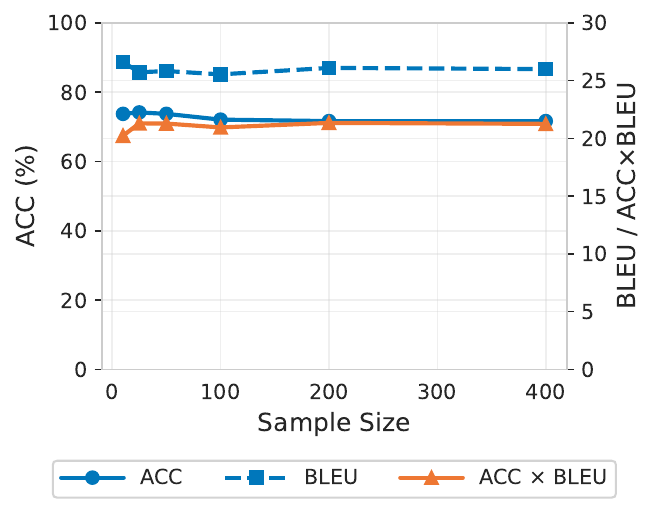}
    \caption{Sample size $N$.}
    \label{gemma4b:ablation_samplesize}
  \end{subfigure}
  \hfill
  \begin{subfigure}[t]{0.32\textwidth}
    \centering
    \includegraphics[width=\linewidth]{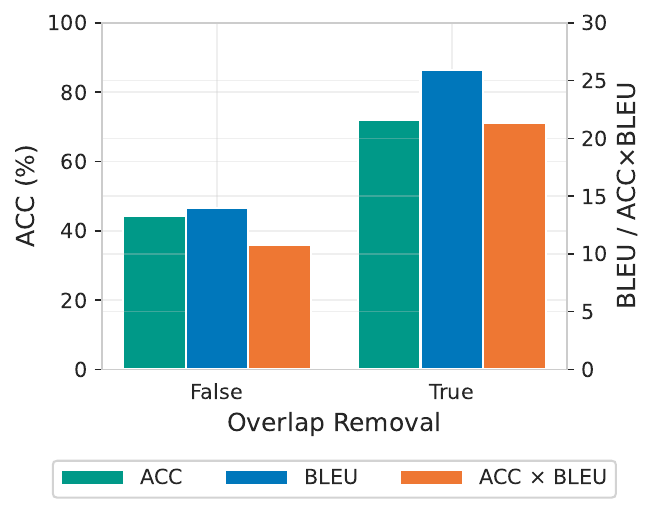}
    \caption{Feature overlap removal.}
    \label{gemma4b:ablation_ignore_shared}
  \end{subfigure}
  \caption{Ablation of number of top-$k$ language-specific features, sample size, and feature-overlap removal on Gemma~3~4B. (a)~Even $k{=}2$ achieves strong performance. (b)~Steering metrics are stable with as few as 10 sentences. (c)~Removing feature-overlap removal degrades all steering metrics significantly.}
  \label{gemma4b:ablation_topk_sample_shared}
\end{figure*}

\begin{figure*}[ht!]
  \centering
  \begin{subfigure}[t]{0.32\textwidth}
    \centering
    \includegraphics[width=\linewidth]{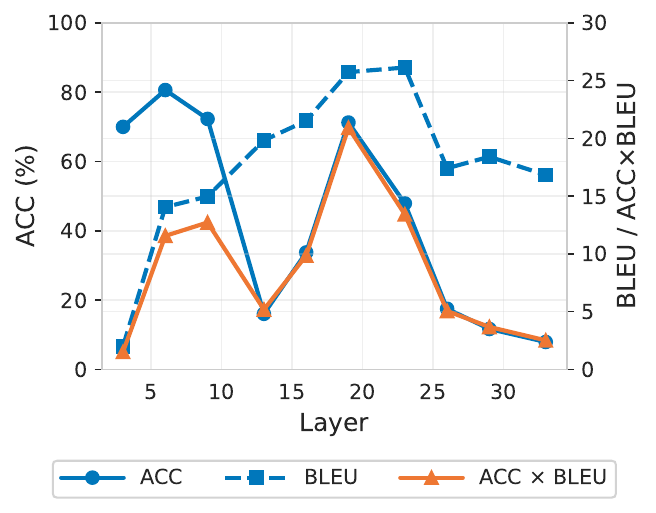}
    \caption{Intervention layer.}
    \label{gemma4b:ablation_layer}
  \end{subfigure}
  \quad
  \begin{subfigure}[t]{0.32\textwidth}
    \centering
    \includegraphics[width=\linewidth]{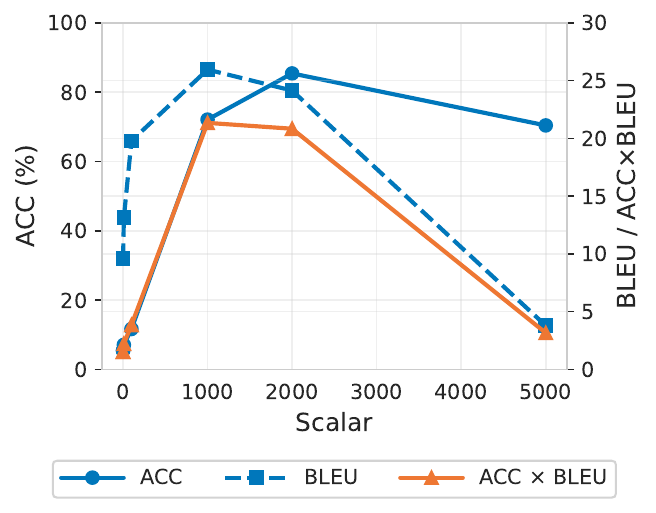}
    \caption{Scaling coefficient $\alpha$.}
    \label{gemma4b:ablation_scalar}
  \end{subfigure}
  \caption{Ablation of intervention layer and scaling coefficient $\alpha$ on Gemma~3~4B. (a)~ACC$\times$BLEU peaks in middle layers, where steering redirects language without disrupting content. (b)~Intermediate $\alpha$ balances steering strength and translation quality.}
  \label{gemma4b:ablation_layer_scalar}
\end{figure*}

\begin{figure*}[ht!]
  \centering
  \begin{subfigure}[t]{0.32\textwidth}
    \centering
    \includegraphics[width=\linewidth]{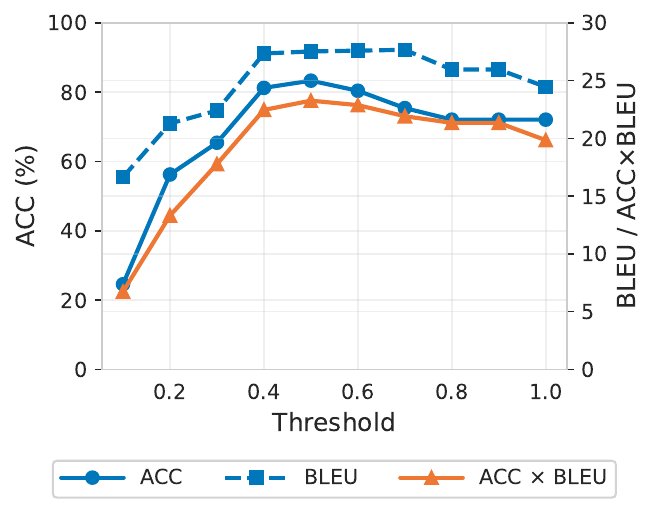}
    \caption{Frequency threshold $\tau$.}
    \label{gemma4b:ablation_threshold}
  \end{subfigure}
  \quad
  \begin{subfigure}[t]{0.32\textwidth}
    \centering
    \includegraphics[width=\linewidth]{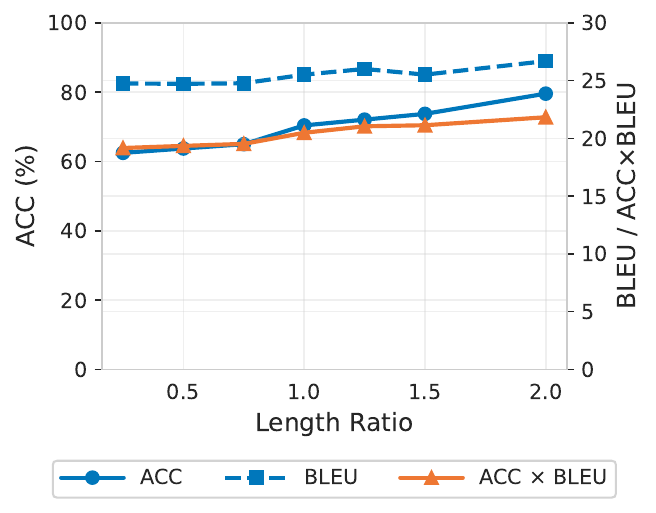}
    \caption{Sequence length ratio.}
    \label{gemma4b:ablation_randlenratio}
  \end{subfigure}
  \caption{Ablation of frequency threshold $\tau$ and length ratio between random-token sequences and target-language sentences on Gemma~3~4B. (a)~Performance generally trends upward with increasing $\tau$ and stabilizes around $\tau = 0.8$. (b)~Steering metrics remain generally stable across length ratios, showing that exact length matching is not required.}
  \label{gemma4b:ablation_threshold_randlen}
\end{figure*}

Figures~\ref{gemma4b:ablation_topk_sample_shared}--\ref{gemma4b:ablation_threshold_randlen} and Table~\ref{tab:seed_stability_gemma4b} show the ablation results for Gemma~3~4B. Results are generally consistent with Llama~3.1~8B.

\begin{table}[h]
  \centering
  \caption{Steering metrics stability across 3 random seeds on Gemma~3~4B. Small standard deviations indicate robustness to random seeds.}
  \label{tab:seed_stability_gemma4b}
  \begin{tabular}{lcc}
    \toprule
    Metric & Mean & Std \\
    \midrule
    ACC (\%) & 69.86 & 1.339 \\
    BLEU & 25.67 & 0.317 \\
    ACC $\times$ BLEU & 20.47 & 0.040 \\
    \bottomrule
  \end{tabular}
\end{table}

\FloatBarrier
\subsection{Ablation on Gemma~3~1B}

\begin{figure*}[ht!]
  \centering
  \begin{subfigure}[t]{0.32\textwidth}
    \centering
    \includegraphics[width=\linewidth]{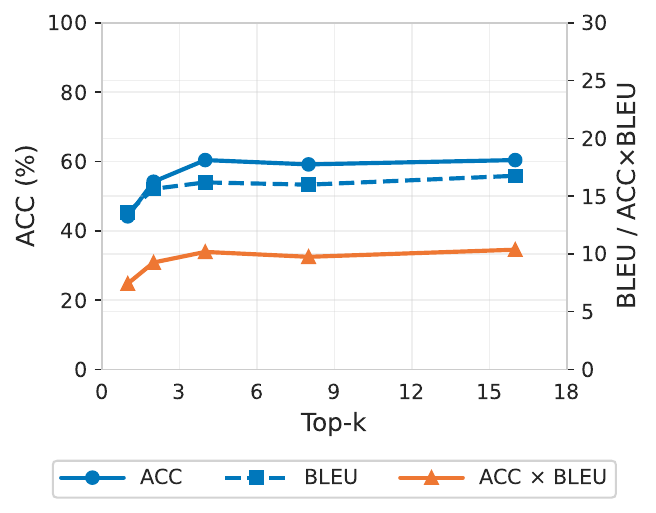}
    \caption{Top-$k$ features.}
    \label{gemma1b:ablation_topk}
  \end{subfigure}
  \hfill
  \begin{subfigure}[t]{0.32\textwidth}
    \centering
    \includegraphics[width=\linewidth]{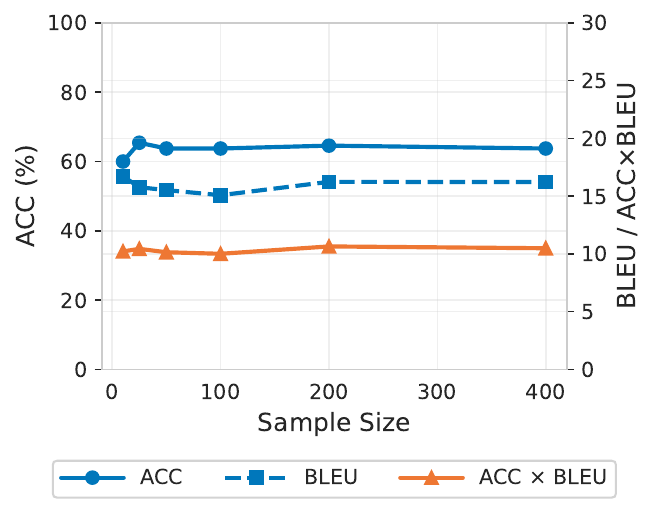}
    \caption{Sample size $N$.}
    \label{gemma1b:ablation_samplesize}
  \end{subfigure}
  \hfill
  \begin{subfigure}[t]{0.32\textwidth}
    \centering
    \includegraphics[width=\linewidth]{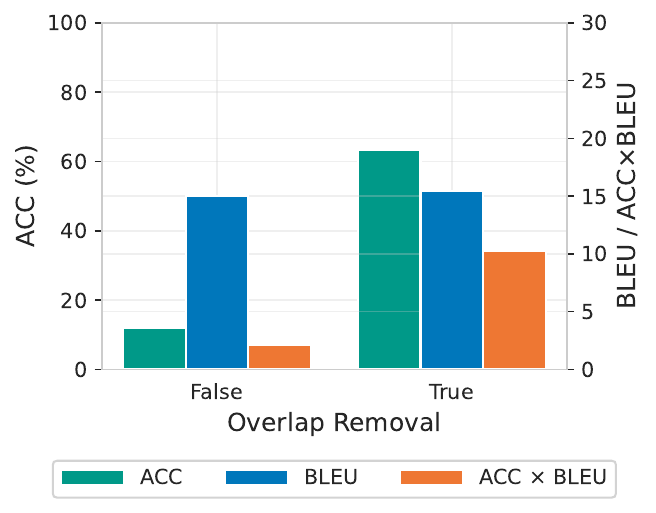}
    \caption{Feature overlap removal.}
    \label{gemma1b:ablation_ignore_shared}
  \end{subfigure}
  \caption{Ablation of number of top-$k$ language-specific features, sample size, and feature-overlap removal on Gemma~3~1B. (a)~Even $k{=}2$ achieves strong performance. (b)~Steering metrics are stable with as few as 10 sentences. (c)~Removing feature-overlap removal degrades all steering metrics dramatically.}
  \label{gemma1b:ablation_topk_sample_shared}
\end{figure*}

\begin{figure*}[ht!]
  \centering
  \begin{subfigure}[t]{0.32\textwidth}
    \centering
    \includegraphics[width=\linewidth]{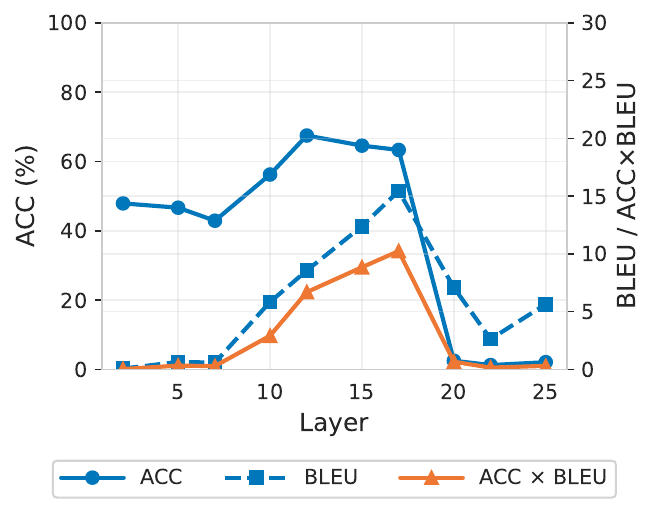}
    \caption{Intervention layer.}
    \label{gemma1b:ablation_layer}
  \end{subfigure}
  \quad
  \begin{subfigure}[t]{0.32\textwidth}
    \centering
    \includegraphics[width=\linewidth]{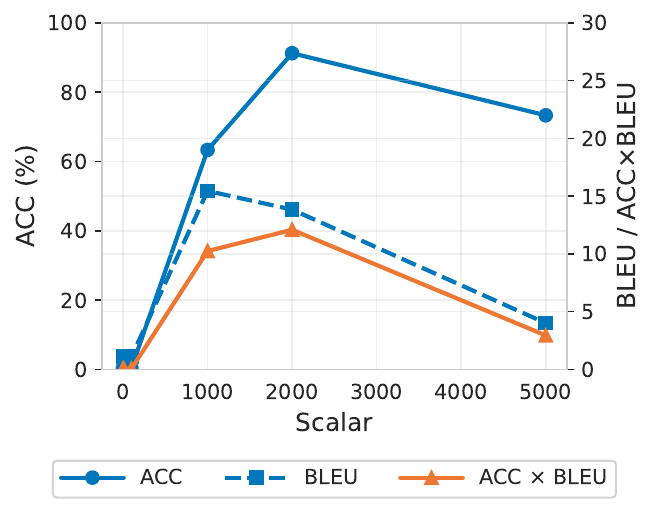}
    \caption{Scaling coefficient $\alpha$.}
    \label{gemma1b:ablation_scalar}
  \end{subfigure}
  \caption{Ablation of intervention layer and scaling coefficient $\alpha$ on Gemma~3~1B. (a)~ACC$\times$BLEU peaks in middle layers, where steering redirects language without disrupting content. (b)~Intermediate $\alpha$ balances steering strength and translation quality.}
  \label{gemma1b:ablation_layer_scalar}
\end{figure*}

\begin{figure*}[ht!]
  \centering
  \begin{subfigure}[t]{0.32\textwidth}
    \centering
    \includegraphics[width=\linewidth]{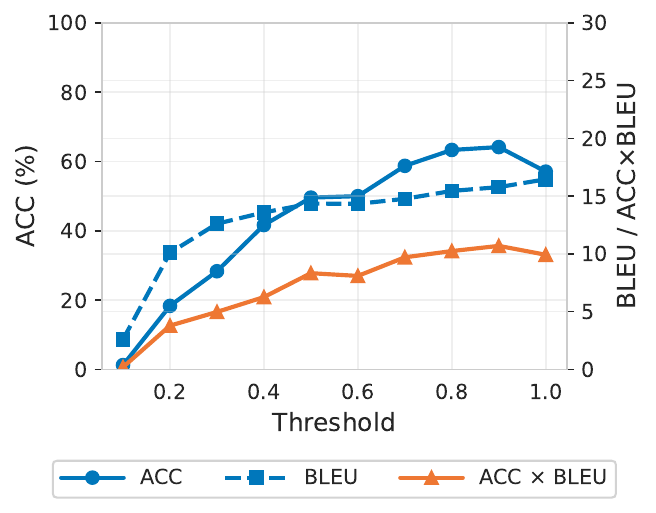}
    \caption{Frequency threshold $\tau$.}
    \label{gemma1b:ablation_threshold}
  \end{subfigure}
  \quad
  \begin{subfigure}[t]{0.32\textwidth}
    \centering
    \includegraphics[width=\linewidth]{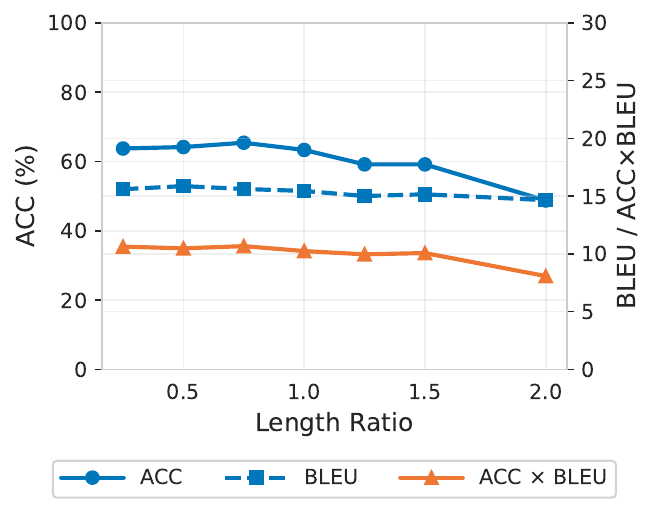}
    \caption{Sequence length ratio.}
    \label{gemma1b:ablation_randlenratio}
  \end{subfigure}
  \caption{Ablation of frequency threshold $\tau$ and length ratio between random-token sequences and target-language sentences on Gemma~3~1B. (a)~Performance trends upward with increasing $\tau$ and stabilizes around $\tau = 0.8$. (b)~Steering metrics remain stable across length ratios, showing that exact length matching is not required.}
  \label{gemma1b:ablation_threshold_randlen}
\end{figure*}

Figures~\ref{gemma1b:ablation_topk_sample_shared}--\ref{gemma1b:ablation_threshold_randlen} and Table~\ref{tab:seed_stability_gemma1b} show the ablation results for Gemma~3~1B. Results are generally consistent with Llama~3.1~8B.

\begin{table}[h]
  \centering
  \caption{Steering metrics stability across 3 random seeds on Gemma~3~1B. Small standard deviations indicate robustness to random seeds.}
  \label{tab:seed_stability_gemma1b}
  \begin{tabular}{lcc}
    \toprule
    Metric & Mean & Std \\
    \midrule
    ACC (\%) & 63.06 & 1.273 \\
    BLEU & 15.77 & 0.292 \\
    ACC $\times$ BLEU & 10.30 & 0.066 \\
    \bottomrule
  \end{tabular}
\end{table}

\FloatBarrier
\subsection{Ablation on the Filter Negative Set}
\label{sec:appendix_ablation_filter_negatives}

LangFIR uses random-token sequences as the filter negative set. We ablate this choice against two alternatives that require additional data: non-parallel text from the other eleven languages (Other Languages) and non-parallel English sentences (Non-parallel English). Table~\ref{tab:filter_negatives} reports FLORES+ results averaged over the twelve languages.

Text from other languages is slightly better than random tokens in ACC $\times$ BLEU (+0.5 to +0.8 across the three models), likely because a mixture of languages also surfaces language-agnostic features. Non-parallel English shows model-specific variation: relative to random tokens, it gains 0.7 points on Llama~3.1~8B and 1.0 on Gemma~3~1B, but loses 3.3 on Gemma~3~4B. We attribute this to two competing effects. Random tokens cast a wider net over language-agnostic features, which helps when the noise is diverse but may occasionally over-filter. Non-parallel English is a more conservative filter, but may remove English-biased features that overlap with closely related languages. Consistent with this, we verify on Llama~3.1~8B that random tokens surface a slightly larger set of language-agnostic features than Non-parallel English. English likely works as a negative set because general concepts are entangled with English representations (Section~\ref{sec:analyses}), so English sentences can also surface many language-agnostic features. Overall, random tokens remain competitive with both alternatives while requiring no data beyond the target-language corpus.

\begin{table}[h]
  \centering
  \caption{Ablation of filter negative set on Llama~3.1~8B, Gemma~3~4B, and Gemma~3~1B. The best result per metric within each model is \textbf{bolded}. Random tokens achieve competitive performance without requiring any additional language data.}
  \label{tab:filter_negatives}
  \begin{tabular}{llccc}
    \toprule
    Model & Filter Negative Set & ACC (\%) & BLEU & ACC $\times$ BLEU \\
    \midrule
    \multirow{3}{*}{Llama~3.1~8B}
      & Random Tokens        & 78.8 & 23.2 & 18.9 \\
      & Other Languages      & 80.0 & \textbf{25.6} & \textbf{19.6} \\
      & Non-parallel English & \textbf{81.2} & 23.5 & \textbf{19.6} \\
    \midrule
    \multirow{3}{*}{Gemma~3~4B}
      & Random Tokens        & 68.8 & \textbf{26.1} & 19.7 \\
      & Other Languages      & \textbf{72.1} & 25.3 & \textbf{20.5} \\
      & Non-parallel English & 54.6 & 24.3 & 16.4 \\
    \midrule
    \multirow{3}{*}{Gemma~3~1B}
      & Random Tokens        & 60.4 & \textbf{16.2} & 9.8 \\
      & Other Languages      & 63.3 & 15.8 & 10.3 \\
      & Non-parallel English & \textbf{63.7} & 15.8 & \textbf{10.8} \\
    \bottomrule
  \end{tabular}
\end{table}

\FloatBarrier
\section{Cross-Lingual Answer Generation}
\label{sec:appendix_xquad}

This appendix details the cross-lingual answer generation experiment, covering both the experimental setup and the full steering results.

\subsection{Experimental Setup}
\label{sec:appendix_xquad_setup}

The setup mirrors the multilingual generation control task (Section~\ref{sec:multilingual_experiment_setup}, Appendix~\ref{sec:appendix_experimental_details}). We describe only the task-specific differences below.

\textbf{Dataset and Preprocessing.} We use XQuAD~\citep{artetxe2020cross}, an extractive question answering dataset of 1{,}190 question--context--answer triples that is entirely parallel across 11 languages. For each language, 20 examples form a validation set for hyperparameter tuning, and 100 disjoint examples are used for final evaluation.

\textbf{Evaluation metrics.} We compute F1 and exact match (EM) against the gold target-language answer following the standard SQuAD v1.1 evaluation~\citep{rajpurkar2016squad}, which normalizes answers by lowercasing, removing punctuation and articles, and collapsing whitespace. We use character-level tokenization for Chinese.

\textbf{Hyperparameter tuning.} Hyperparameters are re-tuned for this task, selecting the configuration with the highest F1 averaged over the five languages. Table~\ref{tab:tuned_hp_xquad} reports the selected hyperparameters for each method and model on the cross-lingual answer generation task.

\begin{table}[h]
\centering
\caption{Selected hyperparameters for the cross-lingual answer generation task across all three models.}
\label{tab:tuned_hp_xquad}
\begin{tabular}{llp{6cm}}
\toprule
\textbf{Model} & \textbf{Method} & \textbf{Selected hyperparameters} \\
\midrule
\multirow{8}{*}{Gemma-3-1B}
  & PCA            & layer 12, $\alpha$=10000, top-$k$=1 \\
  & LDA            & layer 25, $\alpha$=0.1 \\
  & Zhong (Mono)   & layer 20, $\alpha$=0.6, top-$k$=200, anchor=9 \\
  & Zhong (Para)   & layer 20, $\alpha$=0.6, top-$k$=200 \\
  & SAE-DM         & layer 10, $\alpha$=1000 \\
  & Gated-DM       & layer 10, $\alpha$=1000 \\
  & DiffMean       & layer 10, $\alpha$=1000 \\
  & LangFIR        & layer 15, $\alpha$=1000, $\tau$=0.8, top-$k$=2 \\
\midrule
\multirow{8}{*}{Gemma-3-4B}
  & PCA            & layer 16, $\alpha$=10000, top-$k$=1 \\
  & LDA            & layer 29, $\alpha$=1000 \\
  & Zhong (Mono)   & layer 26, $\alpha$=0.2, top-$k$=200, anchor=12 \\
  & Zhong (Para)   & layer 23, $\alpha$=0.2, top-$k$=200 \\
  & SAE-DM         & layer 26, $\alpha$=10000 \\
  & Gated-DM       & layer 13, $\alpha$=10000 \\
  & DiffMean       & layer 26, $\alpha$=10000 \\
  & LangFIR        & layer 16, $\alpha$=1000, $\tau$=0.8, top-$k$=1 \\
\midrule
\multirow{8}{*}{Llama-3.1-8B}
  & PCA            & layer 12, $\alpha$=2.5, top-$k$=32 \\
  & LDA            & layer 21, $\alpha$=0.5 \\
  & Zhong (Mono)   & layer 12, $\alpha$=0.2, top-$k$=200, anchor=11 \\
  & Zhong (Para)   & layer 12, $\alpha$=0.2, top-$k$=400 \\
  & SAE-DM         & layer 21, $\alpha$=2.5 \\
  & Gated-DM       & layer 21, $\alpha$=10 \\
  & DiffMean       & layer 21, $\alpha$=10 \\
  & LangFIR        & layer 18, $\alpha$=10, $\tau$=1.0, top-$k$=100 \\
\bottomrule
\end{tabular}
\end{table}

\FloatBarrier
\subsection{Results}
\label{sec:appendix_xquad_results}

Table~\ref{tab:xquad_em} reports average EM for the cross-lingual answer generation experiment, and Tables~\ref{tab:xquad_f1_full} and~\ref{tab:xquad_em_full} report the full per-language F1 and EM results for all three models. The same pattern as F1 holds on EM (Table~\ref{tab:xquad_em}): every steering method beats Prompt Only on average. Among steering methods, Gated-DiffMean leads on average (8.9) while LangFIR remains competitive (8.3) and is best on Gemma 3 4B.

\begin{table*}[h]
\centering
\caption{Cross-lingual answer generation results: EM averaged over five target languages (es, de, vi, tr, zh). The best result among steering methods within each model is \textbf{bolded}.}
\label{tab:xquad_em}
\small
\setlength{\tabcolsep}{4pt}
\resizebox{\textwidth}{!}{%
\begin{tabular}{l cc @{\hspace{1.4em}} cccccccc}
\toprule
& \multicolumn{2}{c}{\textbf{Non-steering}} & \multicolumn{8}{c}{\textbf{Steering}} \\
\cmidrule(lr){2-3} \cmidrule(lr){4-11}
\textbf{Model} & No Steering & Prompt Only & PCA & LDA & Zhong (M) & Zhong (P) & SAE-DM & Gated-DM & DiffMean & LangFIR \\
\midrule
Gemma 3 1B   & 7.8 & 7.2 & 5.8 & 7.8 & 6.8 & \textbf{8.2} & 7.0 & 7.8 & 7.0 & 6.8 \\
Gemma 3 4B   & 8.8 & 3.8 & 9.8 & 8.8 & 6.2 & 6.0 & 9.8 & 9.0 & 9.2 & \textbf{10.0} \\
Llama 3.1 8B & 6.8 & 6.2 & 5.8 & 6.6 & 5.0 & 5.2 & 5.8 & \textbf{9.8} & \textbf{9.8} & 8.2 \\
\midrule
\textbf{Avg} & 7.8 & 5.7 & 7.1 & 7.7 & 6.0 & 6.5 & 7.5 & \textbf{8.9} & 8.7 & 8.3 \\
\bottomrule
\end{tabular}%
}
\end{table*}

\begin{table*}[h!]
    \centering
    \caption{Per-language F1 scores for Gemma 3 1B, Gemma 3 4B, and Llama 3.1 8B. Best result among steering methods within each model is \textbf{bolded}.}
    \label{tab:xquad_f1_full}
    \footnotesize
    \setlength{\tabcolsep}{5pt}
    \begin{tabular}{llrrrrrr}
    \toprule
    \textbf{Model} & \textbf{Method} & es & de & vi & tr & zh & \textbf{Avg} \\
    \midrule
    \multirow{10}{*}{Gemma 3 1B} & No Steering & 10.1 & 14.9 & 15.5 & 11.9 & 10.9 & 12.7 \\
     & Prompt Only & 9.9 & 14.5 & 15.0 & 11.1 & 10.8 & 12.3 \\
    \cmidrule{2-8}
     & PCA & 8.4 & 12.8 & 12.5 & 10.0 & 9.1 & 10.6 \\
     & LDA & 10.1 & 14.9 & 15.5 & 11.9 & 10.9 & 12.7 \\
     & Zhong (Mono) & 10.5 & 13.4 & 9.6 & 11.0 & 11.3 & 11.1 \\
     & Zhong (Para) & 14.1 & \textbf{19.4} & 12.3 & 13.4 & 16.3 & 15.1 \\
     & SAE-DM & 14.5 & 16.1 & 16.6 & 12.2 & 7.2 & 13.3 \\
     & Gated-DM & 11.0 & 15.2 & 15.9 & 12.0 & 14.0 & 13.6 \\
     & DiffMean & 13.7 & 16.0 & \textbf{20.2} & \textbf{13.5} & 8.6 & 14.4 \\
     & LangFIR & \textbf{16.4} & 15.6 & 15.4 & 13.3 & \textbf{22.8} & \textbf{16.7} \\
    \midrule
    \multirow{10}{*}{Gemma 3 4B} & No Steering & 13.6 & 17.2 & 17.5 & 13.3 & 11.5 & 14.6 \\
     & Prompt Only & 5.5 & 6.7 & 8.7 & 10.2 & 8.3 & 7.9 \\
    \cmidrule{2-8}
     & PCA & 14.3 & 18.4 & 18.2 & 14.1 & 12.7 & 15.6 \\
     & LDA & 14.3 & 16.5 & 17.6 & 13.4 & 11.5 & 14.7 \\
     & Zhong (Mono) & 9.9 & 15.2 & 11.7 & 11.3 & 8.1 & 11.3 \\
     & Zhong (Para) & 13.2 & 13.8 & 11.6 & 11.2 & 16.7 & 13.3 \\
     & SAE-DM & \textbf{17.9} & \textbf{24.6} & \textbf{26.4} & 15.9 & 27.6 & \textbf{22.5} \\
     & Gated-DM & 15.2 & 18.2 & 23.6 & 17.3 & 22.3 & 19.3 \\
     & DiffMean & 16.1 & 22.3 & 24.9 & \textbf{18.0} & 27.4 & 21.7 \\
     & LangFIR & 13.0 & 17.0 & 18.5 & 15.0 & \textbf{36.4} & 20.0 \\
    \midrule
    \multirow{10}{*}{Llama 3.1 8B} & No Steering & 11.2 & 14.7 & 15.3 & 11.3 & 9.0 & 12.3 \\
     & Prompt Only & 9.6 & 14.3 & 15.3 & 11.3 & 9.9 & 12.1 \\
    \cmidrule{2-8}
     & PCA & 10.5 & 13.5 & 15.0 & 9.4 & 9.5 & 11.6 \\
     & LDA & 10.9 & 14.6 & 14.6 & 10.9 & 7.9 & 11.8 \\
     & Zhong (Mono) & 9.3 & 10.6 & 11.4 & 12.8 & 8.8 & 10.6 \\
     & Zhong (Para) & 9.6 & 11.1 & 11.0 & 13.1 & 10.9 & 11.2 \\
     & SAE-DM & 10.2 & 13.8 & 13.8 & 9.9 & 7.6 & 11.1 \\
     & Gated-DM & 18.6 & \textbf{23.6} & 23.5 & 14.7 & 23.4 & 20.8 \\
     & DiffMean & 18.9 & 23.4 & \textbf{23.6} & 14.8 & 22.8 & 20.7 \\
     & LangFIR & \textbf{22.0} & 20.4 & 19.5 & \textbf{21.1} & \textbf{25.3} & \textbf{21.6} \\
    \bottomrule
    \end{tabular}
    \end{table*}

\begin{table*}[h!]
    \centering
    \caption{Per-language EM scores for Gemma 3 1B, Gemma 3 4B, and Llama 3.1 8B. Best result among steering methods within each model is \textbf{bolded}.}
    \label{tab:xquad_em_full}
    \footnotesize
    \setlength{\tabcolsep}{5pt}
    \begin{tabular}{llrrrrrr}
    \toprule
    \textbf{Model} & \textbf{Method} & es & de & vi & tr & zh & \textbf{Avg} \\
    \midrule
    \multirow{10}{*}{Gemma 3 1B} & No Steering & 7.0 & 12.0 & 9.0 & 8.0 & 3.0 & 7.8 \\
     & Prompt Only & 6.0 & 11.0 & 8.0 & 8.0 & 3.0 & 7.2 \\
    \cmidrule{2-8}
     & PCA & 5.0 & 9.0 & 6.0 & 7.0 & 2.0 & 5.8 \\
     & LDA & \textbf{7.0} & 12.0 & \textbf{9.0} & 8.0 & \textbf{3.0} & 7.8 \\
     & Zhong (Mono) & \textbf{7.0} & 11.0 & 6.0 & 8.0 & 2.0 & 6.8 \\
     & Zhong (Para) & \textbf{7.0} & \textbf{15.0} & 7.0 & \textbf{9.0} & \textbf{3.0} & \textbf{8.2} \\
     & SAE-DM & \textbf{7.0} & 11.0 & 8.0 & \textbf{9.0} & 0.0 & 7.0 \\
     & Gated-DM & \textbf{7.0} & 12.0 & \textbf{9.0} & 8.0 & \textbf{3.0} & 7.8 \\
     & DiffMean & 6.0 & 11.0 & 8.0 & \textbf{9.0} & 1.0 & 7.0 \\
     & LangFIR & 5.0 & 11.0 & 7.0 & \textbf{9.0} & 2.0 & 6.8 \\
    \midrule
    \multirow{10}{*}{Gemma 3 4B} & No Steering & 9.0 & 13.0 & 10.0 & 9.0 & 3.0 & 8.8 \\
     & Prompt Only & 1.0 & 1.0 & 6.0 & 7.0 & 4.0 & 3.8 \\
    \cmidrule{2-8}
     & PCA & \textbf{10.0} & \textbf{14.0} & 11.0 & 10.0 & 4.0 & 9.8 \\
     & LDA & \textbf{10.0} & 12.0 & 10.0 & 9.0 & 3.0 & 8.8 \\
     & Zhong (Mono) & 6.0 & 9.0 & 5.0 & 8.0 & 3.0 & 6.2 \\
     & Zhong (Para) & 7.0 & 9.0 & 4.0 & 7.0 & 3.0 & 6.0 \\
     & SAE-DM & 9.0 & \textbf{14.0} & \textbf{12.0} & 10.0 & 4.0 & 9.8 \\
     & Gated-DM & 8.0 & 12.0 & 11.0 & \textbf{11.0} & 3.0 & 9.0 \\
     & DiffMean & 7.0 & 13.0 & \textbf{12.0} & 10.0 & 4.0 & 9.2 \\
     & LangFIR & 9.0 & 13.0 & 11.0 & 10.0 & \textbf{7.0} & \textbf{10.0} \\
    \midrule
    \multirow{10}{*}{Llama 3.1 8B} & No Steering & 6.0 & 9.0 & 8.0 & 7.0 & 4.0 & 6.8 \\
     & Prompt Only & 4.0 & 8.0 & 8.0 & 7.0 & 4.0 & 6.2 \\
    \cmidrule{2-8}
     & PCA & 6.0 & 8.0 & 7.0 & 5.0 & 3.0 & 5.8 \\
     & LDA & 6.0 & 9.0 & 7.0 & 7.0 & 4.0 & 6.6 \\
     & Zhong (Mono) & 5.0 & 6.0 & 5.0 & 8.0 & 1.0 & 5.0 \\
     & Zhong (Para) & 5.0 & 6.0 & 5.0 & 8.0 & 2.0 & 5.2 \\
     & SAE-DM & 5.0 & 8.0 & 7.0 & 6.0 & 3.0 & 5.8 \\
     & Gated-DM & \textbf{10.0} & \textbf{15.0} & \textbf{10.0} & 9.0 & \textbf{5.0} & \textbf{9.8} \\
     & DiffMean & \textbf{10.0} & \textbf{15.0} & \textbf{10.0} & 9.0 & \textbf{5.0} & \textbf{9.8} \\
     & LangFIR & 8.0 & 10.0 & 8.0 & \textbf{11.0} & 4.0 & 8.2 \\
    \bottomrule
    \end{tabular}
    \end{table*}

\end{document}